\documentclass{article}

\PassOptionsToPackage{numbers, compress}{natbib}
\usepackage[preprint]{neurips_2026}

\usepackage[utf8]{inputenc} 
\usepackage[T1]{fontenc}    
\usepackage{hyperref}       
\usepackage{url}            
\usepackage{booktabs}       
\usepackage{amsfonts}       
\usepackage{nicefrac}       
\usepackage{microtype}      
\usepackage{xcolor}         
\usepackage{multirow}
\usepackage{bm}
\usepackage{amsmath}
\usepackage{amsfonts}
\usepackage{graphicx} 

\usepackage{array}
\usepackage{colortbl}
\usepackage{siunitx}
\usepackage{bigstrut}
\usepackage{float}
\usepackage{booktabs}       
\usepackage{amsfonts}       
\usepackage{nicefrac}       
\usepackage{xcolor}         
\usepackage{adjustbox}
\usepackage{siunitx}
\usepackage{tikz}
\usepackage{gensymb}
\usepackage{amsmath}
\usepackage{tabularray}

\usepackage{calculator}

\usepackage{longtable}
\usepackage{adjustbox}

\usepackage{pgfplots}
\usepackage{wrapfig}

\usepackage{pifont}
\usepackage{dsfont}
\usepackage{etoolbox}
\usepackage{color}

\newif\ifshowedits

\newcommand{\addeditor}[3]{%
  \definecolor{#1color}{rgb}{#3}
  \expandafter\newcommand\csname #1\endcsname[1]{%
  \ifshowedits
    {\color{#1color} ##1}%
  \else
    {##1}%
  \fi
  }%
  \expandafter\newcommand\csname #1rmk\endcsname[1]{%
  \ifshowedits
    {\color{#1color} {\bf [#2: ##1]}}
  \fi
  }%
  \expandafter\newcommand\csname #1rpl\endcsname[2]{%
  \ifshowedits
    {\color{#1color} ##1 \sout{##2}}
  \else
    {##1}
  \fi
  }%
}

\newcommand{\createtextvar}[1]{
  \expandafter\newcommand\csname #1\endcsname{%
  {\text{#1}}
}%
}
\newcommand{\textvars}[1]{\forcsvlist{\createtextvar}{#1}}

\usepackage[bigfiles]{pdfbase}
\ExplSyntaxOn
\NewDocumentCommand\embedvideo{smm}{
  \group_begin:
  \leavevmode
  \tl_if_exist:cTF{file_\file_mdfive_hash:n{#3}}{
    \tl_set_eq:Nc\video{file_\file_mdfive_hash:n{#3}}
  }{
    \IfFileExists{#3}{}{\GenericError{}{File~`#3'~not~found}{}{}}
    \pbs_pdfobj:nnn{}{fstream}{{}{#3}}
    \pbs_pdfobj:nnn{}{dict}{
      /Type/Filespec/F~(#3)/UF~(#3)
      /EF~<</F~\pbs_pdflastobj:>>
    }
    \tl_set:Nx\video{\pbs_pdflastobj:}
    \tl_gset_eq:cN{file_\file_mdfive_hash:n{#3}}\video
  }
  \pbs_pdfobj:nnn{}{dict}{
    /Type/RichMediaInstance/Subtype/Video
    /Asset~\video
    /Params~<</FlashVars (
      source=#3&
      skin=SkinOverAllNoFullNoCaption.swf&
      skinAutoHide=true&
      skinBackgroundColor=0x5F5F5F&
      skinBackgroundAlpha=0.75
    )>>
  }
  \pbs_pdfobj:nnn{}{dict}{
    /Type/RichMediaConfiguration/Subtype/Video
    /Instances~[\pbs_pdflastobj:]
  }
  \pbs_pdfobj:nnn{}{dict}{
    /Type/RichMediaContent
    /Assets~<<
      /Names~[(#3)~\video]
    >>
    /Configurations~[\pbs_pdflastobj:]
  }
  \tl_set:Nx\rmcontent{\pbs_pdflastobj:}
  \pbs_pdfobj:nnn{}{dict}{
    /Activation~<<
      /Condition/\IfBooleanTF{#1}{PV}{XA}
      /Presentation~<</Style/Embedded>>
    >>
    /Deactivation~<</Condition/PI>>
  }
  \hbox_set:Nn\l_tmpa_box{#2}
  \tl_set:Nx\l_box_wd_tl{\dim_use:N\box_wd:N\l_tmpa_box}
  \tl_set:Nx\l_box_ht_tl{\dim_use:N\box_ht:N\l_tmpa_box}
  \tl_set:Nx\l_box_dp_tl{\dim_use:N\box_dp:N\l_tmpa_box}
  \pbs_pdfxform:nnnnn{1}{1}{}{}{\l_tmpa_box}
  \pbs_pdfannot:nnnn{\l_box_wd_tl}{\l_box_ht_tl}{\l_box_dp_tl}{
    /Subtype/RichMedia
    /BS~<</W~0/S/S>>
    /Contents~(embedded~video~file:#3)
    /NM~(rma:#3)
    /AP~<</N~\pbs_pdflastxform:>>
    /RichMediaSettings~\pbs_pdflastobj:
    /RichMediaContent~\rmcontent
  }
  \phantom{#2}
  \group_end:
}
\ExplSyntaxOff

\usepackage{graphicx}

\newcommand{\mycomment}[1]{}

\newcommand{\calL}{{\cal L}}

\newcommand{\calO}{{\cal O}}
\newcommand{\calP}{{\cal P}}
\newcommand{\calQ}{{\cal Q}}

\newcommand{\calV}{{\cal V}}

\newcommand{\calX}{{\cal X}}

\newcommand{\bbf}{{\bf f}}  
\newcommand{\bg}{{\bf g}}

\newcommand{\bq}{{\bf q}}

\newcommand{\bs}{{\bf s}}
\newcommand{\bt}{{\bf t}}

\newcommand{\bv}{{\bf v}}

\newcommand{\bx}{{\bf x}}
\newcommand{\by}{{\bf y}}

\newcommand{\bD}{{\bf D}}

\newcommand{\bP}{{\bf P}}
\newcommand{\bQ}{{\bf Q}}
\newcommand{\bR}{{\bf R}}
\newcommand{\bS}{{\bf S}}
\newcommand{\bT}{{\bf T}}

\newcommand{\bX}{{\bf X}}

\newcommand{\IR}{{\mathds{R}}}

\newcommand{\vcomment}[1]{}

\addeditor{vincent}{VL}{0.0, 0.5, 0.0}
\addeditor{arslan}{AA}{0.0, 0.0, 0.8}
\addeditor{tom}{TR}{1.0, 0.0, 1.0}
\addeditor{nicolas}{NV}{0.0, 0.8, 0.5}
\showeditstrue
\showeditsfalse

\title{Articulation in Prime: Primitive-Based Articulated Object Understanding from a Single Casual Video}

\author{%
  Arslan Artykov, \ 
  Tom Ravaud, \ 
  Nicolás Violante-Grezzi, \ 
  Vincent Lepetit \\[1.5ex]
  LIGM, CNRS, Univ Gustave Eiffel, ENPC, Institut Polytechnique de Paris, France \\
}

\begin{document}

\maketitle

\begin{abstract}

Retrieving the 3D kinematics of articulated objects from monocular video is a fundamental challenge in computer vision. Existing methods rely on complex video setups or cues such as long-term point tracking or wide-baseline matching, but are frequently brittle under severe occlusions, rapid camera ego-motion, or weak local features. Learning-based methods, meanwhile, struggle to generalize beyond their training categories. We propose a category-agnostic optimization framework that treats articulated object understanding as a primitive-fitting problem. Geometric primitives serve as a proxy representation that avoids the pitfalls of unstable point tracks; a novel mechanism organizes them into coherent parts constrained by revolute and prismatic joints. Our formulation jointly optimizes part segmentation and joint parameters, recovering complex kinematics from a single casually captured video. A visibility-aware procedure handles partial observations and occlusions inherent to real-world data. We also propose the AiP-synth and AiP-real benchmarks, featuring significant camera motion and heavy occlusions, and outperform existing methods. Project page: \url{https://aartykov.github.io/Articulation-in-Prime/}

\end{abstract}

\section{Introduction}
\label{sec:intro}

Articulated objects are composed of rigid parts connected by joints, which constrain the parts to move according to the type of joint, e.g., prismatic and revolute. Understanding articulated objects from images is a fundamental problem in computer vision, where the goal is to segment the object into articulated parts and recover joint parameters. 

Early approaches required elaborate acquisition protocols  impractical for real-world use, such as multi-view image sets at two distinct articulation states, or a 3D object scan paired with an interaction video. We instead focus on monocular video of an object exhibiting articulated motion, which simplifies acquisition but introduces technical challenges such as camera ego-motion and object occlusions. To address these, existing approaches~\cite{liu2023building,artykov-iccvw25-articulated} rely on long-term point tracks or wide-baseline matches to estimate part motion, leading to brittle reconstructions when local features fail. Learning-based alternatives struggle to generalize to unseen object categories, even when trained on synthetic data with ground-truth annotations.

We formulate articulated object understanding as a primitive-fitting problem over an input video. Our approach leverages the global structure of primitives as a \textit{proxy representation} to guide optimization, making it more robust than tracking-based methods and generalizing across a diverse set of objects. For each frame, we backproject depth into 3D points, then optimize primitives to fit the resulting point clouds and move according to predicted joints. A novel mechanism organizes primitives into parts constrained by articulated motions, and a visibility-aware procedure handles partial observations inherent to real-world captures. We evaluate on existing benchmarks and on AiP-synth and AiP-real, two new benchmarks we introduce featuring large camera motions and heavy occlusions.

\section{Related Work}%
\textbf{Learning-based articulated object understanding.}
Early methods~\cite{mu2021asdf} extend implicit representations~\cite{park2019deepsdf} to category-level modeling by embedding joint angles in a learned manifold. CAPTRA~\cite{captra} tracks articulated motion directly from point cloud sequences, while ANCSH~\cite{li2020category} generalizes NOCS~\cite{wang2019normalized} to represent canonical configurations of articulated objects. However, unlike our approach, these methods are typically category-specific. Ota et al.~\cite{ota-hsaur} identify movable components and validates them through physical interaction. Ditto~\cite{jiang2022ditto} estimates geometry and relative state changes from multi-view 3D point clouds.

\textbf{Articulated object reconstruction from a point cloud video.}
Optimization-based methods on point cloud sequences such as ReArt~\cite{liu2023building} and Chat et al.~\cite{chao2025part} (code not available) recover a hierarchy of piecewise rigid bodies that generalizes across categories. Other methods~\cite{shi2021self} rely on complete point clouds  of the object at each timestep. To improve real-world applicability, we instead limit our input to single-view point clouds. P$^3$-Net~\cite{shi2022p} uses recurrent neural networks for part and motion estimation; other unsupervised approaches~\cite{zhong2023multi} focus only on part segmentation.

\textbf{Articulated object reconstruction from multi-view input.}
Watch-It-Move~\cite{noguchi-cvpr22-watchitmove} uses geometric cues across multiple video sequences to infer object articulation. To mitigate the requirement for dense temporal data, PARIS~\cite{liu2023paris} and DTA~\cite{weng2024neural} use only two sets of images of fixed object states. More recently, Wu et al.~\cite{wu2025reartgs} leverage 3DGS~\cite{kerbl20233d} for high-fidelity geometry reconstruction; however, their method does not address part and joint estimation. Other works~\cite{peng2025itaco, ai2026Articulation, kerr2024rsrd} estimate joint parameters by combining object scans with interaction videos. But in contrast to our approach, this requires a separate acquisition stage to obtain a scan.

\textbf{Articulated object understanding from a single video input.}
Similarly to our method, REACTO~\cite{song2024reacto}, Artykov et al.~\cite{artykov-iccvw25-articulated}, Articulate-Anything~\cite{le2024articulate}, Artipoint~\cite{werby2025articulated}, and FreeArtGS~\cite{dai2026freeartgs} (code not available) operate on single-video inputs. REACTO reconstructs surface geometry without estimating joint parameters, while Articulate-Anything estimates joint parameters for single-part objects but bypasses geometry reconstruction by retrieving meshes from a external database. Artipoint leverages hand-centric priors and persistent 2D trajectories for keypoint tracking for joint parameter extraction. 

\textbf{Object decomposition into primitives.} 
Representing an object as a set of geometric primitives~\cite{tulsiani2017learning, yang2021unsupervised, deng2020cvxnet} is a classical problem in computer vision and graphics. Optimization-based methods~\cite{liu2022robust, liu2023marching, ganeshan2025residual}, directly optimize the parameters of the primitives directly, while learning-based methods~\cite{tulsiani2017learning,paschalidou2019superquadrics, deng2020cvxnet, yang2021unsupervised, fedele2025superdec} predict them from an observation of the object. 
We choose the popular superquadrics~\cite{fedele2025superdec, liu2022robust, paschalidou2019superquadrics}, as they offer a good trade-off between expressiveness and compactness of representation.
Unlike these methods, which target accurate object reconstruction, our approach uses primitives as a \textit{proxy} for structures with geometric and dynamic coherence.

\textvars{prism,rev,type,static}
\textvars{obs,reg,rec,flow}

\section{Method}

\begin{figure}
    \centering
    \includegraphics[width=\linewidth]{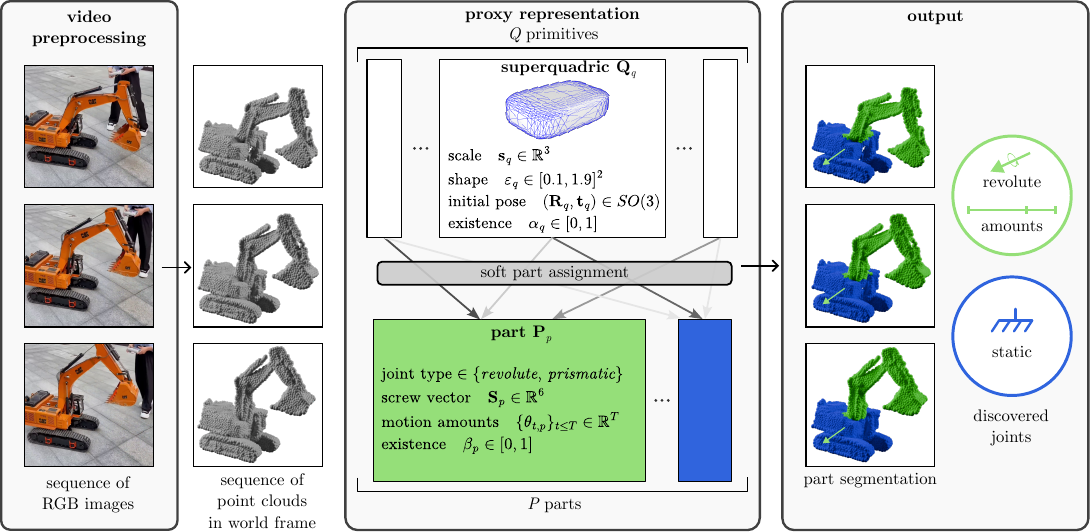}
    \vspace{-0.5cm}
    \caption{\textbf{Overview.} Given a sequence of depth maps and camera poses extracted from a video, we backproject the frames into a sequence of partial point clouds in world coordinates. We jointly optimize a set of superquadric primitives and part assignments, where each primitive is softly assigned to a part via a differentiable allocation, and each part carries its own joint parameters (revolute or prismatic) and per-timestep motion amounts. The optimization is driven by reconstruction and scene flow losses, with regularization encouraging a sparse part and primitive decomposition.}
    \label{fig:overview}
\end{figure}

The input to our method is a sequence of depth maps $\bD_{t\le T}$ of an object undergoing articulated motion and corresponding camera poses extracted from a video of $T$ frames.
For real sequences, we use ViPE~\cite{huang2025vipe} to obtain depth maps and camera poses and like previous works~\cite{artykov-iccvw25-articulated, ai2026Articulation, peng2025itaco}, we use SAM2~\cite{ravi2024sam2} to get the object masks.

We backproject the sequence of depth maps into a sequence of point clouds $\bX_t = \{\bx_{t,n}\}_{n \le N}$, where the 3D points $\bx_{t,n}$ are expressed in world coordinates. 
Point cloud $\bX_t$ contains only the points of the object seen in frame $t$. Thus, each $\bX_t$ contains only a portion of the surface of the object due to camera movement and occlusions.


\textbf{Parts and Primitives.} 
We use convex superquadrics for the primitives. A superquadric is parameterized by a 3D rotation $\bR_q$ and translation $\bt_q$, its scale $\bs_q \in \IR^3$, and $\varepsilon_q \in [0.1, 1.9]^2$, the 2-vector controlling its shape (more details are given in the appendix (Sec.~\ref{supp:superquadric})).
These primitives are assigned to parts via a soft differentiable assignment, and the motion of each primitive is a weighted aggregation of the part motions. At convergence, a primitive is assigned to only one part.
We consider a set $\calP =\{ \bP_p\}_{p \le P}$ of $P$ parts and a  
a set $\calQ = \{ \bQ_q\}_{q \le Q}$ of $Q$ primitives. 
During optimization, the probability $v_{q,p}$ for primitive $\bQ_q$ to belong to part $\bP_p$ is computed as:
\begin{equation}
  v_{q,p} = \frac{\exp(h_{q,p})}{\sum_{p'} \exp(h_{q,p'})} \> ,
  \label{eq:vqp}
\end{equation}
where $h_{q, p}$ are the entries of an assignment matrix. Because the numbers of parts and primitives are typically small, we directly optimize on the $h_{q, p}$ values.

To segment the point clouds into parts, we first introduce the probability $w_{t,n,q}$ for point $\bx_{t, n}$ to belong to primitive $\bQ_q$, which we compute as:
\begin{equation}
  w_{t,n,q} = \frac{\exp(\bbf_{t,n} . \bg_q) }{\sum_{q'} \exp(\bbf_{t,n} . \bg_{q'})} \> ,
  \label{eq:wnq}
\end{equation}
where $\bg_q$ is an optimized feature vector for primitive $\bQ_q$ and $\bbf_{t,n}$ is a feature vector for point $\bx_{t,n}$ predicted by a small MLP which we optimize together with the other parameters. This MLP takes the point's 3D coordinates as input.
Since the number of points is very large, it is more efficient to optimize on the MLP parameters and the feature vectors of the primitives instead of directly optimizing an assignment matrix.



%

\textbf{Joint Parameters.}
Each part $\bP_p$ has joint parameters. To half of the parts, we assign a prismatic joint, and a revolute joint to the remaining parts. We optimize on more parts than we expect to find on the target object. The superfluous parts will not be assigned any primitive after optimization and can thus be discarded trivially, allowing us to automatically identify the number of parts and the nature of their joint. 


We parameterize the joint parameters using twist vectors from screw theory. A twist vector $\bS_\rev$ for a revolute motion can be written as $
\bS_\rev = [ \omega, (-\omega \times \bq_\bot)]$ with $\bq_\bot = \bq - (\bq.\omega)\omega$, under the constraint that $\|\omega\| = 1$, so we optimize over 3-vectors $\omega$ and $\bq$. 
A twist vector $\bS_\prism$ for a prismatic motion can be written as $\bS_\prism = [0, \bv]$ with $\|\bv\| = 1$, and we parameterize it with $\bv$. Each part has thus a screw vector $\bS_p$, and a motion amount per timestep $\theta_{t,p}$.
We can compute the transformation matrix $\bT_{t,p}$ (3D rotation and translation) for part $p$ at time $t$ by applying Rodrigues' formula to $(\theta_{t,p} \bS_p)$. If at the end of the optimization the motion amounts for a part are small enough, then the part is considered static.

We can now express how a primitive $q$ moves by computing its screw vector as:
\begin{equation}
    \bT_{t,q} = \left(\sum_p v_{q,p} \theta_{t,p} \bS_p\right) . [\bR_q, \bt_q] \> ,
\end{equation}
by taking the linear combination of the parts' screw vector and applying it to the rotation and translation of the primitive.

\textbf{Objective Function.}
We optimize all parameters using Adam~\cite{kingma-icml15-adam} over a global objective function that balances observation and regularization terms:
\begin{equation}
\calL = \calL_\obs + \calL_\reg \>.
\end{equation}
Optimization details and weights for each loss term are given in the appendix (Sec.~\ref{appendix:optim}); we do not show the weights in the equations below for simplicity.

The observation loss $\calL_\obs$ measures how well the primitives fit the 3D points and vice-versa, and how well the primitive motions fit the scene flow between consecutive time steps:
\begin{equation}
    \calL_\obs = \calL_\rec^{\calQ\rightarrow\calX} + \calL_\rec^{\calX\rightarrow\calQ} + \calL_\flow \> .
\end{equation}

\textbf{Reconstruction Loss.}
Each superquadric is represented by points $\mathbf{y}$ sampled from its surface (please refer to the appendix (Sec.~\ref{supp:superquadric_sampling}) for sampling details). The term $\calL_\rec^{\calQ\rightarrow\calX}$ ensures that the primitives remain close to observed points. In contrast with previous work using superquadrics~\cite{paschalidou2019superquadrics, fedele2025superdec}, we use a single camera and the scene is dynamic, so we do not observe complete point clouds, only points that are visible from the camera. Simply taking $\calL_\rec^{\calQ\rightarrow\calX}$ as the Chamfer distance between the primitives and the point clouds would result in a biased estimate as many points on the superquadrics are not explained by these partial point clouds. To solve this issue, we consider only the points on the superquadrics that are visible from the camera
\begin{equation}
  \calL_\rec^{\calQ\rightarrow\calX} = 
  \frac{1}
  {\sum_{t \le T} \sum_{q \le Q} \sum_{\by \in \calV_t(\bT_{t,q}.\bQ_q)} \gamma_{t,\by,q}}
   \sum_{t \le T} \sum_{q \le Q} \sum_{\by \in \calV_t(\bT_{t,q}.\bQ_q)} \gamma_{t,\by,q} d(\by, \bX_t) \> ,
\end{equation}
where
\begin{equation}
    d(\by, \bX) = \min_{\bx \in \bX}{\|\by - \bx\|_2^2} \> ,
\end{equation}
and $\calV_t(\bT . \bQ)$ is the visible part of primitive $\bQ$ under pose $\bT$ at time $t$. 
$\gamma_{t,\by,q} \in [0, 1]$ is the probability that point $\by$ on primitive $\bQ_q$ is not occluded by any other primitive.
Inspired by NeRF/3DGS~\cite{kerbl20233d,Mildenhall2020nerf}, we compute it as
\begin{equation}
    \gamma_{t,\by,q}  = \alpha_q \prod_{q' \in \calO_t(\by)} {(1 - \alpha_{q'})} \> , 
\label{eq:inter_primitive_occlusion}
\end{equation}
where $\calO_t(\by)$ is the set of primitives that occlude point $\by$ at time $t$ and $\alpha_q$ is the probability of existence of primitive $\bQ_q$. We optimize it directly using $\calL_\text{prim-existence}$ introduced below~(Eq.\eqref{eq:L_prim-existence}).

Conversely, the term $\calL_\rec^{\calX\rightarrow\calQ}$ ensures that observed points are explained by primitives:
\begin{equation}
    \calL_\rec^{\calX\rightarrow\calQ} = 
    \frac{1}{T.N}\sum_{t \le T}\sum_{n \le N}{\sum_{q \le Q}{w_{t,n,q} d_t(\bx_{t,n}, \bT_{t,q} . \bQ_q)}} \> ,
\end{equation}
where $d_t(\bx, \bT . \bQ)$ is the distance between 3D point $\bx$ and primitive $\bQ$ under pose $\bT$ at time $t$: 
\begin{equation}
    d_t(\bx, \bT . \bQ) = \min_{\by\in\calV_t(\bT . \bQ)} \|\bx - \by\|_2^2 \> .
\end{equation}

\textbf{Flow Loss.}
The term $\calL_\flow$ ensures that the parts move according to the scene flow. Our ablation study shows it is crucial to get good results. For each 3D point $\bx_{t,n}$, we predict its location $\hat{\bx}_{t,n}$ for the next time step from the part poses $\bT_{p,t}$ over time:
\begin{equation}
  \hat{\bx}_{t+1,n} = \sum_{p\le P} u_{t,n,p} \bT_{p,t+1} \bT_{p,t}^{-1} \bx_{t,n} \> ,
  \label{eq:predicted_flow}
\end{equation}
where $u_{t,n,p}= \sum_{q \le Q} v_{q,p} w_{t,n,q}$ is the probability for point $\bx_{t,n}$ to belong to part $p$. 

$\calL_\flow$ can then be computed as:
\begin{equation}
  \calL_\flow = \frac{1}{(T-1).N} \sum_{t<T} \sum_{n\le N} 
  \| ( \hat{\bx}_{t+1,n} - \bx_{t,n}) - \bbf_{t,n} \|_1 \> ,
\end{equation}
where $\bbf_{t,n}$ is the scene flow at point $\bx_{t,n}$. In practice, we use GMSF~\cite{zhang2023gmsf} to compute the scene flow from the sequence of point clouds.

\textbf{Regularization Loss.}
Regularization loss $\calL_\reg$ mostly aims at minimizing the number of parts, the number of primitives per part, and at encouraging smooth articulated motions: 
\begin{equation}
    \calL_\reg = \calL_\text{parts} +  \calL_\text{prims} + \calL_\text{motion} + \calL_\text{match}\> .
\end{equation}
\textbf{Term} $\boldsymbol{\calL}_\text{parts} = \calL_\text{part-sparsity} + \calL_\text{part-existence}$:
\begin{equation}
    \calL_\text{part-sparsity} = \left( \frac{1}{P} \sum_{p \leq P} \sqrt{\frac{1}{Q} \sum_{q \leq Q} v_{q,p} } \right)^2 \> ,
\end{equation}
and, like previous works did to encourage   primitives~\cite{yang2021unsupervised, fedele2025superdec} to "exist" or not, we use a binary cross-entropy loss:
\begin{equation}
    \calL_\text{part-existence} = -\frac{1}{P}\sum_{p \le P}{\left[\hat{\beta}_p\log{\beta_p} + (1 - \hat{\beta}_p)\log{(1 - \beta_p)}\right]} \> .
\end{equation}
where we optimize directly on the $\beta_p$, the probabilities for parts of existing and $\hat{\beta}_p = 1_{\frac{1}{Q}\sum_{q \le Q}{v_{q, p}} > \tau_{\beta}}$.
At the end of optimization, we can check if a part $p$ actually exists by similarly thresholding $\beta_p$.

\textbf{Term} $\boldsymbol{\calL}_\text{prims} = \calL_\text{prim-sparsity} + \calL_\text{prim-existence}$ plays a similar role for primitives:
\begin{equation}
    \calL_\text{prim-sparsity} = \left(
    \frac{1}{Q}
    \sum_{q \leq Q}
    \sqrt{\frac{1}{T.N} \sum_{t \leq T} 
     \sum_{n \leq N} w_{t, n, q} } \right)^2 
\end{equation}
limits the number of primitives per part, and
\begin{equation}
    \calL_\text{prim-existence} = -\frac{1}{Q}\sum_{q \le Q} \left[\hat{\alpha}_q\log{\alpha_q} + (1 - \hat{\alpha}_q)\log{(1 - \alpha_q}\right]
    \label{eq:L_prim-existence}
\end{equation}
encourages the primitives to "exist" or not. We optimize on the $\alpha_q$, the probabilities for primitives of existing, with  $\hat{\alpha}_q=1_{\sum_{t\le T}\sum_{n\le N}w_{t,n,q} > \tau_\alpha}$.
At the end of optimization, we can check if a primitive $q$ actually exists by similarly thresholding $\alpha_q$ by $\tau_\alpha$.

\textbf{Term} $\boldsymbol{\calL}_\text{motion}$ limits the magnitudes of the linear or angular velocities of the parts:
\begin{equation}
  \calL_\text{motion} = \sum_{t < T} \sum_{p \leq P} (\theta_{t+1,p} -\theta_{t,p})^2 \> . 
\end{equation}

Finally, \textbf{term} $\boldsymbol{\calL}_\text{match}$ encourages primitives to be matched to only one part since minimizing the entropy of the $v_{q,p}$'s encourages them to tend to a one-hot vector:
\begin{equation}
    \calL_\text{match} = -\frac{1}{Q} \sum_{q \le Q} \sum_{p \le P} v_{q,p} \log v_{q,p}  \> .
\end{equation}

\section{Experiments}

\subsection{Baseline Methods} 

Existing approaches in articulated object modeling typically rely on multi-view setups or 3D scans paired with aligned interaction sequences, which complicates the identification of directly comparable baselines. To evaluate our approach, we compare it against the following methods.

\textbf{Video2Articulation}~\cite{peng2025itaco} is an optimization-based approach that inputs a 3D object scan along with a scan-aligned interaction video and segments the object’s dynamic part and estimates joint parameters.
\textbf{Articulate-Anything}~\cite{le2024articulate} inputs a single video, but it assumes access to a mesh database from which object parts are retrieved, and it does not predict part labels or per-timestep motion values, so we denote these entries as “N/A” in our tables. 
\textbf{ReArt}~\cite{liu2023building} is the most comparable to our approach, as it takes a dynamic point cloud sequence as input. It segments object parts in a canonical frame and propagates them to other timesteps using optimized joint parameters. 
\textbf{Artipoint}~\cite{werby2025articulated}
leverages human-object interaction to identify affordance points and uses long-term point tracking to follow moving parts. Similar to Articulate-Anything, it does not estimate part labels or motion states. Since Artipoint requires human-object interaction, we restrict its evaluation to real-world sequences.

\subsection{Datasets} 

\textbf{Video2Articulation-S.} We evaluate on the simulated dataset introduced by Video2Articulation~\cite{peng2025itaco}, which consists of 73 videos spanning 11 categories from the PartNet-Mobility dataset~\cite{mo2019partnet}. 

\textbf{Arti4D.} We evaluate on the Arti4D dataset introduced by Artipoint~\cite{werby2025articulated}, which consists of object interaction sequences captured in real-world settings. The dataset comprises four scenes characterized by repetitive object instances. Consequently, we selected 11 representative objects from across the full set of scenes for our evaluation.

\textbf{AiP-synth.} We introduce a novel dataset comprising 16 challenging categories. We curated a diverse selection of objects—including bottles, cameras, and globes—specifically to evaluate the robustness of our framework across a wide spectrum of kinematic and articulation configurations.  We utilize the SAPIEN~\cite{xiang2020sapien} environment to render RGB-D sequences of objects sourced from the PartNet-Mobility~\cite{mo2019partnet} dataset. To ensure diversity in perspective, our sequences incorporate two distinct camera trajectories: ping-pong and circular arc. The figures of the representative objects from the dataset and representative figures of camera trajectories are provided in the appendix (Sec.~\ref{supp:aip_synth}).

\textbf{AiP-real.} To assess performance in unconstrained real-world environments, we curated a real-world dataset. To demonstrate the robustness and accuracy of our framework, we supplemented our own captures with diverse sequences sourced from YouTube, including challenging objects such as globes and excavators.

\subsection{Metrics}

We evaluate object part segmentation using the mean Intersection-over-Union~(mIoU) metric. For joint axis estimation, we report the sign-agnostic axis–angle error (in degrees), and for pivot localization, we use the point-to-line distance (in centimeters). We further evaluate the motion state using the mean absolute error (L1) between the predicted and ground-truth motion magnitudes, reported in degrees or centimeters as appropriate. Finally, we report joint type accuracy as the percentage of correctly predicted joint types.

\begin{table*}[t]
    \centering
    \caption{\textbf{Quantitative results on the AiP-synth dataset.} `N/A' means the method does not predict the metric. `F' means the method fails on the category. Best results are in \textbf{bold}.}
    \label{tab:quant_results}
    \renewcommand{\tabcolsep}{6pt}
    \resizebox{\textwidth}{!}{
    \begin{tabular}{llccccc}
    \toprule
    \textbf{Joint Type} & \textbf{Method} 
    & \textbf{mIoU $\uparrow$}
    & \textbf{Axis $\downarrow$ (°)} 
    & \textbf{Position $\downarrow$ (cm)} 
    & \textbf{State $\downarrow$ (° / cm)} 
    & \textbf{Type Acc. $\uparrow$ (\%)} \\
    \midrule
    \multirow{4}[2]{*}{Revolute} 
    
    & Articulate-Anything~\cite{le2024articulate} 
     & N/A & 69.82 $\pm$ 37.19 & 70.66 $\pm$ 53.23 & N/A & 33.3 \\

    & Video2Articulation~\cite{peng2025itaco} 
      &0.48 $\pm$ 0.31 &51.20 $\pm$ 34.99 &80.95 $\pm$ 34.20 &57.77 $\pm$ 30.02 &25.0  \\

    & ReArt~\cite{liu2023building} 
    &0.17 $\pm$ 0.31  &74.07 $\pm$ 30.22 &80.71 $\pm$ 35.22 &80.46 $\pm$ 24.18 &25.0 \\

    \rowcolor[rgb]{ .93, .93, .93} \cellcolor[rgb]{1, 1, 1}
    & Ours 
    &\textbf{0.85 $\pm$ 0.20}  &\textbf{0.52 $\pm$ 1.38} &\textbf{0.58 $\pm$ 1.45} &\textbf{1.45 $\pm$ 2.86} &\textbf{100.0} \\

    \midrule
    \multirow{4}[2]{*}{Prismatic} 
    
    & Articulate-Anything~\cite{le2024articulate} 
   & N/A & 45.58 $\pm$ 44.42 & -- & N/A & 50.0 \\

    & Video2Articulation~\cite{peng2025itaco} 
    &0.59 $\pm$ 0.21   &51.46 $\pm$ 33.06 &-- &18.63 $\pm$ 7.24 &75.0 \\

    & ReArt~\cite{liu2023building} 
    &F &F &-- &F &F \\

    \rowcolor[rgb]{ .93, .93, .93} \cellcolor[rgb]{1, 1, 1}
    & Ours 
    &\textbf{0.94 $\pm$ 0.04}  &\textbf{0.00 $\pm$ 0.00} &-- &\textbf{0.14 $\pm$ 0.23} &\textbf{100.0} \\
    \bottomrule
    \end{tabular}
    }
\end{table*}
\sisetup{detect-all=true}
\setlength{\tabcolsep}{5pt}
\begin{table*}[t]
\centering
\def\mywidth{0.65\columnwidth}
\caption{
\textbf{Quantitative results on the AiP-real dataset.} `N/A' means the method does not predict the metric. `F' means the method fails on the category.  Best results in \textbf{bold}. 
}
\resizebox{\mywidth}{!}{
\sisetup{table-auto-round,table-format=.2,table-column-width=1.6cm}

\begin{tabular}{@{}cc|cccccc}

\toprule

\multicolumn{1}{c}{\textbf{Metrics}} &
\multicolumn{1}{c|}{\textbf{Method}} &      
\multicolumn{1}{c}{Box} & 
\multicolumn{1}{c}{Chair} & 
\multicolumn{1}{c}{Sliding Box} & 
\multicolumn{1}{c}{Globe} & 
\multicolumn{1}{c}{Excavator} &

\multicolumn{1}{c}{Mean} \\

\midrule

\multirow{5}[2]{*}{mIoU $\uparrow$} 
                               
& Articulate-Anything & N/A & N/A & N/A & N/A & N/A & N/A \\
                               
& Video2Articulation &0.49 &0.02 &\textbf{0.85} &0.00 &0.55 &0.38 \\

& ReArt & F &0.35 &F &F  &F & 0.35  \\

& Artipoint~\cite{werby2025articulated} & N/A  &N/A  &N/A  &N/A &N/A &N/A  \\

\rowcolor[rgb]{ .93, .93, .93} \cellcolor[rgb]{1, 1, 1}                     
& Ours &\textbf{0.91} &\textbf{0.91} &0.71 &\textbf{0.86} &\textbf{0.89} & \textbf{0.86}\\

\midrule

\multirow{5}[2]{*}{\shortstack{Axis} (°) $\downarrow$} 

& Articulate-Anything & 90.00 & F & 90.00 &  F & F &   90.00 \\

& Video2Articulation &89.02 &10.68 &81.84 &15.21 &\textbf{10.17} &41.38 \\

& ReArt &F &81.25 &F &F &F &81.25 \\

& Artipoint~\cite{werby2025articulated} &\textbf{ 7.54}  &86.09 &35.90  &F &34.38 &40.98  \\

\rowcolor[rgb]{ .93, .93, .93} \cellcolor[rgb]{1, 1, 1}
& Ours &10.18 &\textbf{3.98} &\textbf{6.89} &\textbf{1.13} &10.49 &\textbf{6.53} \\

\midrule

\multirow{5}[2]{*}{\shortstack{Position} (cm) $\downarrow$} 

& Articulate-Anything & 28.45 & F &-- &  F & F &  28.45  \\

& Video2Articulation &135.69 &30.16 &-- &5.20 &5.09 &44.04 \\

& ReArt &F &100.0 &-- &F &F &100.0 \\

& Artipoint~\cite{werby2025articulated} & 4.07  &63.86  &-- &F &6.94 &24.96  \\

\rowcolor[rgb]{ .93, .93, .93} \cellcolor[rgb]{1, 1, 1}
& Ours &\textbf{1.35} &\textbf{6.47} &-- &\textbf{3.01} &\textbf{2.23} &\textbf{3.27} \\

\midrule

\multirow{5}[2]{*}{\shortstack{Type\\Accuracy} ($\%$) $\uparrow$}

& Articulate-Anything & \textbf{100.0} & F & 0.00 & F & F & 50.0\\
 
& Video2Articulation &\textbf{100.0} &\textbf{100.0} &0.00 &\textbf{100.0} &\textbf{100.0} &80.0 \\

& ReArt &F &0.00 &F &F &F &0.00 \\

& Artipoint~\cite{werby2025articulated} & \textbf{100.0}  &\textbf{100.0}  &0.00  &F &\textbf{100.0} &75.0  \\

\rowcolor[rgb]{ .93, .93, .93} \cellcolor[rgb]{1, 1, 1}
& Ours &\textbf{100.0} &\textbf{100.0} &\textbf{100.0} &\textbf{100.0} &\textbf{100.0} &\textbf{100.0} \\

\bottomrule
\end{tabular}%
}
\label{tab:quant_table_real}
\end{table*}

\begin{table*}[t]
    \centering
    \caption{\textbf{Quantitative results on the Video2Articulation-S dataset.} `N/A' means the method does not predict the metric. `F' means the method fails on the category. Best results are in \textbf{bold}.}
    \label{tab:v2a_dataset_quant_results}
    \renewcommand{\tabcolsep}{6pt}
    \resizebox{\textwidth}{!}{
    \begin{tabular}{llccccc}
    \toprule
    \textbf{Joint Type} & \textbf{Method} 
    & \textbf{mIoU $\uparrow$}
    & \textbf{Axis $\downarrow$ (°)} 
    & \textbf{Position $\downarrow$ (cm)} 
    & \textbf{State $\downarrow$ (° / cm)} 
    & \textbf{Type Acc. $\uparrow$ (\%)} \\
    
    \midrule
    \multirow{4}[2]{*}{Revolute} 
    & Articulate-Anything~\cite{le2024articulate} 
    & N/A & 51.43 $\pm$ 44.54 & 33.13 $\pm$ 32.01 & N/A & 85.7 \\
    
    & Video2Articulation~\cite{peng2025itaco} 
    & 0.73 $\pm$ 0.22 & 9.72 $\pm$ 25.13 & 8.41 $\pm$ 5.84 & 8.65 $\pm$ 15.32 & 88.9  \\

    & ReArt~\cite{liu2023building} 
    &0.36 $\pm$ 0.36  &54.58 $\pm$ 34.84 &66.06 $\pm$ 35.20 &75.71 $\pm$ 21.92 &55.5  \\
    
    \rowcolor[rgb]{ .93, .93, .93} \cellcolor[rgb]{1, 1, 1}
    & Ours 
    &\textbf{0.93 $\pm$ 0.05} &\textbf{0.00 $\pm$ 0.00} &\textbf{0.00 $\pm$ 0.00}&\textbf{0.14 $\pm$ 0.27} &\textbf{100.0}   \\

    \midrule
    \multirow{4}[2]{*}{Prismatic} 
    & Articulate-Anything~\cite{le2024articulate} 
    & N/A & 90.00 $\pm$ 0.00 & -- & N/A & 0 \\
    
    & Video2Articulation~\cite{peng2025itaco} 
    & 0.58 $\pm$ 0.34 & 6.83 $\pm$ 0.40 & -- & 4.06 $\pm$ 0.44 & \textbf{100.0} \\


    & ReArt~\cite{liu2023building} 
    &F  &F &-- &F &F  \\
    
    \rowcolor[rgb]{ .93, .93, .93} \cellcolor[rgb]{1, 1, 1}
    & Ours 
    &\textbf{0.88 $\pm$ 0.02}  &\textbf{ 0.00 $\pm$ 0.00} &-- &\textbf{3.00 $\pm$ 0.50} &\textbf{100.0}  \\
    \bottomrule
    \end{tabular}
    }
\end{table*}
\begin{table*}[t]
    \centering
    \caption{\textbf{Quantitative results on the Arti4D dataset.} Best results are in \textbf{bold}.}
    \label{tab:arti4d_quant_results}
    \renewcommand{\tabcolsep}{10pt}
    {\fontsize{7.2}{9}\selectfont
    \begin{tabular}{llccc}
    \toprule
    \textbf{Joint Type} & \textbf{Method} 
    & \textbf{Axis $\downarrow$ (°)} 
    & \textbf{Position $\downarrow$ (cm)} 
    & \textbf{Type Acc. $\uparrow$ (\%)} \\
    \midrule

    & Artipoint~\cite{werby2025articulated} & 24.96 $\pm$ 35.27  & 85.49 $\pm$ 76.82 & 75.0 \\
    \rowcolor[rgb]{.93, .93, .93} 
    \cellcolor[rgb]{1, 1, 1}
    \multirow{-2}{*}{Revolute} & 
    Ours & \textbf{4.77 $\pm$ 3.79} & \textbf{5.44 $\pm$ 2.00} & \textbf{100.0} \\
    
    \midrule

    & Artipoint~\cite{werby2025articulated} &28.84 $\pm$ 11.09   & -- & 64.4 \\
    
    \rowcolor[rgb]{.93, .93, .93}
    \cellcolor[rgb]{1, 1, 1}
    \multirow{-2}{*}{Prismatic} & 
    Ours &\textbf{ 3.60 $\pm$ 1.19} & -- &\textbf{100.0} \\
    
    \bottomrule
    \end{tabular}
    }
\end{table*}

\def\qualitWidth{0.20\linewidth}

\setlength{\tabcolsep}{0pt}


\begin{figure*}[t]
  \centering

\resizebox{0.75\linewidth}{!}{
\begin{tabular}{c@{$\;$}c@{$\;\;$}cccc}

\rotatebox{90}{\hspace{0.1cm} Rep.} &
\rotatebox{90}{\hspace{0.1cm} Frame} &
\includegraphics[trim={2cm 1cm 1cm 1.1cm},clip,width=\qualitWidth]{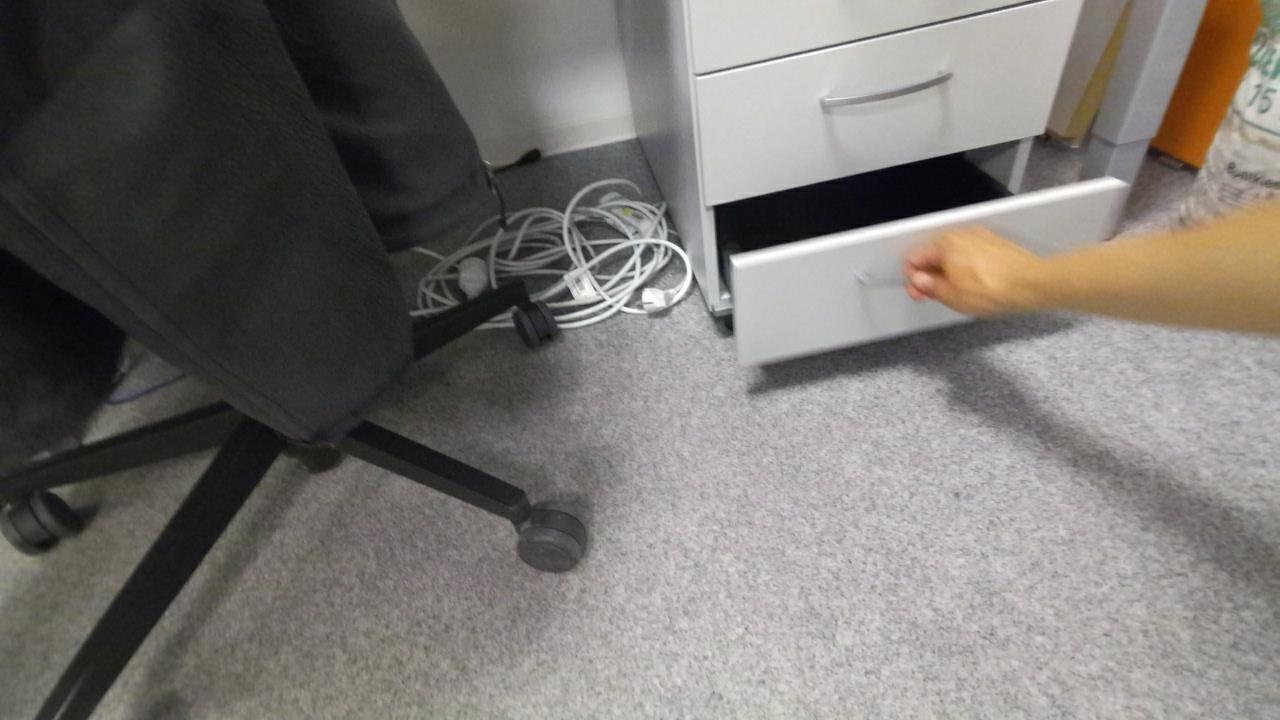} &
\includegraphics[trim={2cm 1cm 1cm 1cm},clip,width=\qualitWidth]{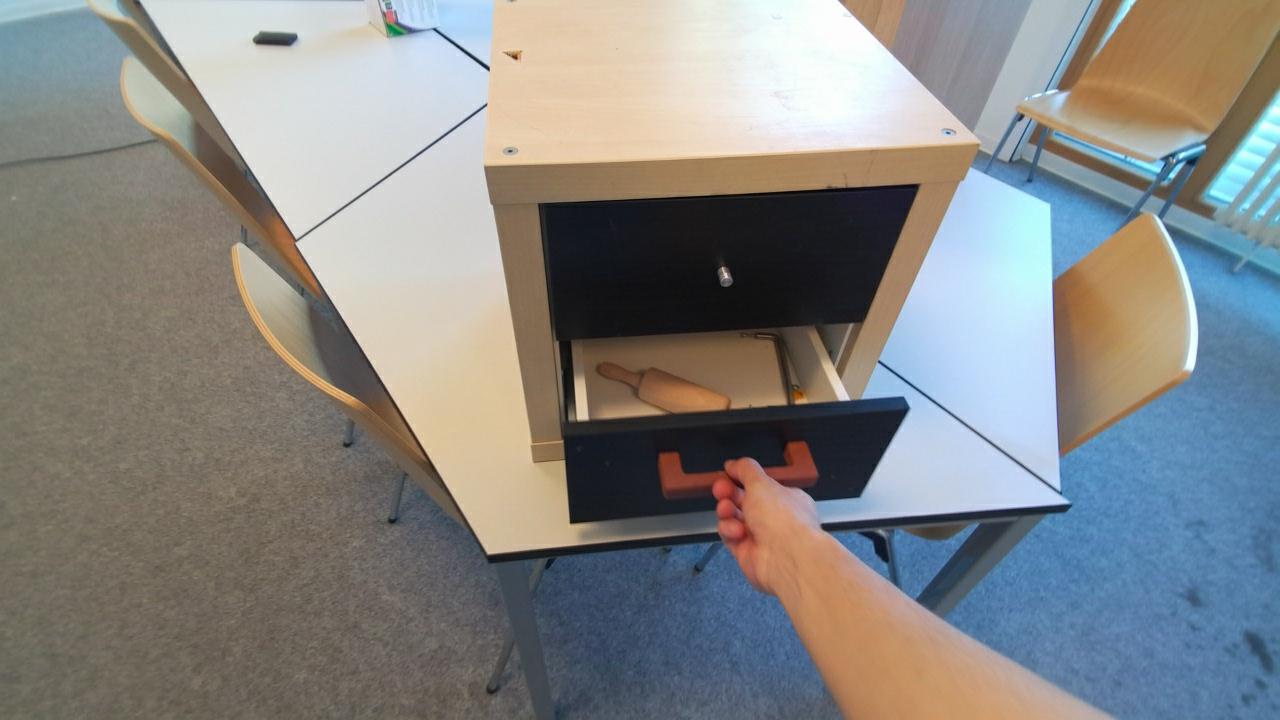} &
\includegraphics[trim={2cm 1cm 1cm 1.1cm},clip,width=\qualitWidth]{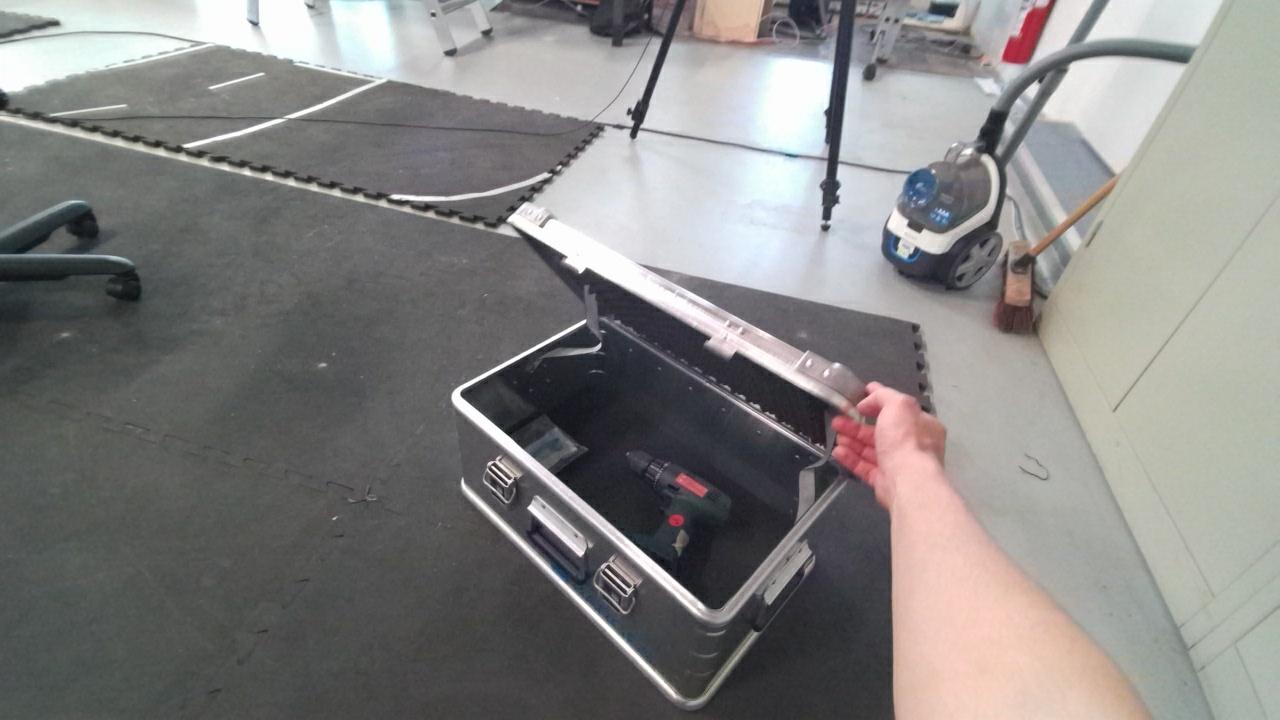} &
\includegraphics[trim={2cm 1cm 1cm 1.1cm},clip,width=\qualitWidth]{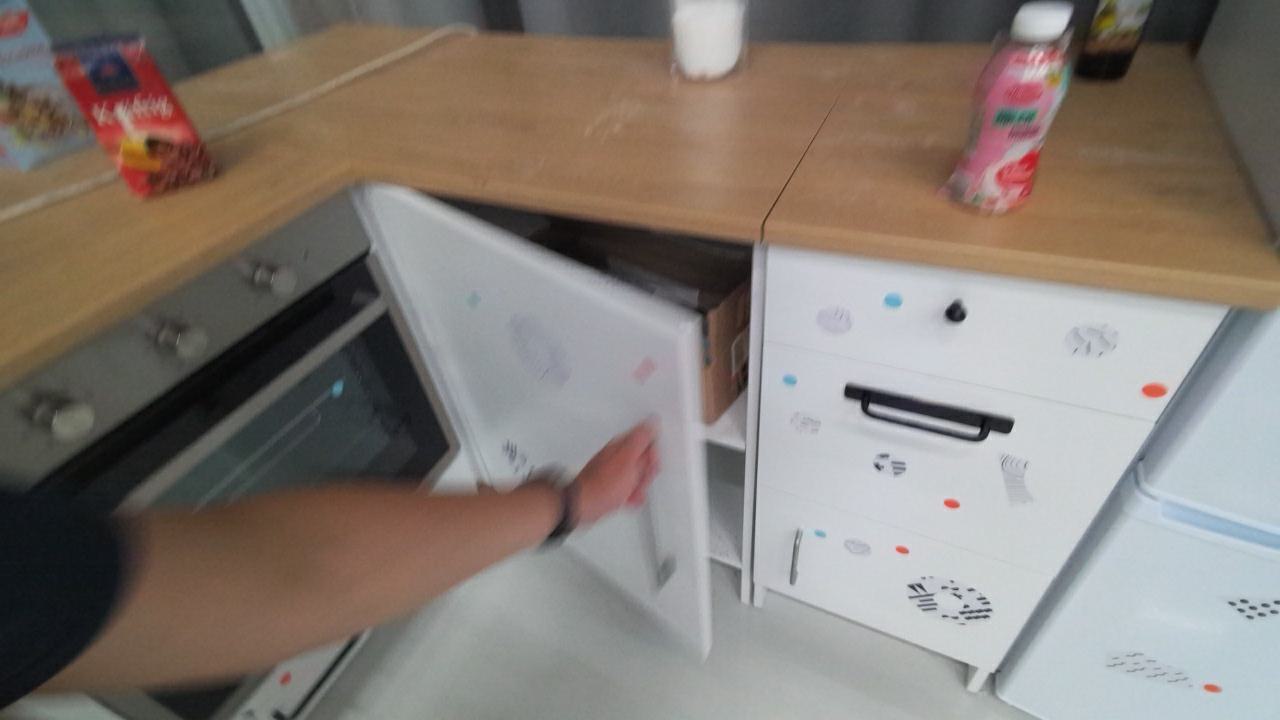} 

\\[-0.0cm]

\rotatebox{90}{\hspace{0.4cm} Results} &
\rotatebox{90}{\hspace{0.8cm}} &
\includegraphics[trim={2cm 1cm 1cm 1.1cm},clip,width=\qualitWidth]{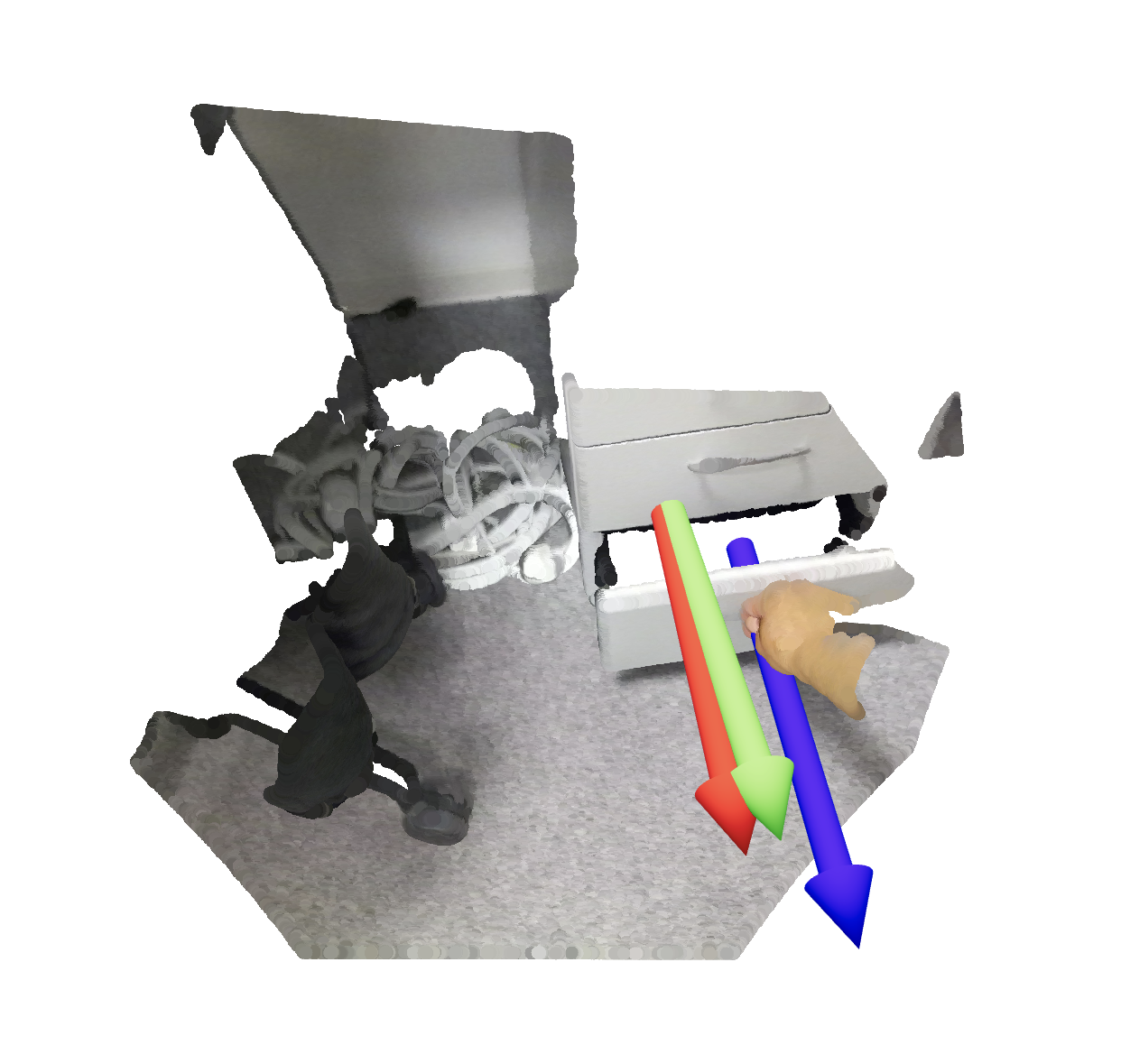} &
\includegraphics[trim={2cm 1cm 1cm 1cm},clip,width=\qualitWidth]{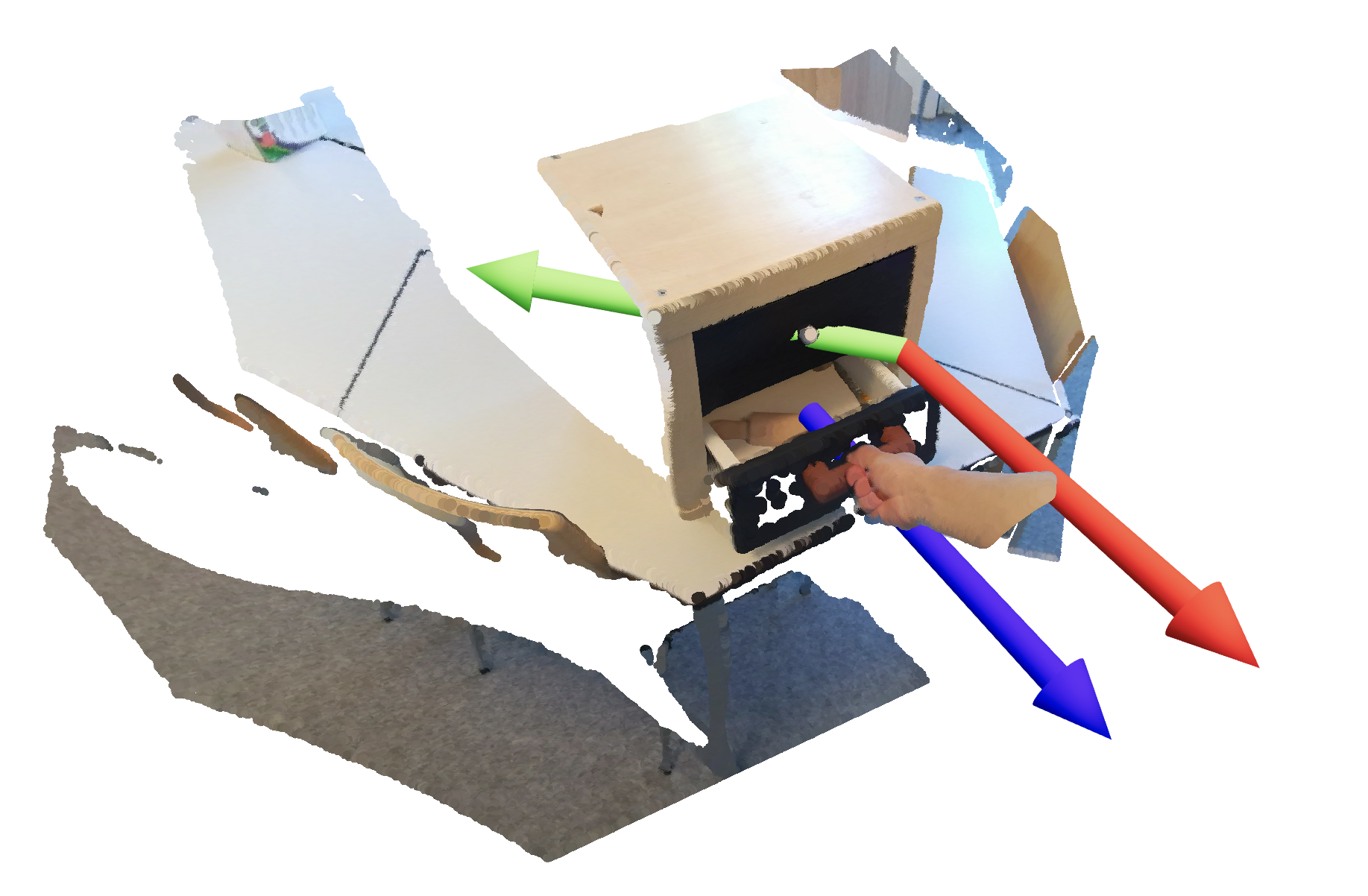} &
\includegraphics[trim={2cm 1cm 1cm 1.1cm},clip,width=\qualitWidth]{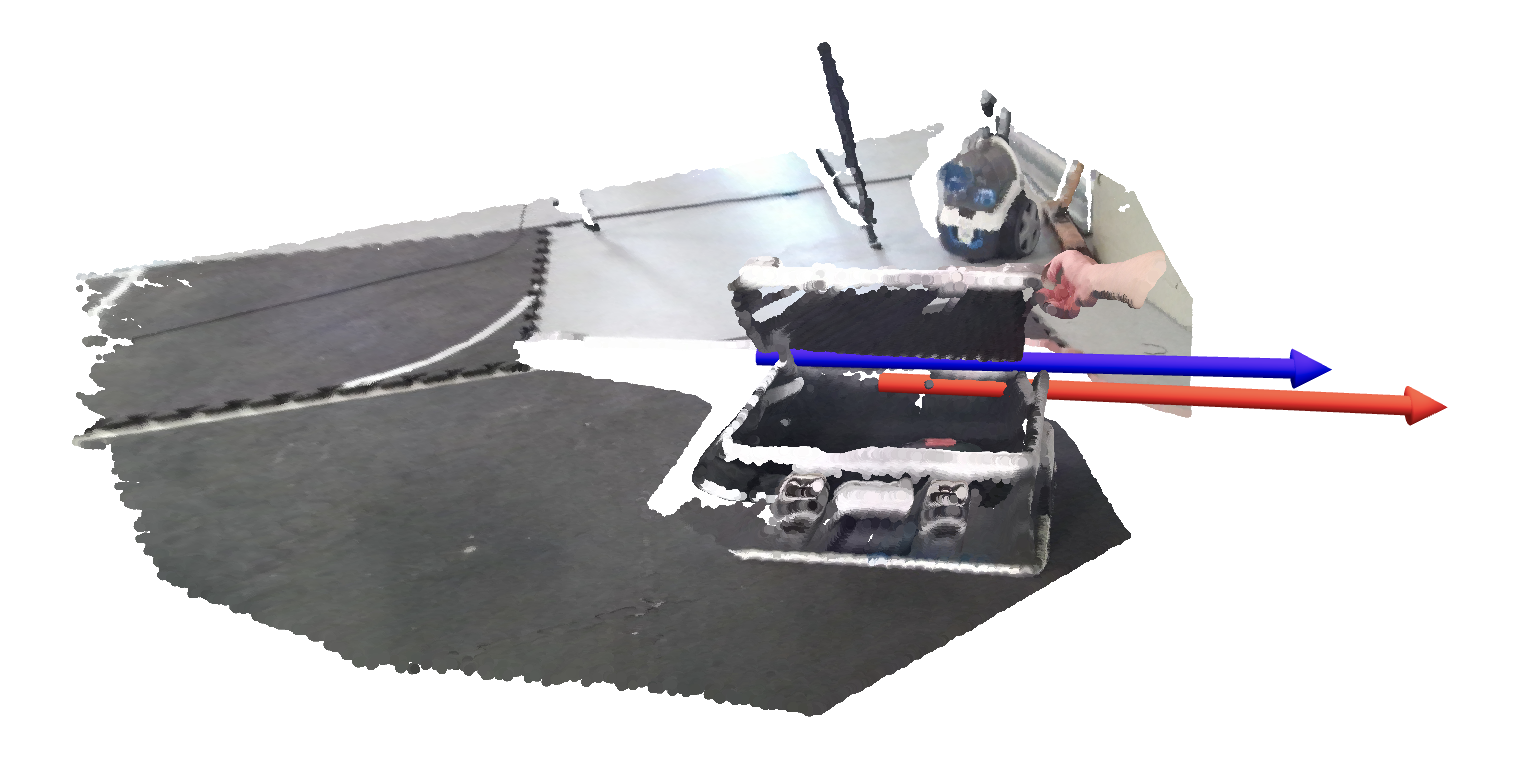} &
\includegraphics[trim={2cm 1cm 1cm 1.1cm},clip,width=\qualitWidth]{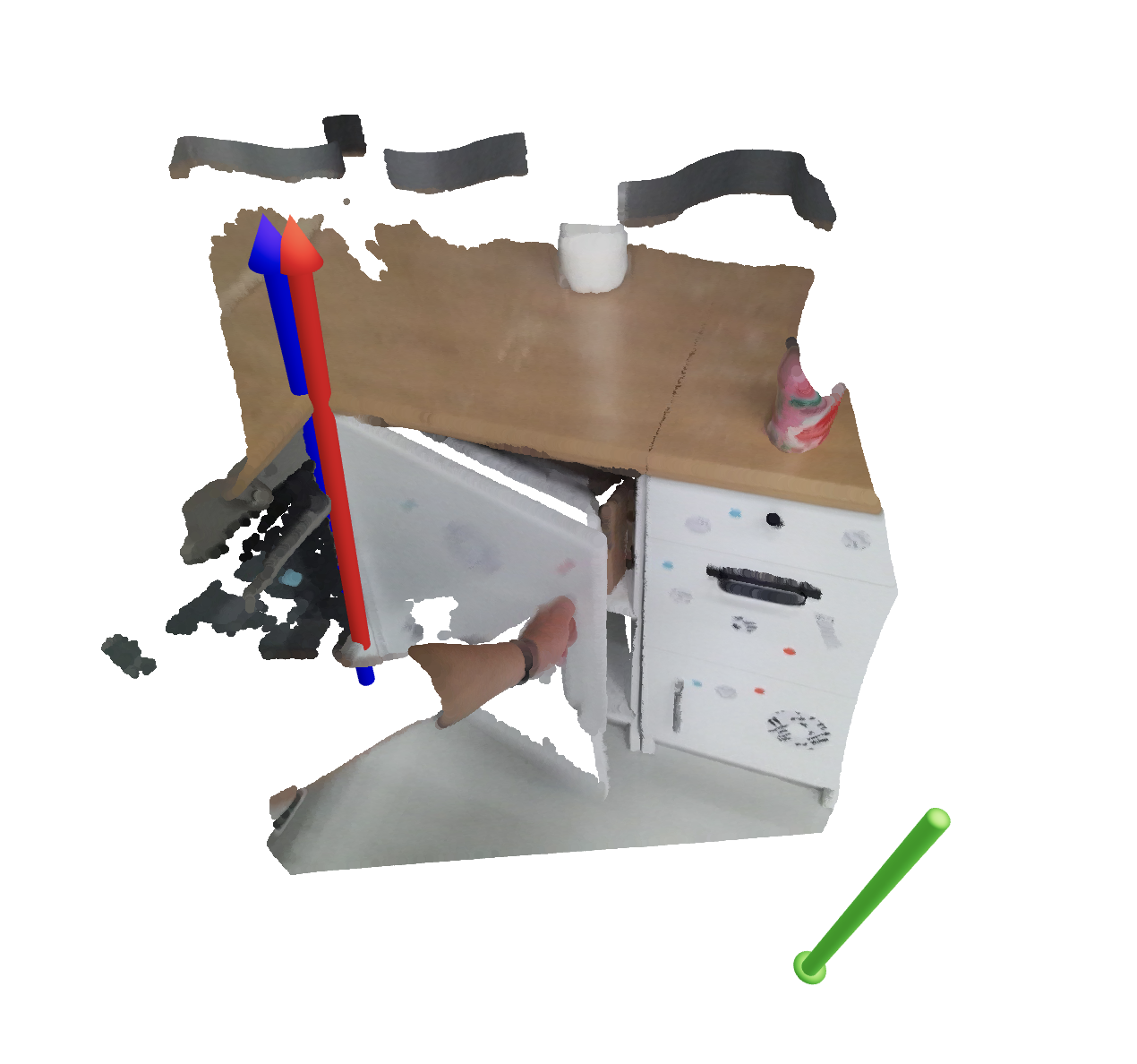} 

\\[-0.0cm]

& & RR080 & DR080 & RH078 & RH201  \\

\end{tabular}
} 
   \caption{\textbf{Qualitative results on the Arti4D dataset.} Red, green, and blue arrows denote Ours, Artipoint, and Ground-Truth joint axes, respectively. For prismatic (translational) joints, only the axis direction is evaluated, while the axis location is not considered. Because Artipoint produces a large error in axis location for scene RH078, the green joint falls outside the camera frustum.}
   \label{fig:qual_fig_arti4d}
\end{figure*}

\subsection{Results}

As illustrated in Tables~\ref{tab:quant_results} and \ref{tab:quant_table_real} and Figures~\ref{fig:qual_fig_synth} and \ref{fig:qual_fig_real}, our method significantly outperforms competing baselines on both the AiP-synth and AiP-real datasets. \arslan{Our framework achieves substantial performance improvements despite utilizing only a single casual video as input. On synthetic sequences, which provide perfect depth maps, scene flow, and camera poses, the proposed method actually attains perfect accuracy across several evaluation metrics.}
In contrast, existing baseline methods exhibit a marked decline in performance, underscoring the inherent difficulty of modeling articulated objects from monocular video sequences characterized by significant occlusions.

Tables~\ref{tab:v2a_dataset_quant_results}, \ref{tab:arti4d_quant_results} and Figure~\ref{fig:qual_fig_arti4d} further demonstrate the robustness and accuracy of our method on challenging benchmarks established by prior work. In both the Video2Articulation-S and Arti4D datasets, the camera motion is primarily quasi-static, rendering them less demanding than our proposed sequences. Consequently, while baseline methods exhibit improved performance on these benchmarks, our approach consistently maintains superior results. Notably, the Arti4D dataset contains significant artifacts and "holes" in the depth maps; despite these imperfections, our method remains robust to low-quality depth inputs.

We provide visualizations of the proxy superquadrics in the appendix (Sec.~\ref{supp:viz_superquadric}).


\def\qualitWidth{0.2\linewidth}

\newcommand{\epicfail}{
\begin{tikzpicture}[x=\qualitWidth,y=\qualitWidth]
    \useasboundingbox (0,0) rectangle (1,1); 
    \draw[gray, line width=0.1cm] (0.3,0.3) -- (0.7,0.7);
    \draw[gray, line width=0.1cm] (0.3,0.7) -- (0.7,0.3);
\end{tikzpicture}
}

\setlength{\tabcolsep}{0pt}

\vspace{-5pt}

\begin{figure*}[t]
  \centering

\resizebox{0.75\linewidth}{!}{
\begin{tabular}{c@{$\;$}c@{$\;\;$}cccccc}

\rotatebox{90}{\hspace{0.7cm}\vphantom{A}Ground} &
\rotatebox{90}{\hspace{0.8cm}\vphantom{A}Truth} &
\includegraphics[trim={3cm 2cm 3cm 1cm},clip,width=\qualitWidth]{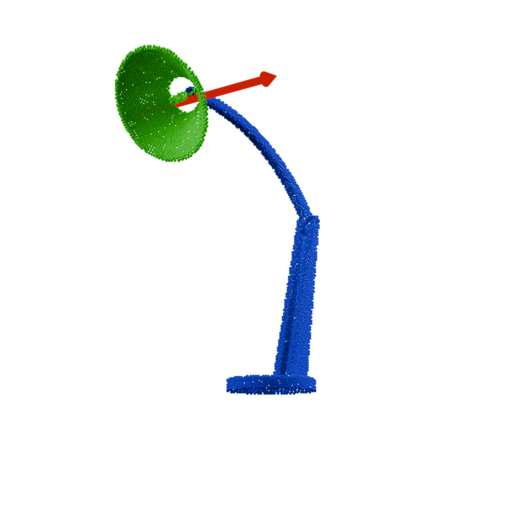} &
\includegraphics[trim={3cm 2cm 3cm 1cm},clip,width=\qualitWidth]{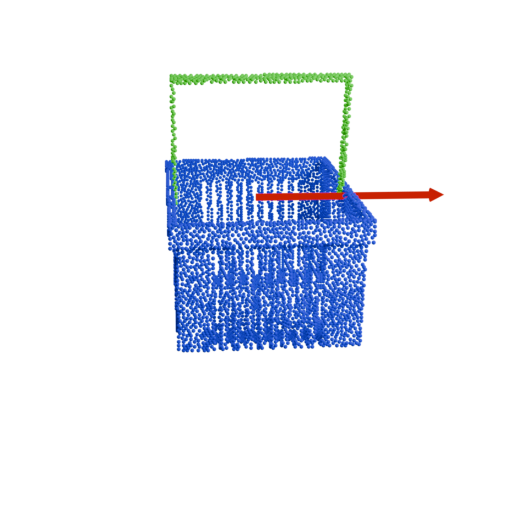} &
\includegraphics[trim={3cm 3cm 3cm 3cm},clip,width=\qualitWidth]{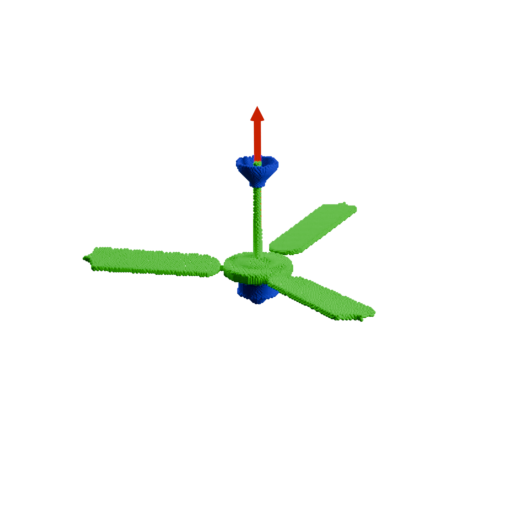} &
\includegraphics[trim={1cm 3cm 1cm 3cm},clip,width=\qualitWidth]{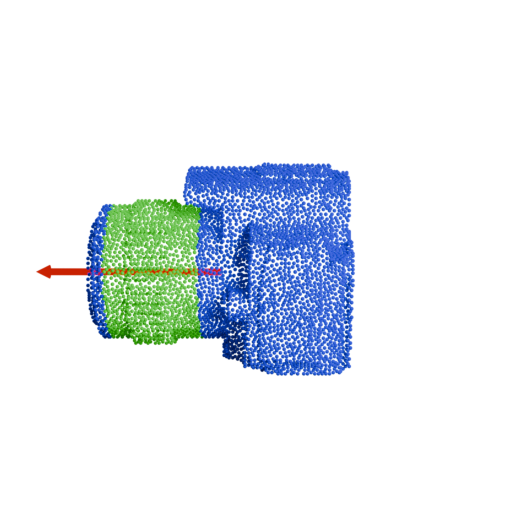} &
\includegraphics[trim={2cm 3cm 3cm 3cm},clip,width=\qualitWidth]{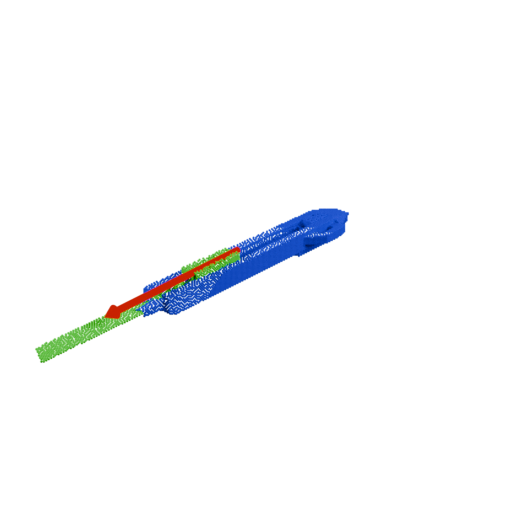}  &
\includegraphics[trim={3cm 3cm 3cm 2cm},clip,width=\qualitWidth]{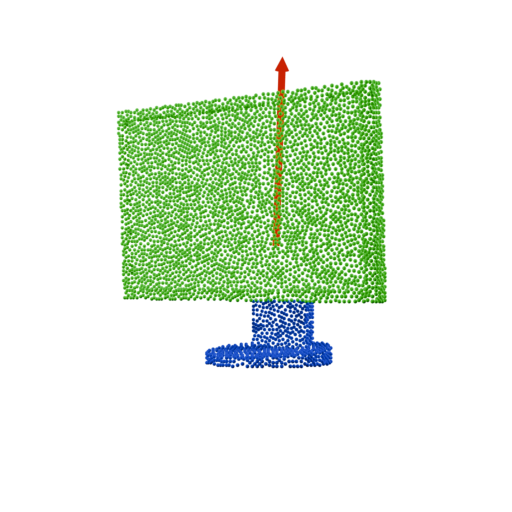} \\[-0.0cm]

\rotatebox{90}{\hspace{0.7cm} Ours} &
\rotatebox{90}{\hspace{0.8cm}} &
\includegraphics[trim={3cm 1cm 3cm 1cm},clip,width=\qualitWidth]{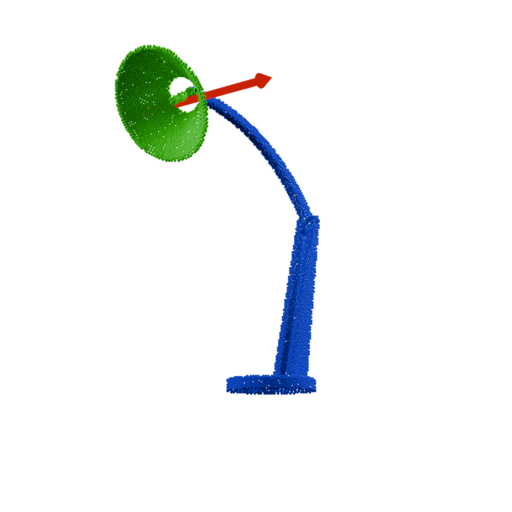}  &
\includegraphics[trim={3cm 1cm 3cm 1cm},clip,width=\qualitWidth]{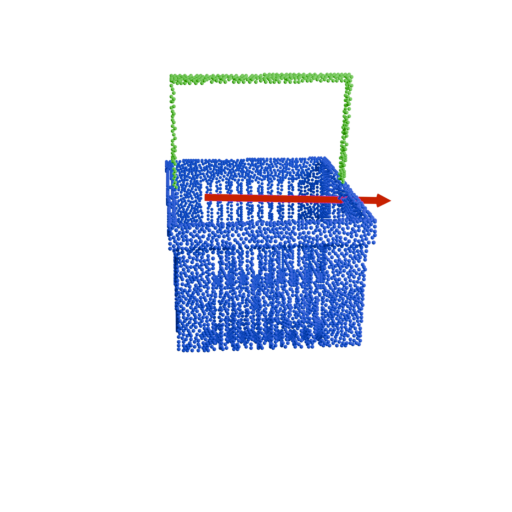}  &
\includegraphics[trim={3cm 1cm 3cm 1cm},clip,width=\qualitWidth]{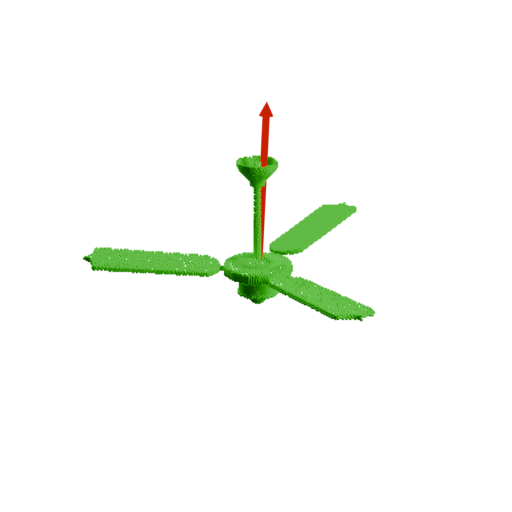}  &
\includegraphics[trim={1cm 3cm 1cm 3cm},clip,width=\qualitWidth]{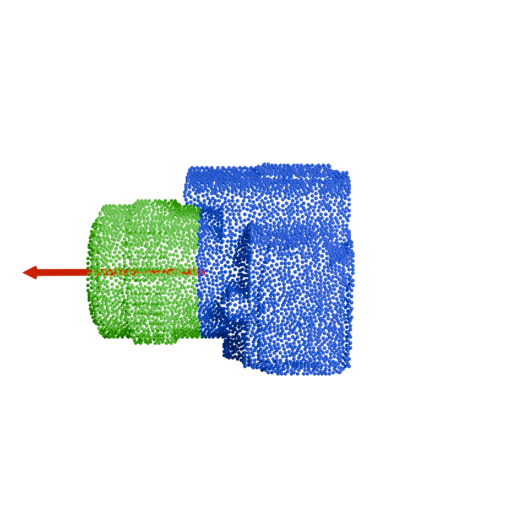}  &
\includegraphics[trim={2cm 3cm 2cm 3cm},clip,width=\qualitWidth]{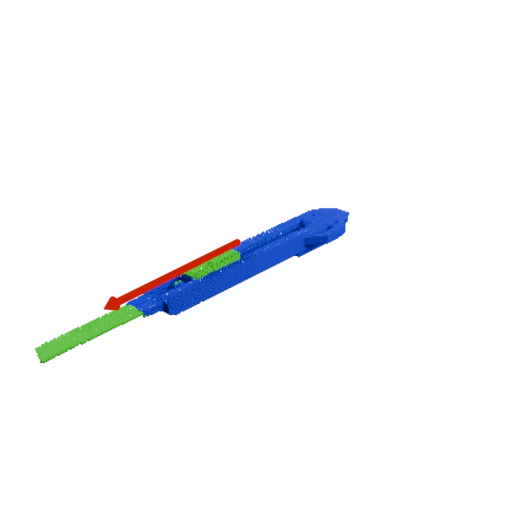}  &
\includegraphics[trim={3cm 1cm 3cm 1cm},clip,width=\qualitWidth]{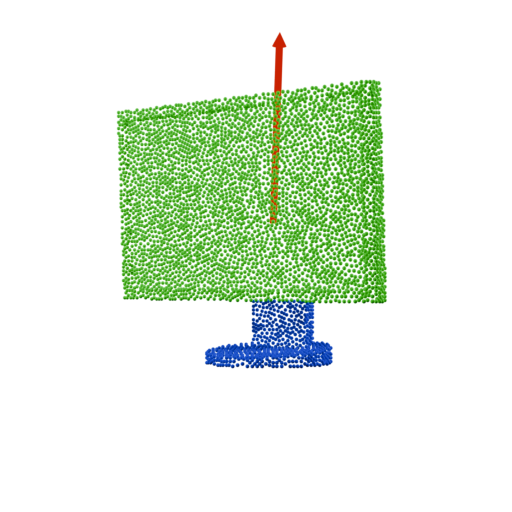}  
\\[-0.0cm]

\rotatebox{90}{\;\;\;\;\;\;\;Articulate-} &
\rotatebox{90}{\;\;\;\;\;\;\;Anything} &
\epicfail&
\epicfail &
\includegraphics[width=\qualitWidth]{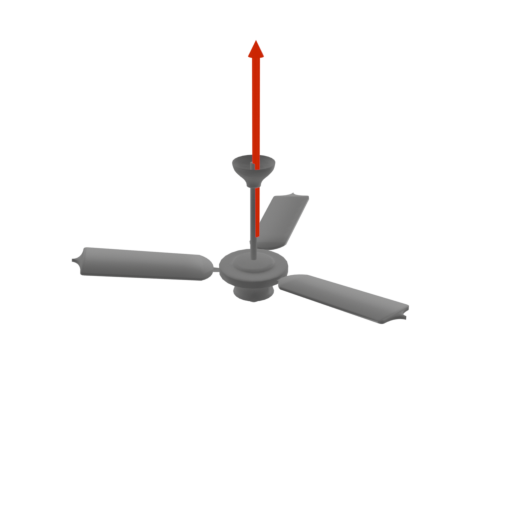} &
\includegraphics[width=\qualitWidth]{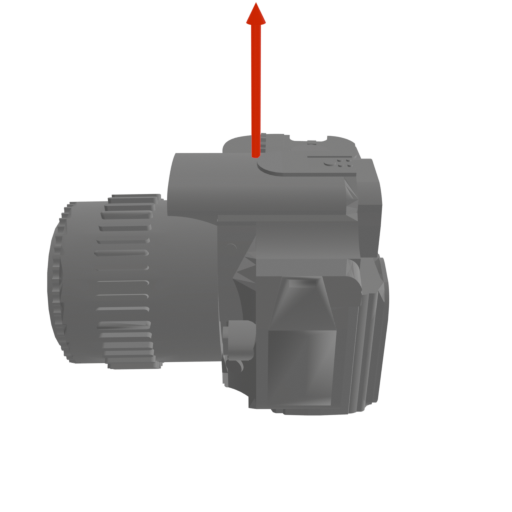} &
\epicfail &
\includegraphics[width=\qualitWidth]{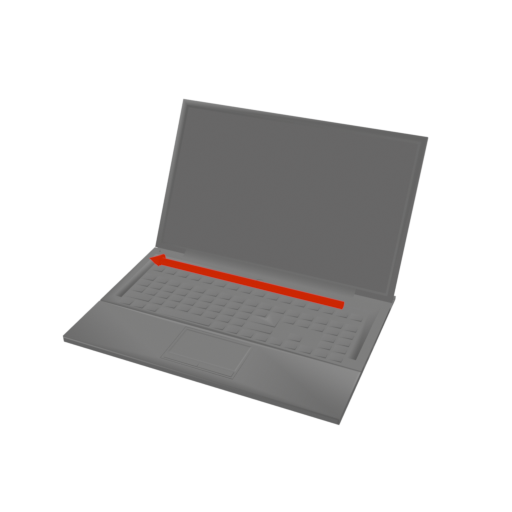} 
\\ [-0.5cm]

\rotatebox{90}{\hspace{0.7cm} ReArt} &
\rotatebox{90}{\hspace{0.8cm}} &
\epicfail &
\includegraphics[trim={4cm 5cm 4cm 5cm},clip,width=\qualitWidth]{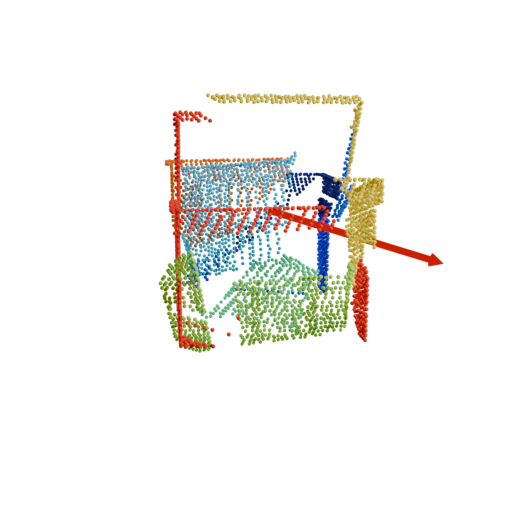} &
\includegraphics[trim={5cm 3cm 5cm 3cm},clip,width=\qualitWidth]{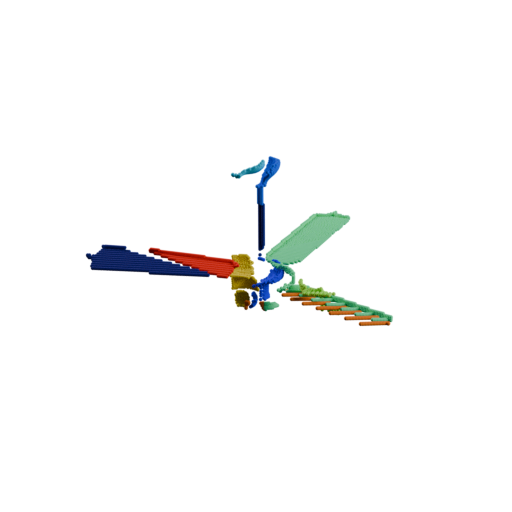} &
\epicfail &
\includegraphics[trim={3cm 3cm 3cm 3cm},clip,width=\qualitWidth]{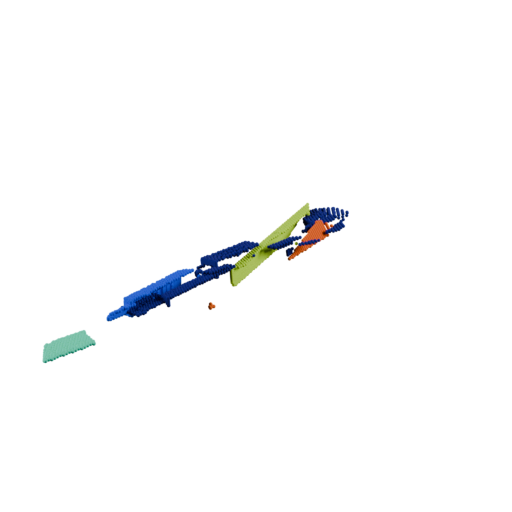} &
\epicfail 
\\ [-0.5cm]

\rotatebox{90}{\hspace{0.6cm}Video2} &
\rotatebox{90}{\hspace{0.4cm}Articulation} &
\includegraphics[trim={3cm 2cm 3cm 1cm},clip,width=\qualitWidth]{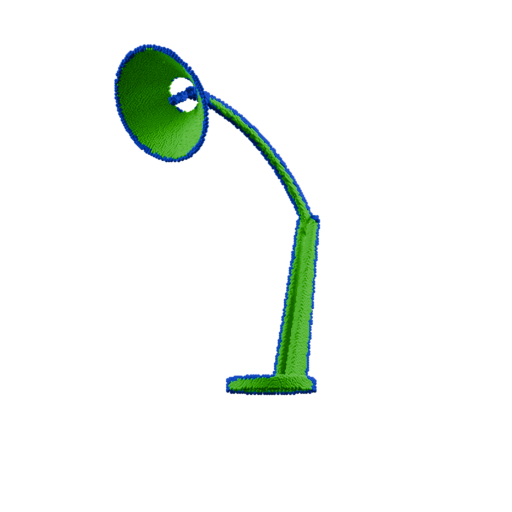}  &
\includegraphics[trim={4cm 5cm 4cm 5cm},clip,width=\qualitWidth]{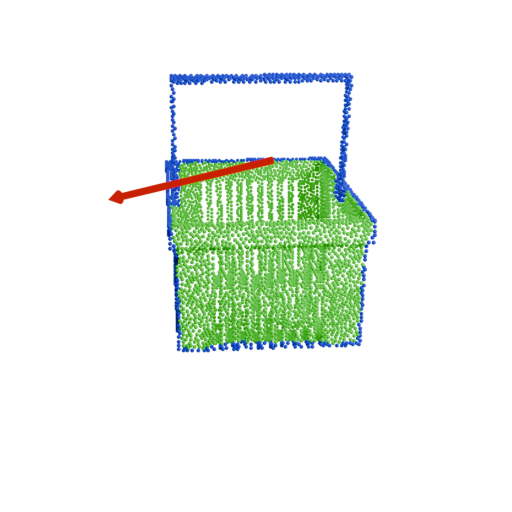} &
\epicfail &
\epicfail &
\includegraphics[trim={2cm 3cm 3cm 3cm},clip,width=\qualitWidth]{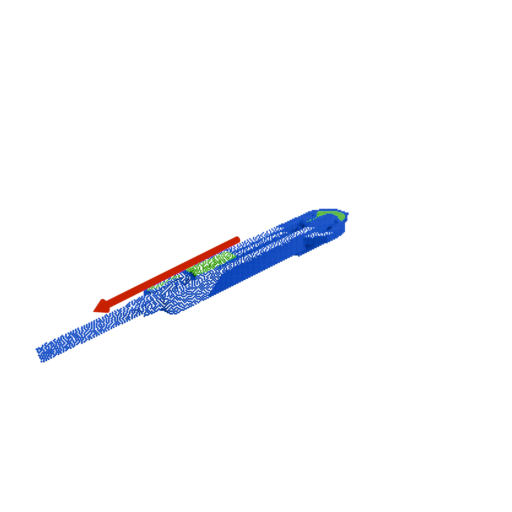}  &

\includegraphics[trim={3cm 2cm 3cm 2cm},clip,width=\qualitWidth]{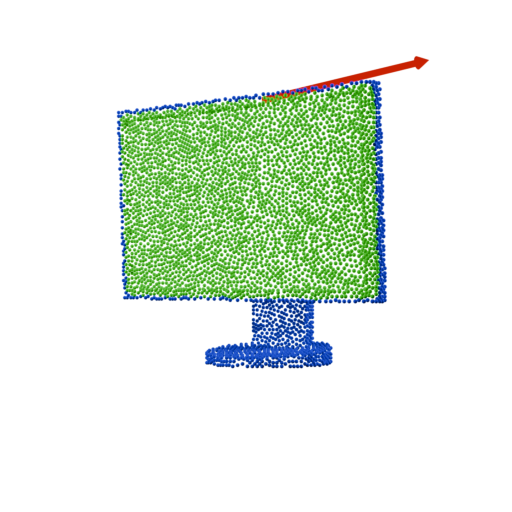} \\[-0.0cm]

& & Lamp & Bucket & Fan & Camera & Blade & Display \\

\end{tabular}
} 
   \caption{\textbf{Qualitative results on randomly selected sequences from the AiP-synth dataset.} Red arrows denote predicted joint axes. Note how our method retrieves the part segmentations and the joint axes much more accurately and robustly than all the other methods. '$\times$' indicates that the method fails on the category. \textit{Video results are provided in the video supplementary material.} }
   \label{fig:qual_fig_synth}
\end{figure*}


\def\qualitWidth{0.13\linewidth} 

\begin{figure*}[t]
  \centering
  \setlength{\tabcolsep}{2pt} 
  
  \resizebox{0.9\linewidth}{!}{
  \begin{tabular}{cccccccc} 
    
    & \footnotesize Rep. Frame & \footnotesize Ground Truth & \footnotesize Ours & \footnotesize Articulate-Anything & \footnotesize ReArt & \footnotesize Video2Articulation & \footnotesize Artipoint \\  \\[-1.5ex]

    \rotatebox{90}{\footnotesize \hspace{0.5cm} Box} &
    \includegraphics[trim={2cm 1cm 1cm 1.1cm},clip,width=\qualitWidth]{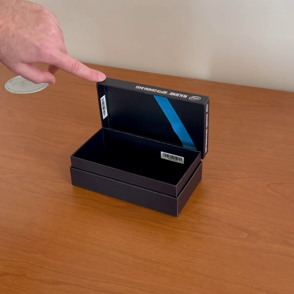} &
    \includegraphics[trim={2cm 0.25cm 2cm 2cm},clip,width=\qualitWidth]{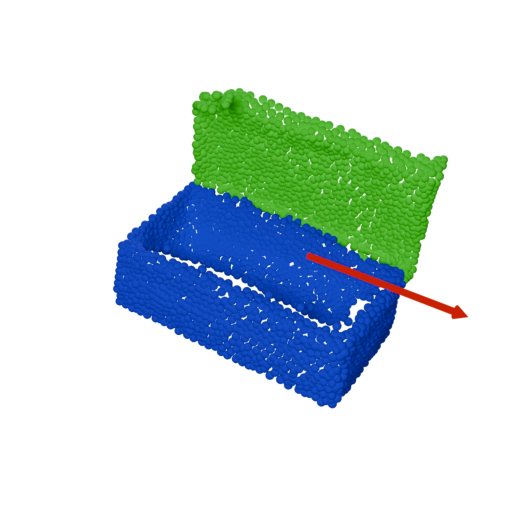} &
    \includegraphics[trim={2cm 1cm 1cm 1.1cm},clip,width=\qualitWidth]{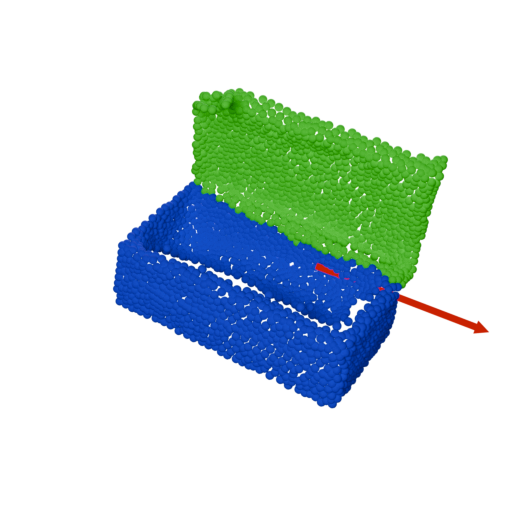} &
    \includegraphics[trim={1cm 1cm 1cm 1cm},clip,width=\qualitWidth]{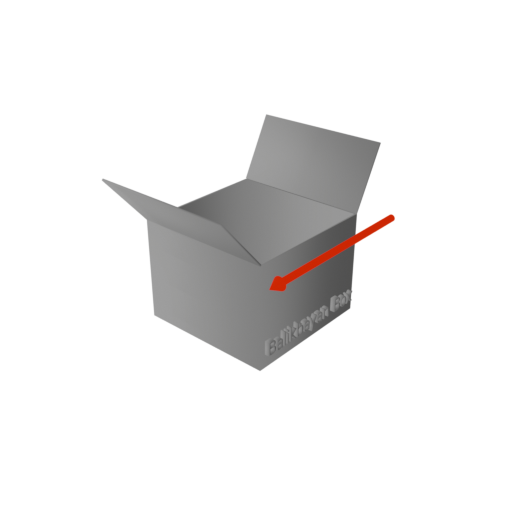} &
    \epicfail &
    \includegraphics[trim={2cm 1.28cm 2cm 1cm},clip,width=\qualitWidth]{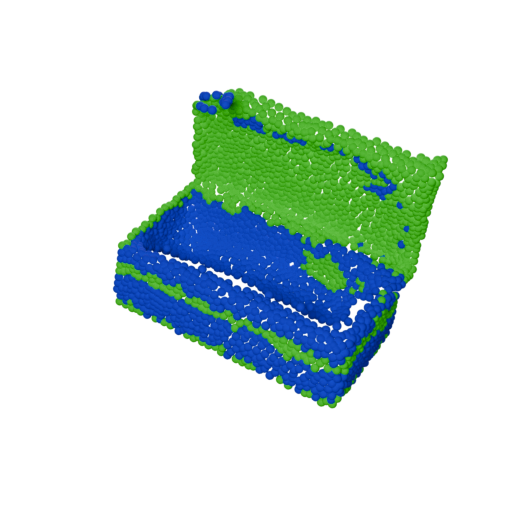} &
    \includegraphics[trim={1cm 1cm 1cm 1cm},clip,width=\qualitWidth]{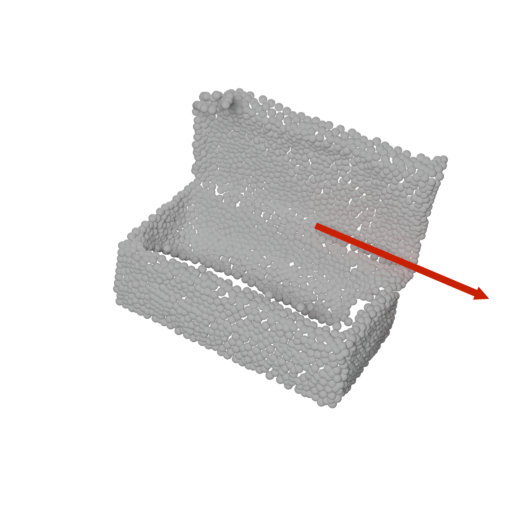} \\

    \rotatebox{90}{\footnotesize \hspace{0.5cm} Chair} &
    \includegraphics[trim={2cm 1cm 1cm 1.1cm},clip,width=\qualitWidth]{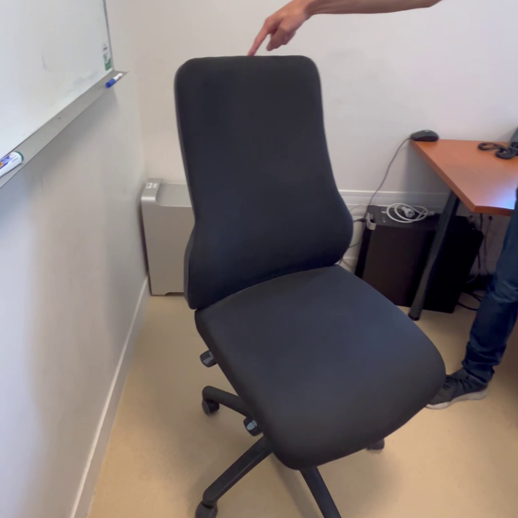} &
    \includegraphics[trim={2cm 1cm 2cm 1cm},clip,width=\qualitWidth]{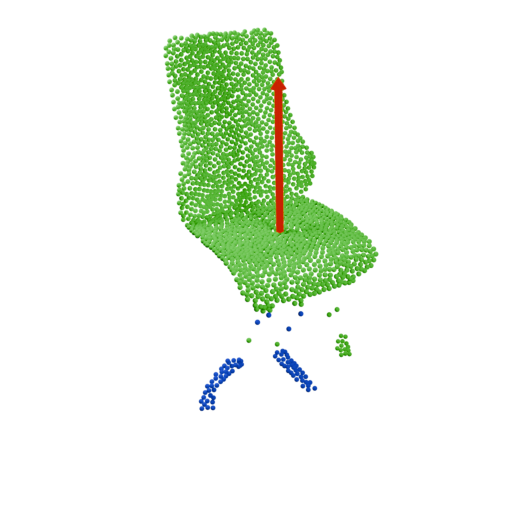} &
    \includegraphics[trim={2cm 1cm 2cm 1cm},clip,width=\qualitWidth]{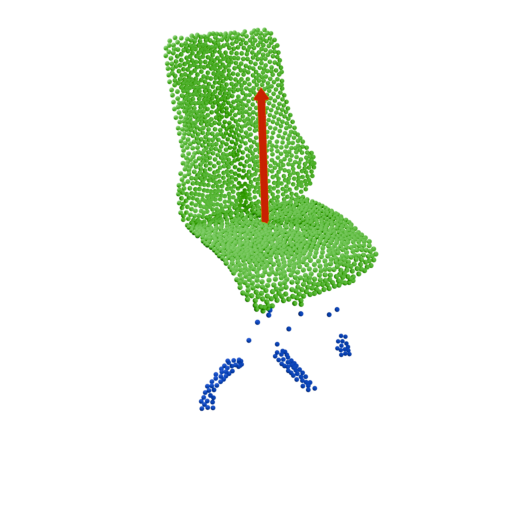} &
    \epicfail &
    \includegraphics[trim={2cm 1.28cm 2cm 1cm},clip,width=\qualitWidth]{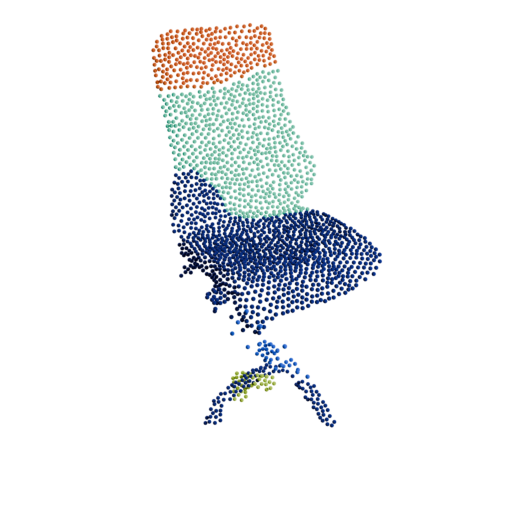} &
    \includegraphics[trim={2cm 1.28cm 2cm 1cm},clip,width=\qualitWidth]{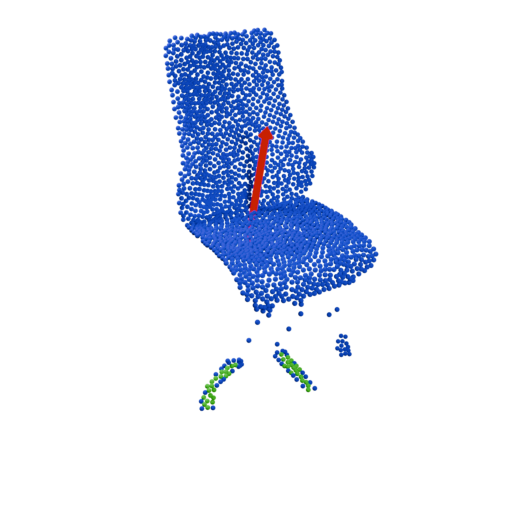} &
    \includegraphics[trim={2cm 1cm 2cm 1cm},clip,width=\qualitWidth]{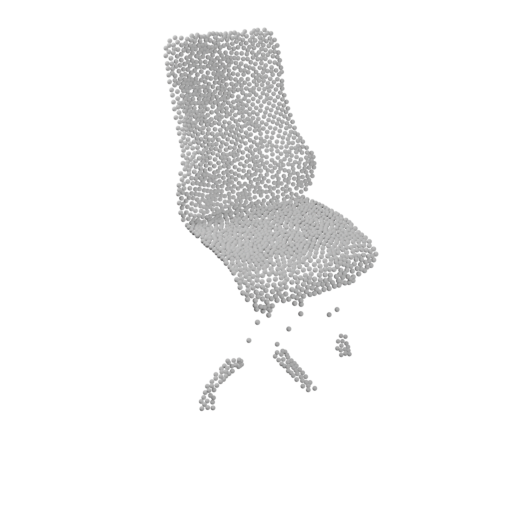} \\

    \rotatebox{90}{\footnotesize \hspace{0.1cm} Sliding Box} &
    \includegraphics[trim={2cm 1cm 1cm 1.1cm},clip,width=\qualitWidth]{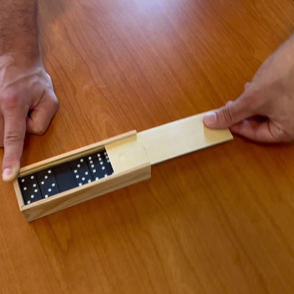} &
    \includegraphics[trim={2cm 1cm 2cm 1cm},clip,width=\qualitWidth]{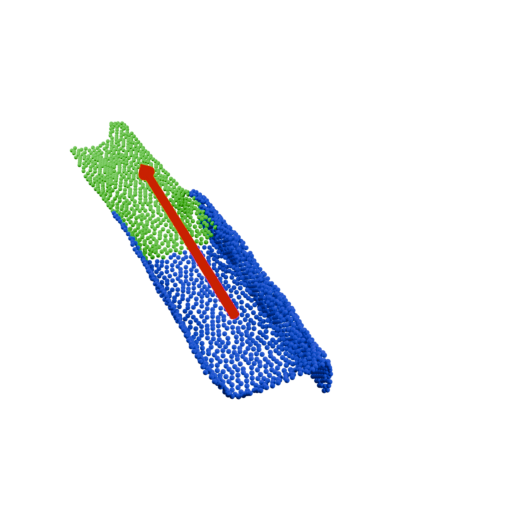} &
    \includegraphics[trim={2cm 1cm 2cm 1cm},clip,width=\qualitWidth]{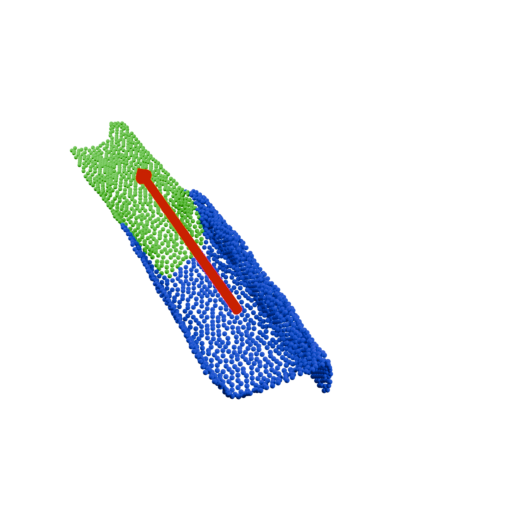} &
    \includegraphics[trim={1cm 1cm 1cm 1cm},clip,width=\qualitWidth]{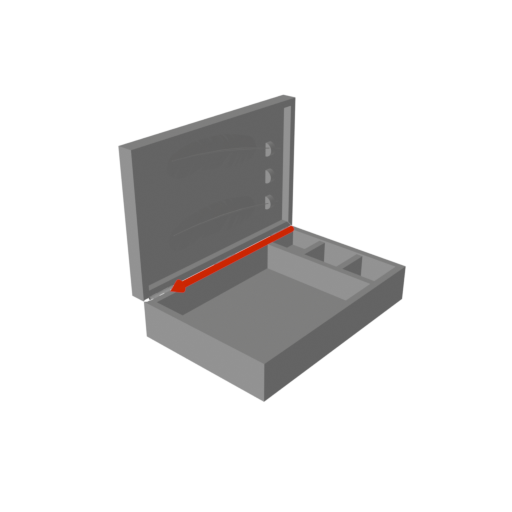} &
    \epicfail &
    \includegraphics[trim={2cm 1cm 2cm 1cm},clip,width=\qualitWidth]{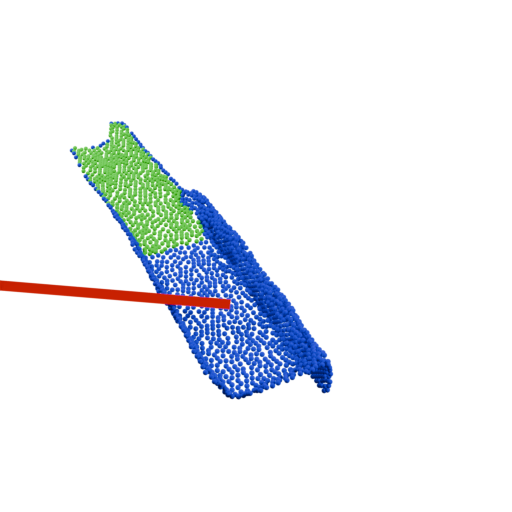} &
    \includegraphics[trim={2cm 1cm 2cm 1cm},clip,width=\qualitWidth]{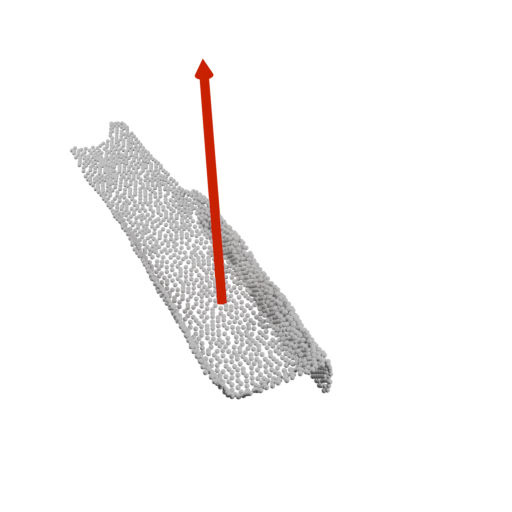} \\

    \rotatebox{90}{\footnotesize \hspace{0.5cm} Globe} &
    \includegraphics[trim={2cm 1cm 1cm 1.1cm},clip,width=\qualitWidth]{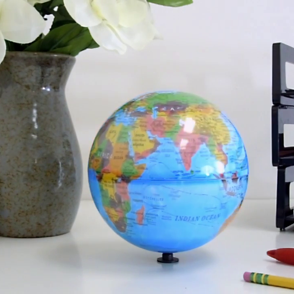} &
    \includegraphics[trim={2cm 1cm 2cm 1cm},clip,width=\qualitWidth]{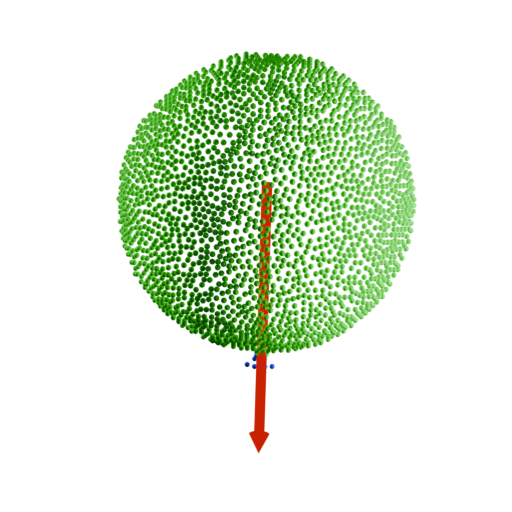} &
    \includegraphics[trim={2cm 1cm 2cm 1cm},clip,width=\qualitWidth]{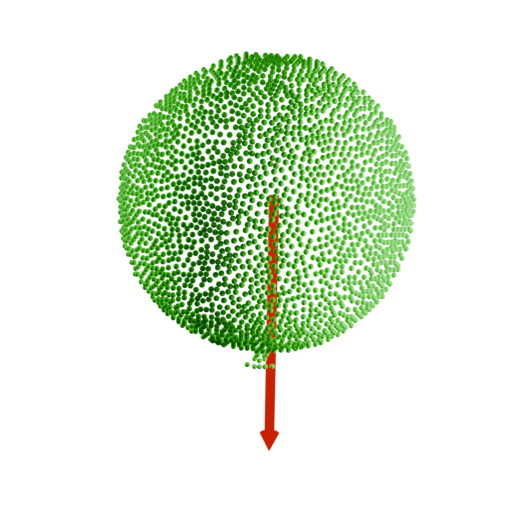} &
    \epicfail &
    \epicfail &
    \includegraphics[trim={2cm 1cm 2cm 1cm},clip,width=\qualitWidth]{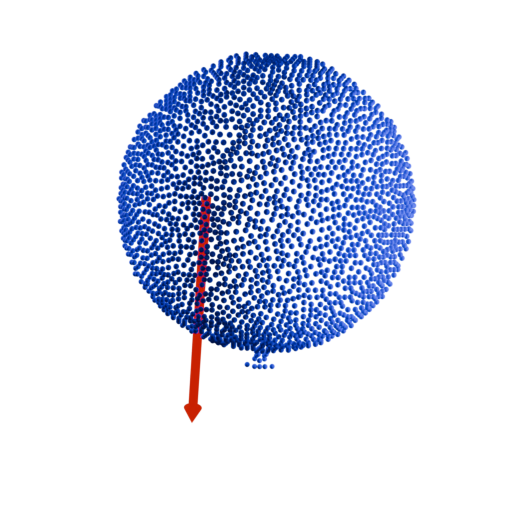} &
    \epicfail \\

    \rotatebox{90}{\footnotesize \hspace{0.1cm} Excavator} &
    \includegraphics[trim={2cm 1cm 1cm 1.1cm},clip,width=\qualitWidth]{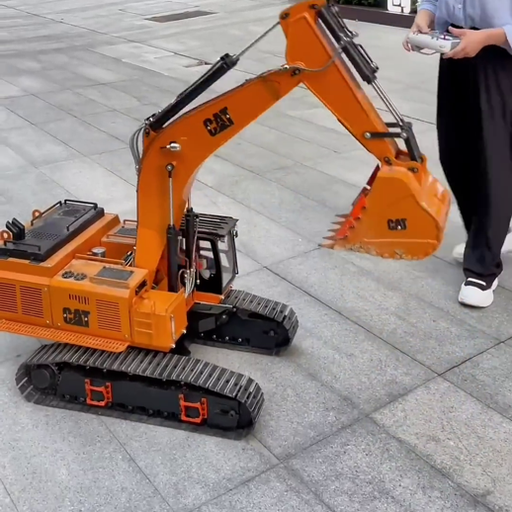} &
    \includegraphics[trim={2cm 1cm 2cm 1cm},clip,width=\qualitWidth]{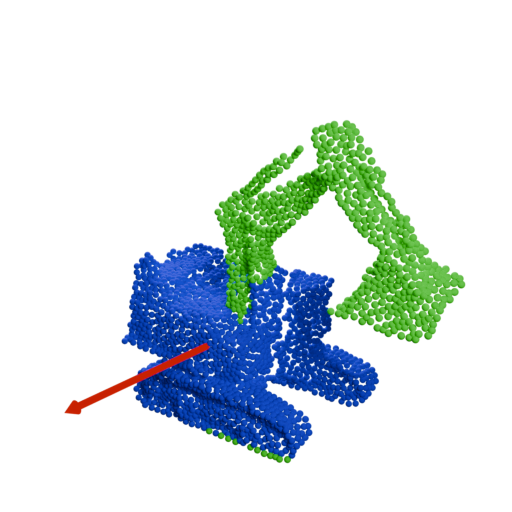} &
    \includegraphics[trim={2cm 1cm 2cm 1cm},clip,width=\qualitWidth]{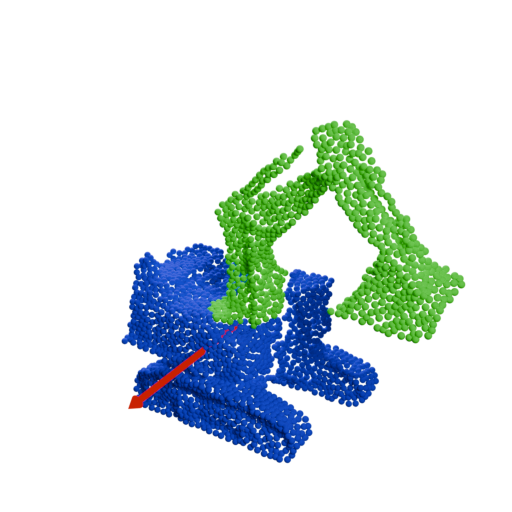} &
    \epicfail &
    \epicfail &
    \includegraphics[trim={2cm 1cm 2cm 1cm},clip,width=\qualitWidth]{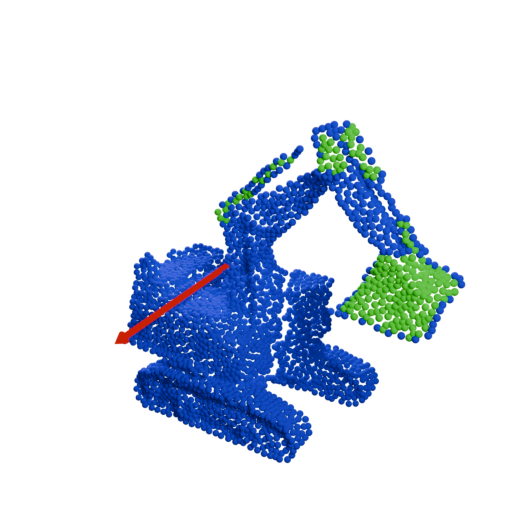} &
    \includegraphics[trim={2cm 1cm 2cm 1cm},clip,width=\qualitWidth]{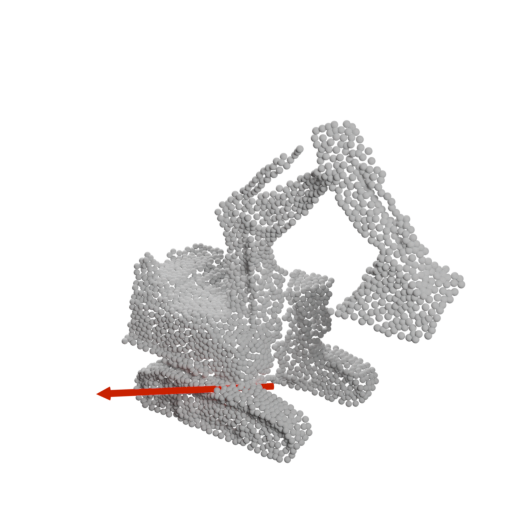} \\

  \end{tabular}
  } 
  
  \vspace{-5pt}
  \caption{\textbf{Qualitative results on the real-world  sequences from the AiP-real dataset.} Red arrows denote predicted joint axes. Note how our method retrieves the part segmentations and the joint axes much more accurately and robustly than all the other methods. '$\times$' indicates that the method fails on the category. \textit{Video results are provided in the video supplementary material.} }
  \label{fig:qual_fig_real}
\end{figure*}

\subsection{Ablation Analysis}

In Table~\ref{tab:ablation}, we report the results of an ablation study on the AiP-synth dataset to evaluate the contribution of each component in our framework. Each experiment is performed using three distinct seeds, and we report the results from the trial yielding the minimum overall loss. 
To assess the impact of the reconstruction loss, we systematically remove $\calL_\rec^{\calX\rightarrow\calQ}$, $\calL_\rec^{\calQ\rightarrow\calX}$, and their combination. 
The utility of the motion cues is validated by removing the scene flow loss. Finally, we investigate the influence of the regularization terms $\calL_\text{reg}$ by ablating the sparsity and existence constraints for both primitives and parts, along with the velocity magnitude and entropy constraints.

We observe a significant degradation in joint parameter estimation when the reconstruction loss terms are removed, as these components ensure that the primitives accurately fit and represent the observed point cloud. Similarly the omission of the flow loss prevents the model from effectively identifying moving components, resulting in a marked decline in dynamic part segmentation. Finally, the regularization terms facilitate convergence toward a physically plausible configuration by encouraging the model to represent the observed geometry with a minimal number of primitives and parts.

\newcommand{\scale}[2]{\MULTIPLY{#1}{#2}{\myr}\ROUND[2]{\myr}{\myr}\myr}

\vspace{-5pt}

\begin{table*}[t]
\centering
\setlength{\tabcolsep}{5.2pt}
\caption{
\label{tab:ablation}
\textbf{Ablation Experiments}. \nicolas{We demonstrate the importance of each loss term used in our optimization.} \arslan{`F' means the method fails to predict the metric. 
The complete method achieves state-of-the-art results across our evaluation suite. While specific sub-configurations may exhibit performance on par with the full method for certain metrics, they suffer from significantly higher error rates across the remaining criteria. 
}
}
\begin{adjustbox}{width=0.75\textwidth}
\begin{tabular}{@{}lcccccc@{}}
\toprule
       & mIoU & Axis & Position & Part Rotation & Part Translation & Type Accuracy \\
\emph{Method} & $\uparrow$ & ($^{\circ}$)~$\downarrow$ & (cm)~$\downarrow$ & ($\degree$)~$\downarrow$ & (cm)~$\downarrow$ & ($\%$)~$\uparrow$ \\
\midrule

w/o flow   &0.26 &13.62 &F &F &3.62 &75.0 \\

w/o $\calL_\rec^{\calX\rightarrow\calQ}$  &0.30 &45.38 &F &F &130.28 &0.00 \\

w/o $\calL_\rec^{\calQ\rightarrow\calX}$  &0.57 &22.50 & 0.00 & 0.00 &25.31 &75.0 \\

w/o $\calL_\rec^{\calX\rightarrow\calQ}$ + $\calL_\rec^{\calQ\rightarrow\calX}$ &0.34 &44.74 &F &F &48.00 &25.0 \\

w/o $\calL_\text{parts}$  &0.55 &44.74 &F & 0.00 &19.78 &50.0 \\

w/o $\calL_\text{prims}$ &0.65 &22.50 & 0.00 & 0.00 &26.18 &75.0 \\

w/o $\calL_{\text{match}}$ &0.55 &22.77 &F &F &42.15 &50.0 \\

w/o $\calL_{\text{motion}}$ &0.58 &45.00 &F &13.76 &39.27 &50.0 \\

Complete method &0.73 & 0.00 & 0.00 &0.53 & 0.00 & 100.0 \\

\bottomrule
\end{tabular}
\end{adjustbox}
\end{table*}

\section{Limitations}
Although we utilize superquadric primitives to obtain a coarse approximation of the object's geometry, this representation is insufficient for generating detailed mesh reconstructions. Consequently, incorporating more expressive primitive representations \cite{groueix2018papier, paschalidou2021neural} presents a promising avenue for future research, particularly for downstream applications like digital twins that necessitate high-fidelity surface reconstructions.

\section{Conclusion}

We presented an unsupervised framework for 
\tom{discovering}
part labels and kinematic joint parameters of articulated objects. Extensive experiments across four challenging benchmarks demonstrate the accuracy and robustness of our approach, even when faced with complex camera trajectories, low-quality depth maps (e.g., in the Arti4D dataset), and intricate articulated configurations such as those found in bottles or cameras. Furthermore, our per-video optimization formulation ensures strong generalization across diverse object categories.

Overall, our method is the first to frame articulated object reconstruction and joint parameter estimation as a primitive-fitting task. We demonstrate that casual monocular video provides sufficient geometric regularity to recover articulated motion without the need for dense correspondences or category-specific training.

\section{Acknowledgements}

This project was funded in part by the European Union (ERC Advanced Grant explorer Funding ID
\#101097259). This work was granted access to the HPC resources of IDRIS under the allocation
2026-AD011014756R2 made by GENCI.


\bibliographystyle{splncs04}
\bibliography{main}

\clearpage

\begingroup
  \begin{center}
    {\LARGE\bfseries Appendix\par}
    \vspace{0.5em}
    {\Large\bfseries Articulation in Prime: Primitive-Based Articulated Object Understanding from a Single Casual Video\par}
  \end{center}
\endgroup
\vspace{1em}

\appendix

\section{Superquadrics}
\label{supp:superquadric}

\begin{figure}[H]
    \centering
    \input{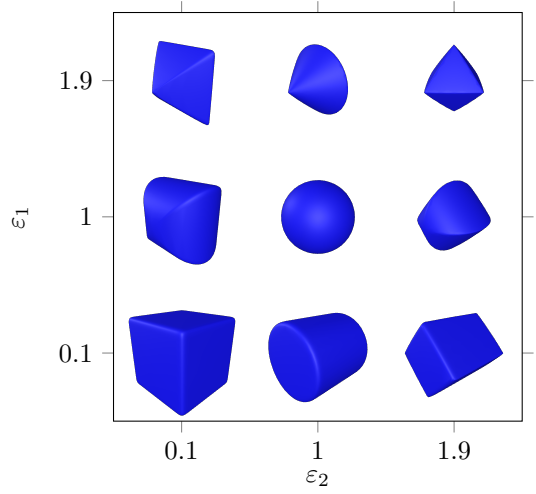}
    \caption{\textbf{Influence of the shape parameters on a unit scale superquadric.} The shape parameters are constrained to the interval $[0.1, 1.9]$ so that superquadrics are convex and more representative of shapes encountered in the real world.}
    \label{fig:superquadric-grid}
\end{figure}

Superquadrics~\cite{barr1981superquadrics} form a parametric family of 3D shapes. Their popularity stems from their ability to represent a wide variety of shape variations with only a small number of parameters. A canonical superquadric is characterized by its scale $\mathbf{s} = (s_x, s_y, s_z) \in \bm{R}^3$ and its shape parameters $\bm{\varepsilon} = (\varepsilon_1, \varepsilon_2) \in \mathbb{R}^2$. It is common practice to constrain the shape parameters to the interval $[0.1, 1.9]$ in order to ensure convex superquadrics without overly sharp edges~\cite{SuperquadricFitting-TPAMI19}. Figure~\ref{fig:superquadric-grid} illustrates how the shape parameters affect the geometry of superquadrics.

\subsection{Equations}

The superquadric surface is defined by the implicit equation:
\begin{equation}
    \left[\left(\frac{x}{s_x}\right)^{2/\varepsilon_2} + \left(\frac{y}{s_y}\right)^{2/\varepsilon_2}\right]^{\varepsilon_2/\varepsilon_1} + \left(\frac{z}{s_z}\right)^{2/\varepsilon_1} = 1
\label{eq:superquadric_implicit}
\end{equation}
or equivalently, by the explicit equation:
\begin{equation}
\left\{
\begin{aligned}
    &x = s_x \cos^{\varepsilon_1}\eta \cos^{\varepsilon_2}\omega \\
    &y = s_y \cos^{\varepsilon_1}\eta \sin^{\varepsilon_2}\omega \\
    &z = s_z \sin^{\varepsilon_1}\eta \\
\end{aligned}
\right.
\quad
\begin{array}{l}
    \eta \in [-\pi/2,\, \pi/2] \\[4pt]
    \omega \in [-\pi,\, \pi]
\end{array}
\label{eq:superquadric_explicit}
\end{equation}

\subsection{Point Sampling}
\label{supp:superquadric_sampling}

The implicit equation~\ref{eq:superquadric_implicit} proves useful for sampling points on the surface of a superquadric. Let $\bm{v} = (v_x, v_y, v_z) \in \mathbb{R}^3$ be a random direction satisfying $\|\bm{v}\|_2 = 1$. Solving for $t \geq 0$ such that $(x, y, z) = t \bm{v}$ lies on the surface of the unit superquadric yields:
\begin{equation}
    t = \left[\left(|v_x|^{2/\varepsilon_2} + |v_y|^{2/\varepsilon_2}\right)^{\varepsilon_2/\varepsilon_1} + |v_z|^{2/\varepsilon_1} \right]^{-\varepsilon_1/2}
\end{equation}
We then apply the scaling parameters to obtain the corresponding point on the surface of the scaled superquadric. This simple sampling procedure is differentiable by construction, fast to compute, and empirically performs on par with the standard uniform sampling method by Pilu \textit{et al.}~\cite{pilu1995equal}.

\subsection{Mesh Construction}
\label{supp:superquadric_mesh}

A mesh representation of a superquadric can be easily derived from the explicit equation~\ref{eq:superquadric_explicit}, following~\cite{monnier2023differentiable}. The idea is to define a mapping between the unit sphere and the scaled superquadric, the sphere being a special case of the superquadrics with shape parameters $\bm{\varepsilon} = (1, 1)$ (Figure~\ref{fig:superquadric-grid}). We first construct a unit icosphere whose vertices all lie on the unit sphere. Then, for each vertex, we compute the pair $(\eta, \omega)$ using the explicit equation of the unit sphere, and apply the explicit equation of the scaled superquadric to get the deformed point. The mesh connectivity is preserved.

\section{Implementation Details}

\subsection{Design Choices}

In our experiments, we optimize over sequences of $T = 25$ frames, sampling a point cloud of $N = 4096$ point from each frame. The maximum number of primitives that can be used in the reconstruction is set to $Q = 10$, while the maximum number of parts is set to $P = 8$. 
Superquadrics are initialized with a uniform scale of $0.5$, shape parameters $(\varepsilon_1, \varepsilon_2) = (0.2, 0.2)$ (close to cuboids), an existence probability of $0.5$, and a random pose at the first timestep. We use the 6D rotation representation~\cite{zhou2019continuity} to parameterize the orientation at the first timestep.

We use a 3-layer MLP to compute point features from the 3D coordinates. Both point and primitive feature vectors are of dimension $16$.

\subsection{Optimization Details}\label{appendix:optim}

To compute the reconstruction loss, we sample $500$ points per primitive using the procedure described in Section~\ref{supp:superquadric_sampling}. Occlusions are computed efficiently by rendering the primitives into a Z-buffer with Nvdiffrast~\cite{Laine2020diffrast}, where superquadric meshes are derived from transformed icospheres as described in Section~\ref{supp:superquadric_mesh}.

We optimize for 10k steps with a learning rate of $5 \times 10^{-3}$, without learning rate warmup of scheduling, and employ the Adam~\cite{kingma-icml15-adam} optimizer.

Table~\ref{tab:loss_weights} reports the weighing coefficients of our loss terms.
\renewcommand{\arraystretch}{1.1}
\begin{table}[h]
    \centering
    \begin{tabular}{@{}cc@{}}
        \toprule
        \textbf{Loss term} & \textbf{Weight} \\
        \midrule
        $\mathcal{L}_{rec}^{\mathcal{Q} \rightarrow \mathcal{X}}$ & 2.0 \\
        $\mathcal{L}_{rec}^{\mathcal{X} \rightarrow \mathcal{Q}}$ & 2.0 \\
        $\mathcal{L}_{\text{flow}}$ & 1.0 \\
        $\mathcal{L}_{\text{part-sparsity}}$ & 0.1 \\
        $\mathcal{L}_{\text{part-existence}}$ & 0.05 \\
        $\mathcal{L}_{\text{prim-sparsity}}$ & 0.01 \\
        $\mathcal{L}_{\text{prim-existence}}$ & 0.05 \\
        $\mathcal{L}_{\text{motion}}$ & 0.1 \\
        $\mathcal{L}_{\text{match}}$ & 0.1 \\
        \bottomrule
    \end{tabular}
    \caption{\textbf{Weighting factors of our loss terms.} These weights were set manually with minimal tuning.}
    \label{tab:loss_weights}
\end{table}

\subsection{Computational Cost}

Our method runs in about 10 minutes per object on a single GPU NVIDIA A6000  with a processor 32x Xeon(R) Silver 4208. We run 5 seeds per object and pick the result with lowest loss after the optimization converges. For the ablations we run only 3 seeds per object.

\section{Details on AiP-synth and AiP-real Datasets}

\subsection{AiP-synth}
\label{supp:aip_synth}

\begin{figure}[!htb]
    \centering
    \includegraphics[width=0.7\linewidth]{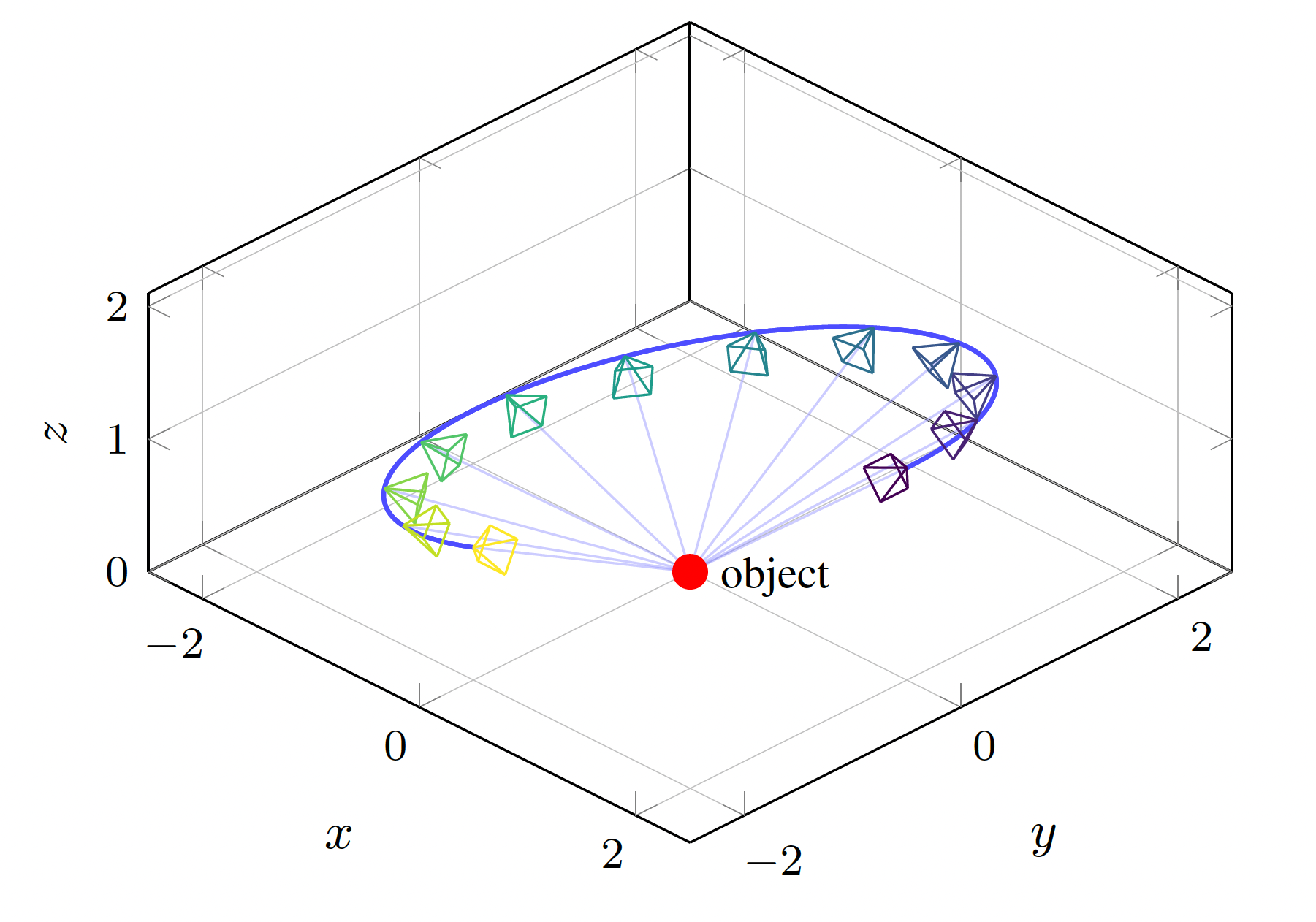}
    \caption{\textbf{A synthetic circular camera trajectory used in AiP-synth.} The camera turns around the object with random start and end angles, and a random height variation amplitude. In the case of the ping-pong trajectory, the camera sweeps back and forth along the circular path.}
    \label{fig:camera_trajectory}
\end{figure}

The dataset comprises a diverse selection of objects extracted from PartNet-Mobility, illustrated in Figure~\ref{fig:category_fig}. To simulate a real capture system, we define a circular trajectory and a ping-pong trajectory. We illustrate them in Figure~\ref{fig:camera_trajectory}.

\def\instanceWidth{0.13\linewidth}

\begin{figure*}
  \centering
  \setlength{\tabcolsep}{1.6pt} 
  \begin{tabular}{cccccccc}
    \includegraphics[trim={3cm 3cm 3cm 3cm},clip,width=\instanceWidth]{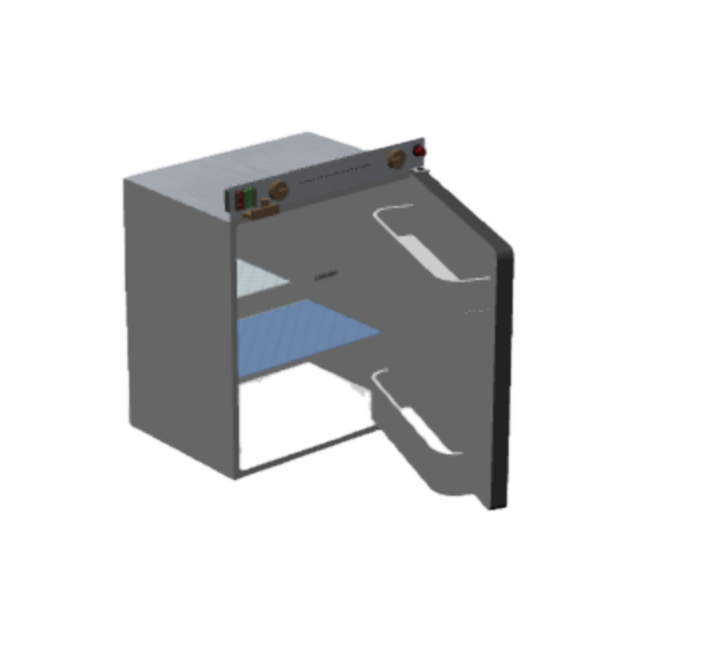} &
    \includegraphics[trim={3cm 3cm 3cm 3cm},clip,width=\instanceWidth]{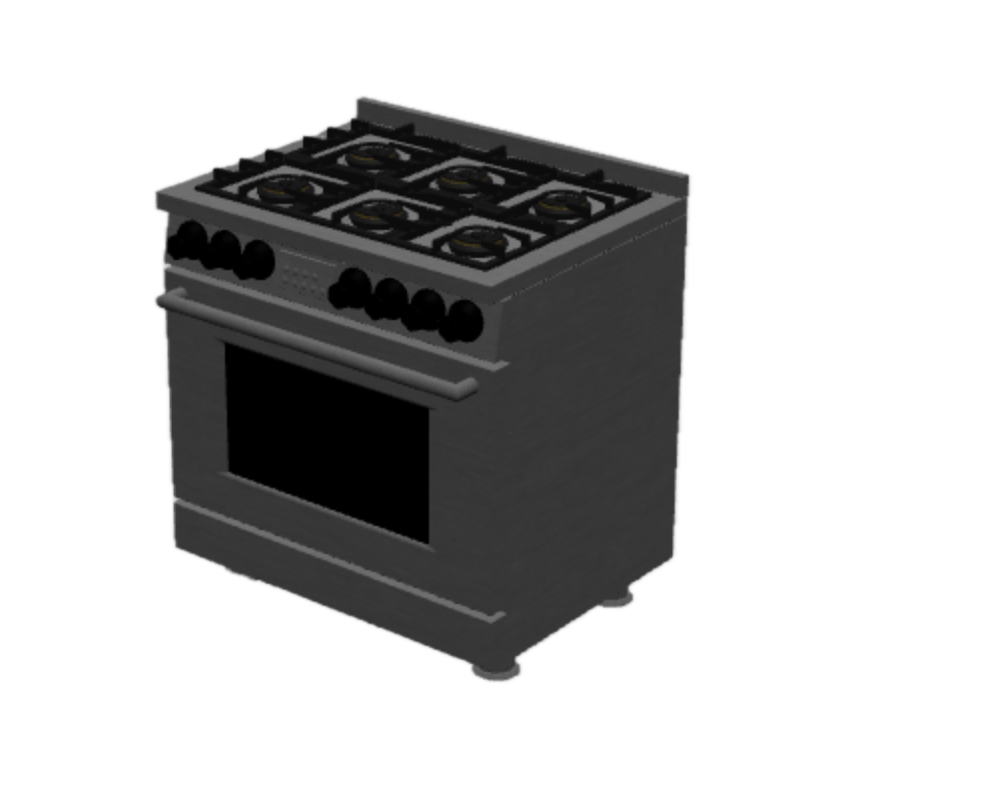} &
    \includegraphics[trim={3cm 3cm 3cm 3cm},clip,width=\instanceWidth]{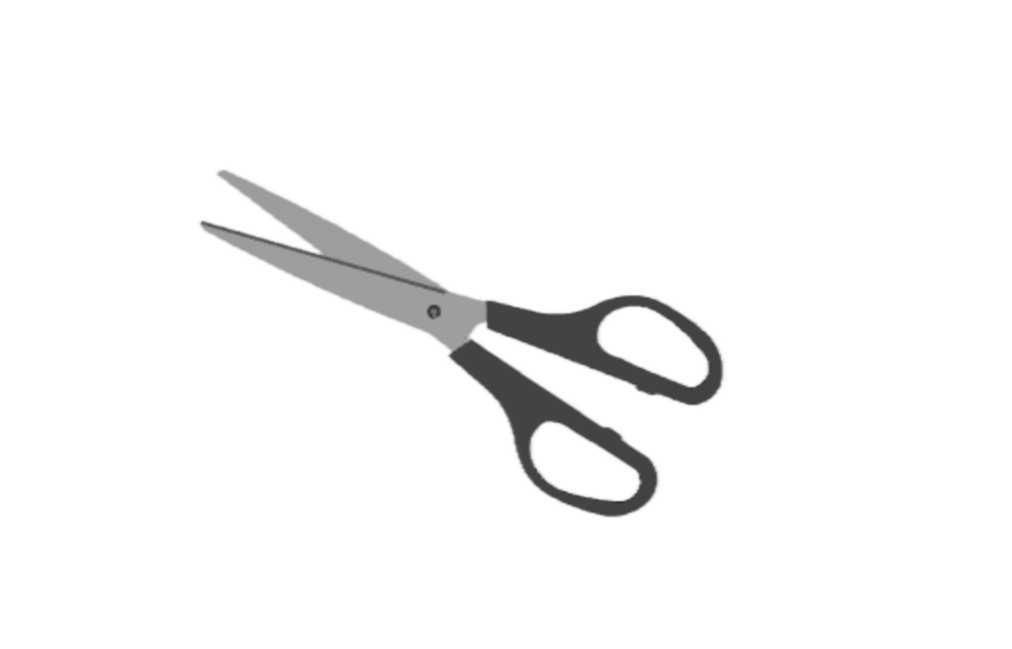} &
    \includegraphics[trim={3cm 3cm 3cm 3cm},clip,width=\instanceWidth]{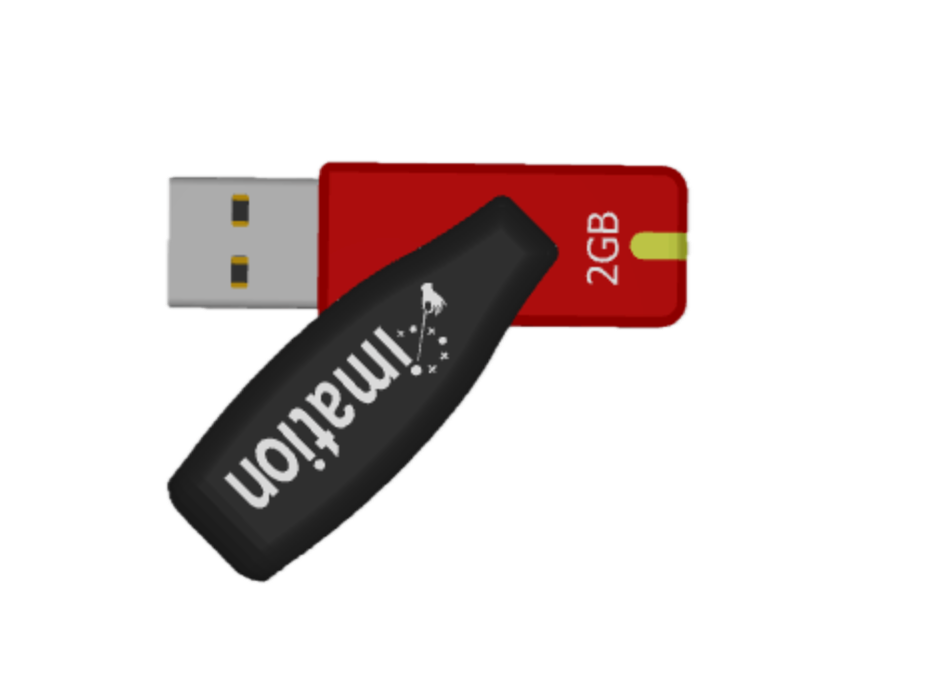} &
    \includegraphics[trim={5cm 3cm 5cm 3cm},clip,width=\instanceWidth]{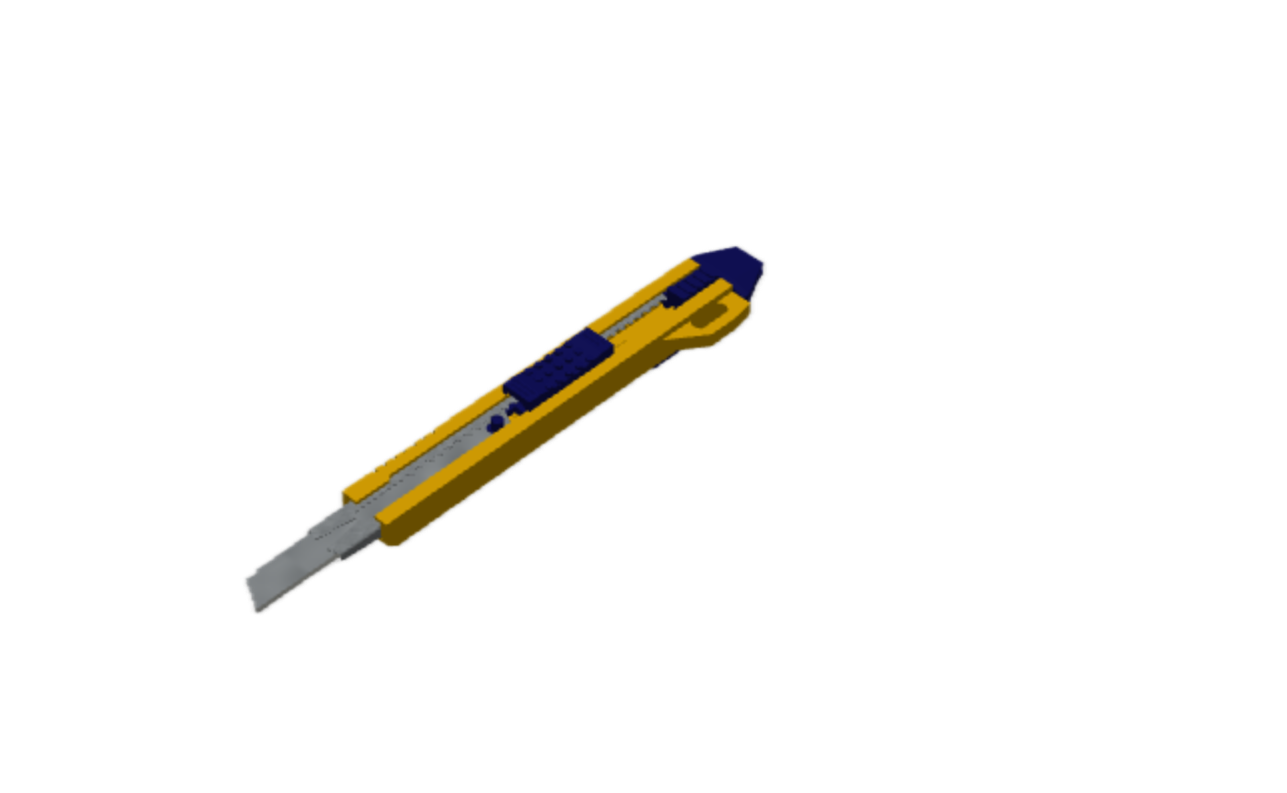} &
    \includegraphics[trim={3cm 3cm 3cm 3cm},clip,width=\instanceWidth]{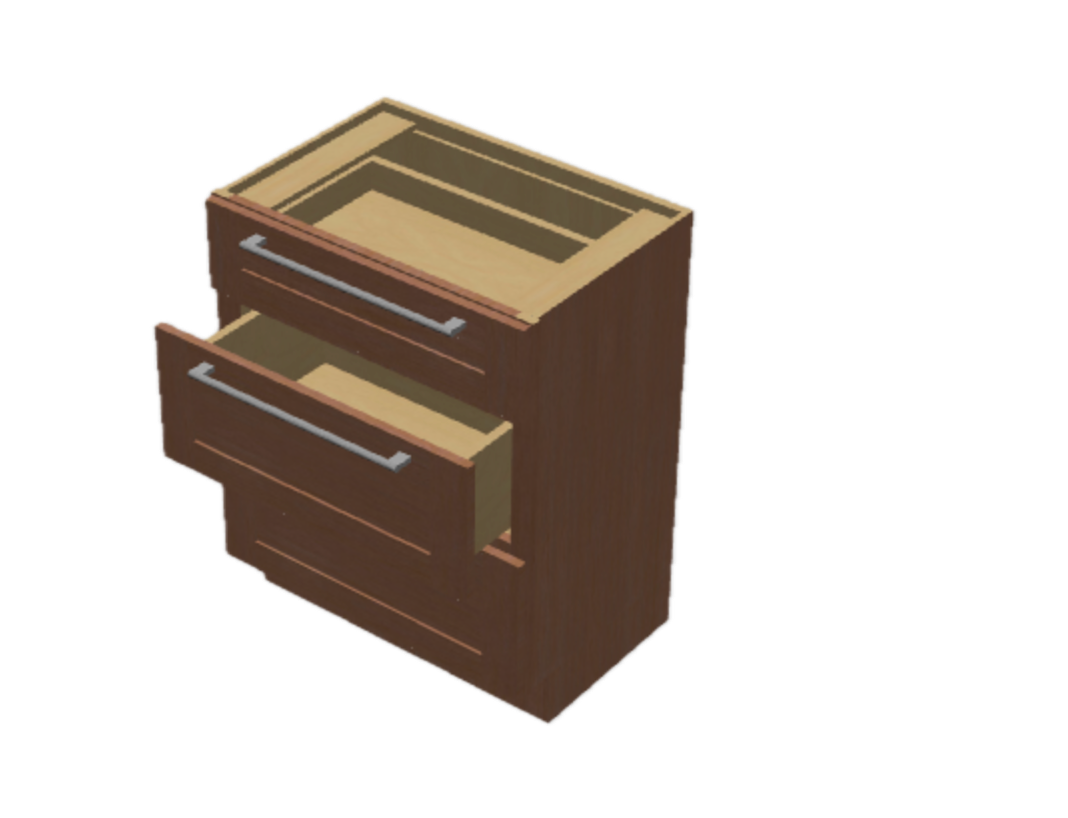} &
    \includegraphics[trim={3cm 3cm 3cm 3cm},clip,width=\instanceWidth]{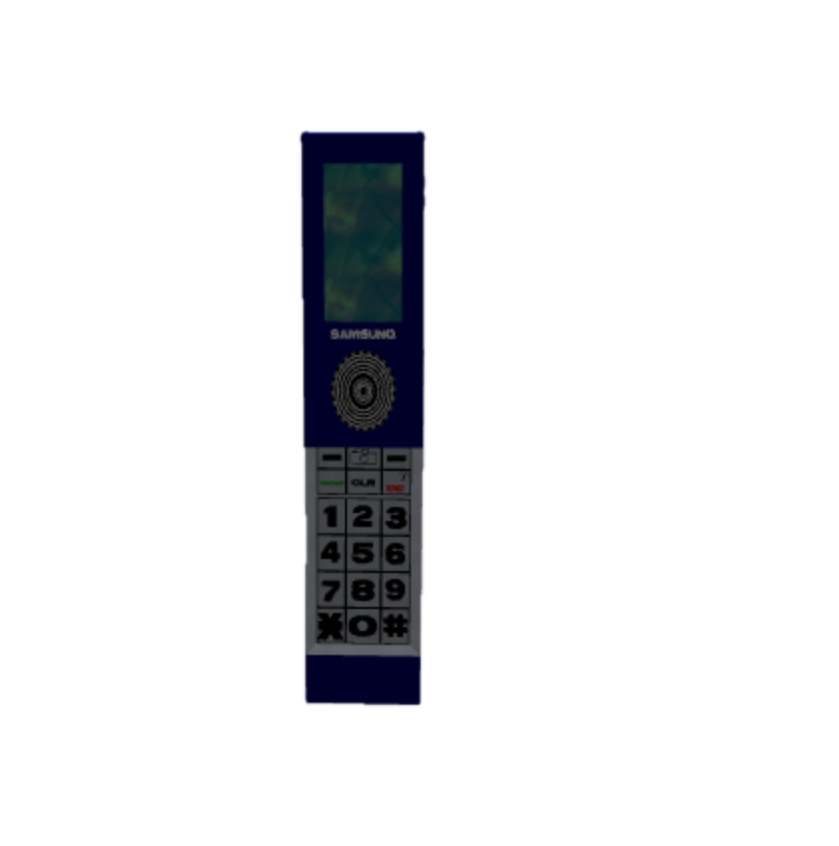}  &
    \includegraphics[trim={3cm 3cm 3cm 3cm},clip,width=\instanceWidth]{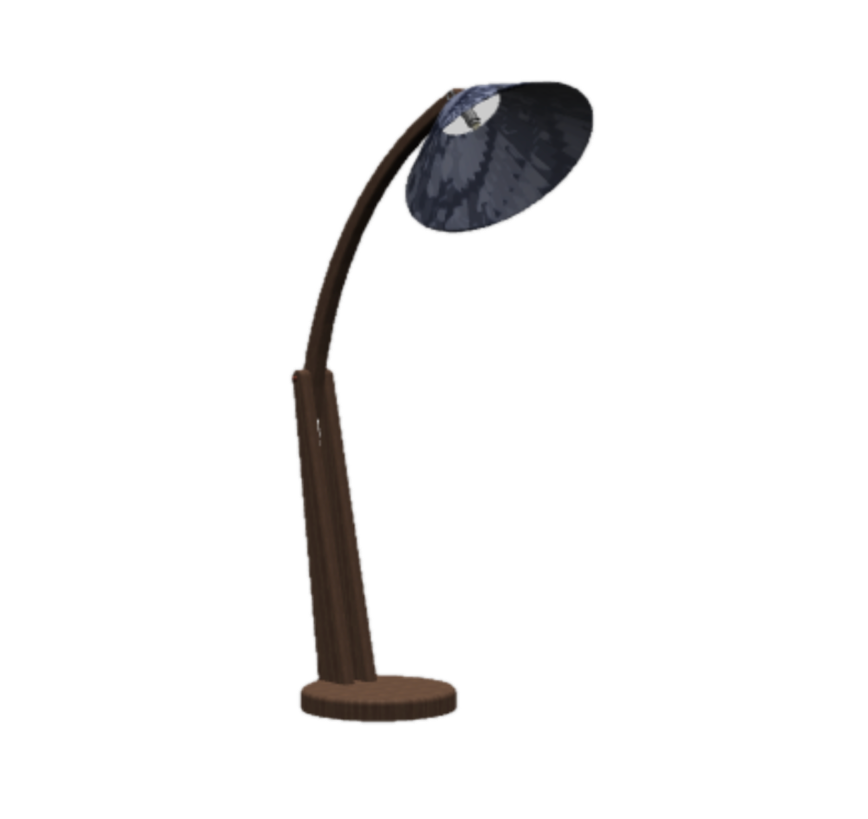}  \\
    \small Fridge & 
    \small Oven & 
    \small Scissors & 
    \small USB & 
    \small Blade & 
    \small Drawer &
    \small Phone &
    \small Lamp \\[4pt]
    \includegraphics[trim={3cm 3cm 3cm 3cm},clip,width=\instanceWidth]{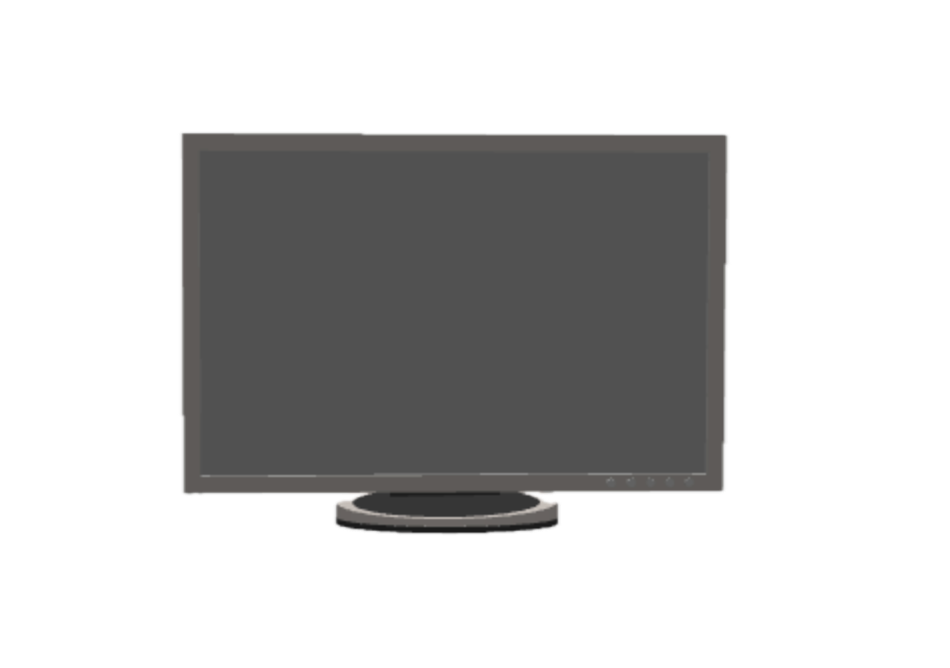} &
    \includegraphics[trim={3cm 3cm 3cm 3cm},clip,width=\instanceWidth]{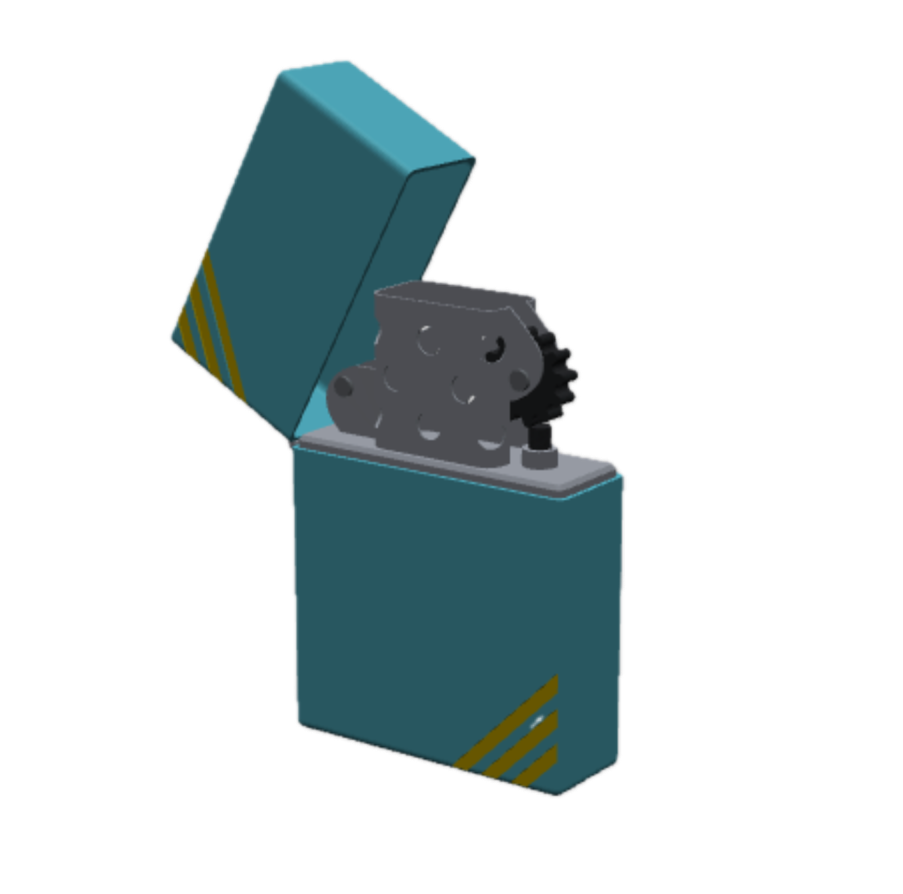} &
    \includegraphics[trim={3cm 3cm 3cm 3cm},clip,width=\instanceWidth]{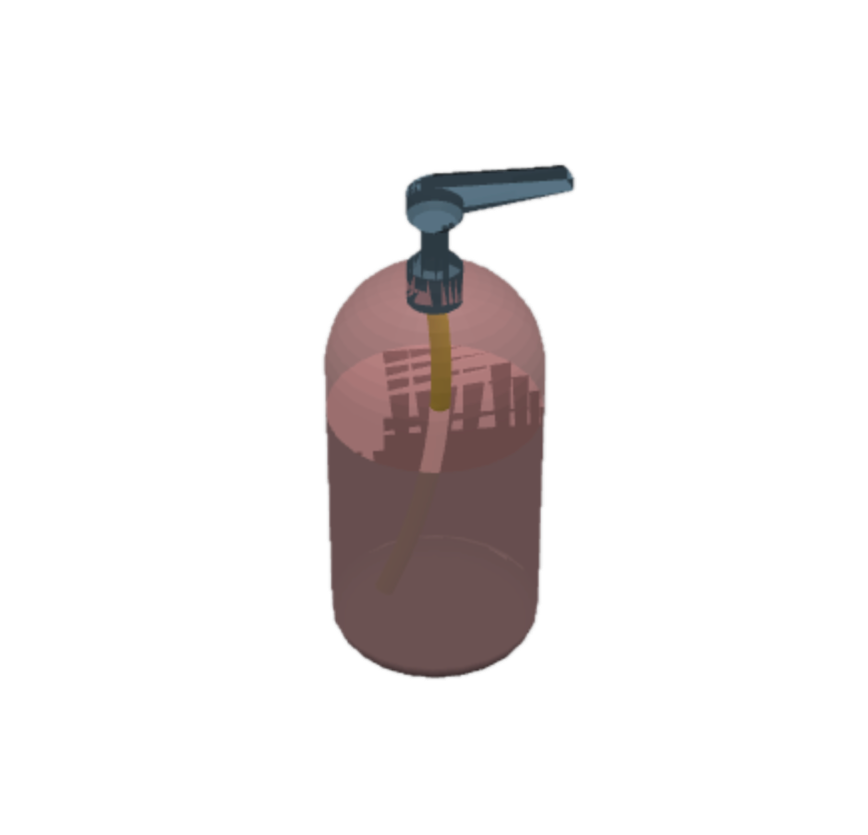} &
    \includegraphics[trim={3cm 3cm 3cm 3cm},clip,width=\instanceWidth]{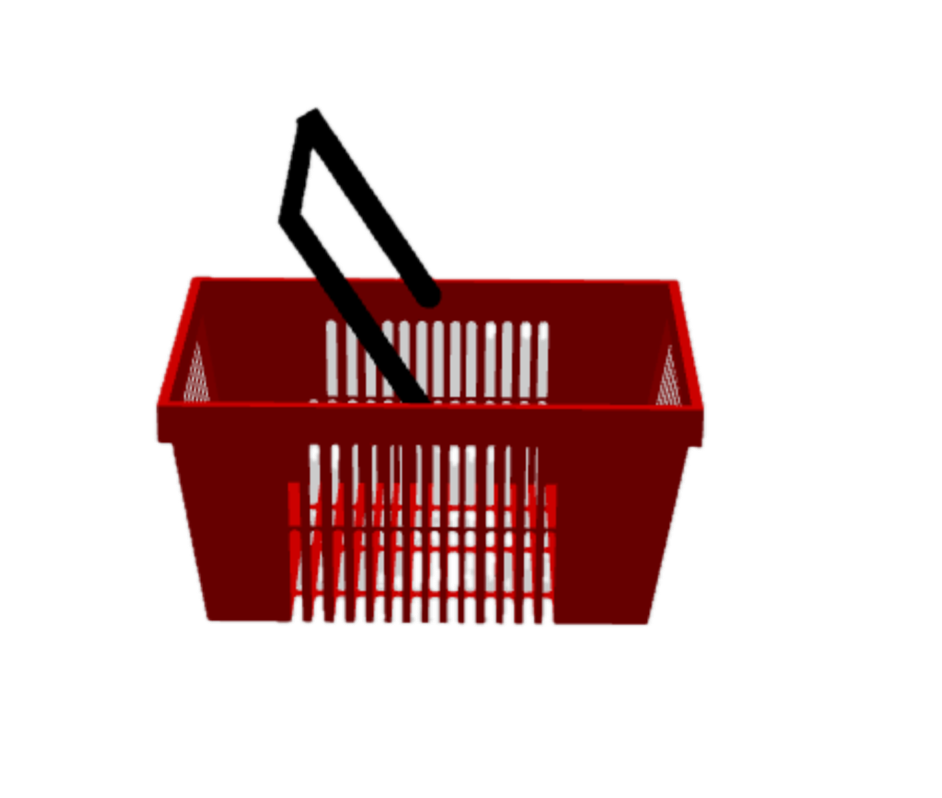} &
    \includegraphics[trim={3cm 3cm 3cm 3cm},clip,width=\instanceWidth]{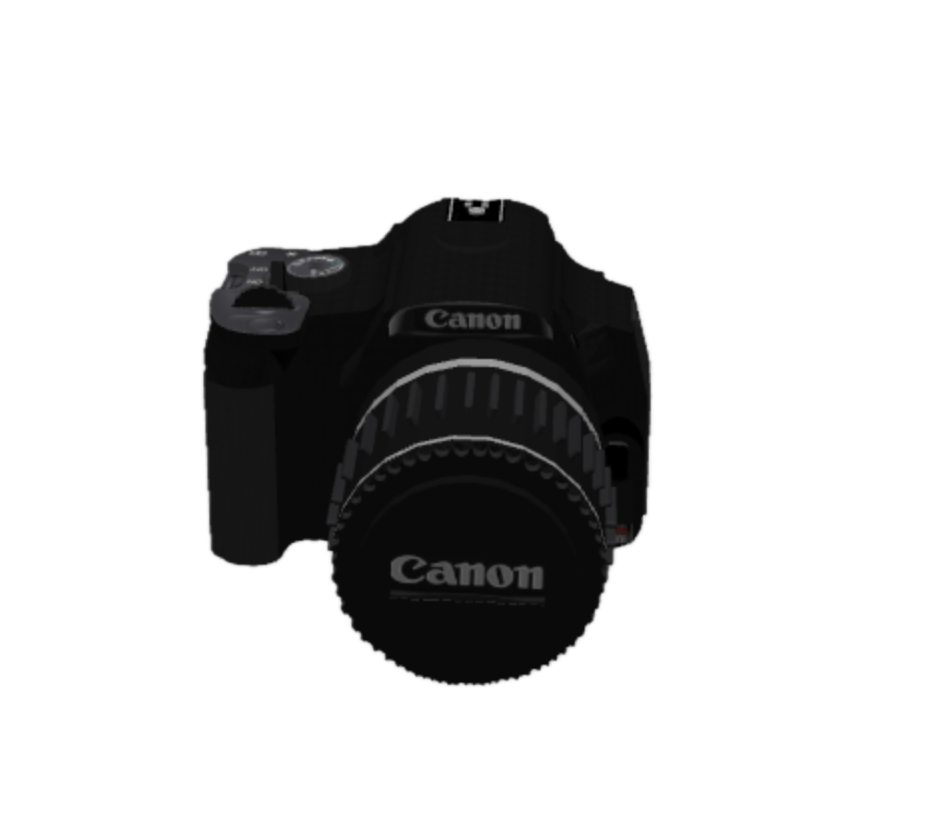} &
    \includegraphics[trim={3cm 3cm 3cm 3cm},clip,width=\instanceWidth]{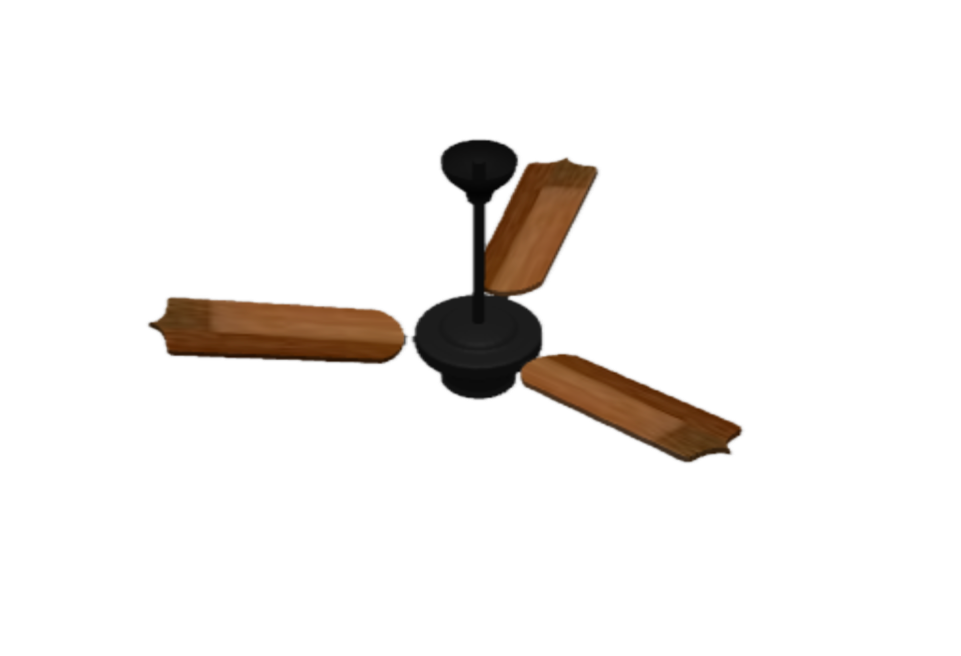} &
    \includegraphics[trim={3cm 3cm 3cm 3cm},clip,width=\instanceWidth]{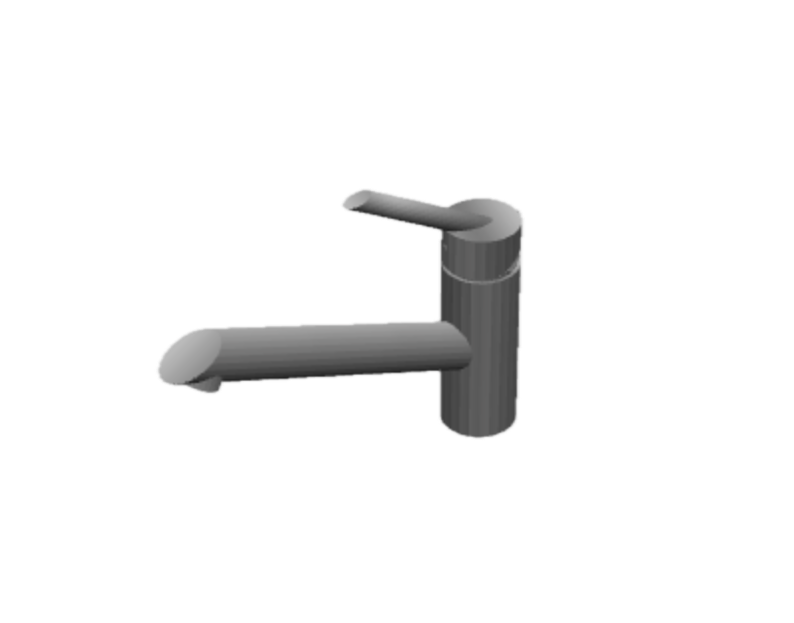}  &
    \includegraphics[trim={3cm 3cm 3cm 3cm},clip,width=\instanceWidth]{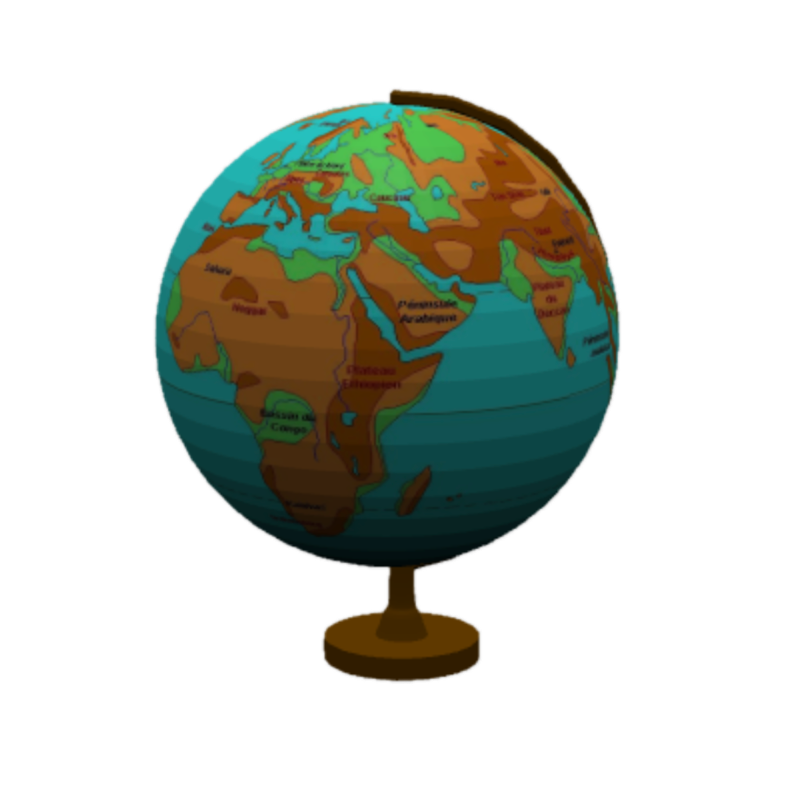}  \\
    \small Display & 
    \small Lighter & 
    \small Bottle & 
    \small Bucket & 
    \small Camera & 
    \small Fan &
    \small Faucet &
    \small Globe \\
  \end{tabular}
  \caption{\textbf{Representative frames of all the objects used in our AiP-synth dataset.} This selection covers a wide range of geometries and kinematics. Some object parts are particularly complex and require the use of multiple primitives. Other objects, such as the globe, exhibit symmetries that make the joint discovery more challenging.}
  \label{fig:category_fig}
\end{figure*}

\paragraph{Circular trajectory.}

Let $t \in [0, 1]$ parametrize the trajectory. The azimuth angle is linearly interpolated between the start and end angles:
\begin{equation}
    \theta(t) = \theta_{\text{start}} + t\,(\theta_{\text{end}} - \theta_{\text{start}})
\end{equation}

Finally, the camera position $\mathbf{p}(t) \in \mathbb{R}^3$ is given by:
\begin{equation}
\mathbf{p}(t) =
\begin{pmatrix}
r \cos\theta(t) \\
r \sin\theta(t) \\
h_0 + \Delta h \cos\theta(t)
\end{pmatrix}
\label{eq:camera_center}
\end{equation}
where $r$ is the radius, $h_0$ the base height, and $\Delta h$ the height variation amplitude.

\paragraph{Ping-pong trajectory.}

We define the oscillation fraction $f(t) \in [0, 1]$ as:
\begin{equation}
f(t) =
\begin{cases}
\dfrac{1 - \cos(2\pi t)}{2} & \text{(smooth turnaround)} \\[6pt]
1 - 2\left|(t - \lfloor t \rfloor) - \tfrac{1}{2}\right| & \text{(constant speed)}
\end{cases}
\end{equation}

The azimuth angle is then interpolated:
\begin{equation}
\theta(t) = \theta_{\text{start}} + f(t)\,(\theta_{\text{end}} - \theta_{\text{start}})
\end{equation}
and the camera position still follows Equation~\ref{eq:camera_center}.
The camera now performs a single back and forth sweep along a circular arc.

\subsection{AiP-real}
\label{supp:aip_real}

Within AiP-real, the \texttt{globe} and \texttt{excavator} sequences are extracted from YouTube videos. The strong performance of our method on these sequences highlights its practicality and robustness in an in-the-wild setting.

Ground-truth labels are difficult to obtain on real-world objects. We leveraged SAM2~\cite{ravi2024sam2} to assist with segmentation annotations. Annotating a joint axis is significantly more ambiguous for a human. To address this, we leveraged our physical understanding of the objects: we manually selected two points lying on the joint axis from the images, and back-projected them to 3D using the depth maps to recover the 3D axis.

\section{Additional Qualitative Results on AiP-synth}

Figure~\ref{fig:qual_fig_merged} shows qualitative results on the remaining sequences of the AiP-synth dataset. Our approach again yields better results than competing methods.


\def\qualitWidth{0.20\linewidth}

\setlength{\tabcolsep}{0pt}

\begin{figure*}[htb]
  \centering

\resizebox{0.9\linewidth}{!}{
\begin{tabular}{c@{$\;$}c@{$\;\;$}cccccccccc}

\rotatebox{90}{\hspace{0.7cm}\vphantom{A}Ground} &
\rotatebox{90}{\hspace{0.8cm}\vphantom{A}Truth} &
\includegraphics[trim={3cm 2cm 3cm 0cm},clip,width=\qualitWidth]{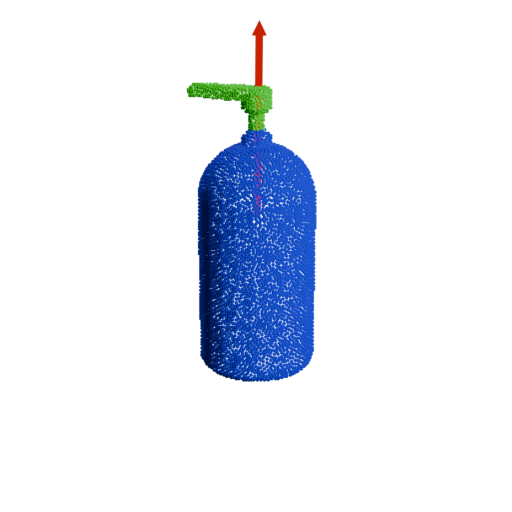} &
\includegraphics[trim={3cm 3cm 3cm 3cm},clip,width=\qualitWidth]{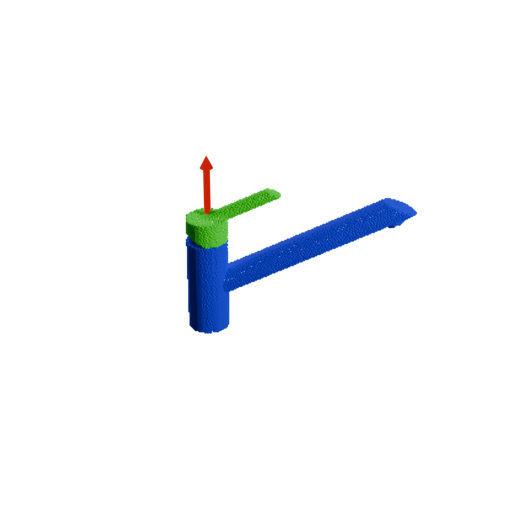} &
\includegraphics[trim={2cm 3cm 3cm 2cm},clip,width=\qualitWidth]{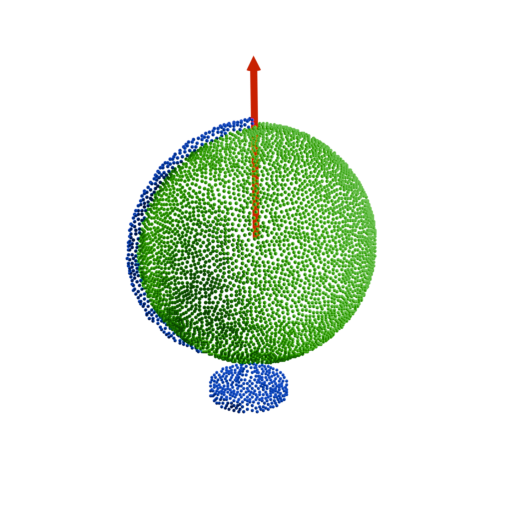}  &
\includegraphics[trim={2cm 3cm 3cm 1cm},clip,width=\qualitWidth]{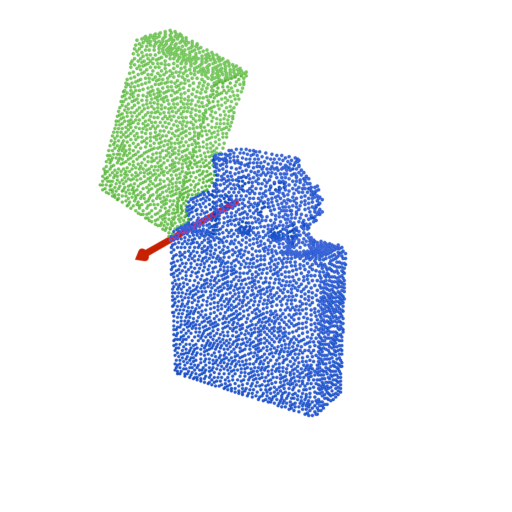}  &
\includegraphics[trim={2cm 3cm 3cm 1cm},clip,width=\qualitWidth]{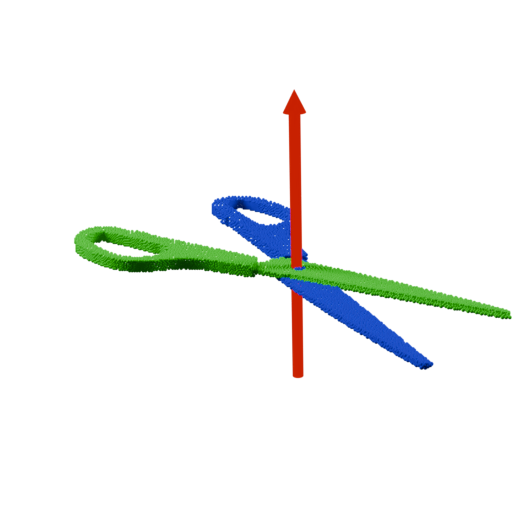}  &
\includegraphics[trim={2cm 2cm 2cm 3cm},clip,width=\qualitWidth]{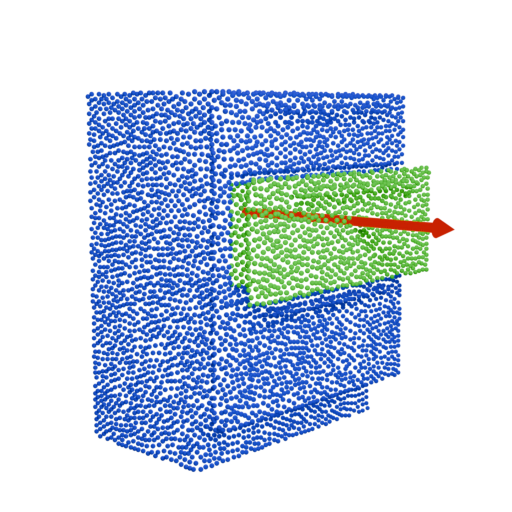}  &
\includegraphics[trim={3cm 2cm 3cm 2cm},clip,width=\qualitWidth]{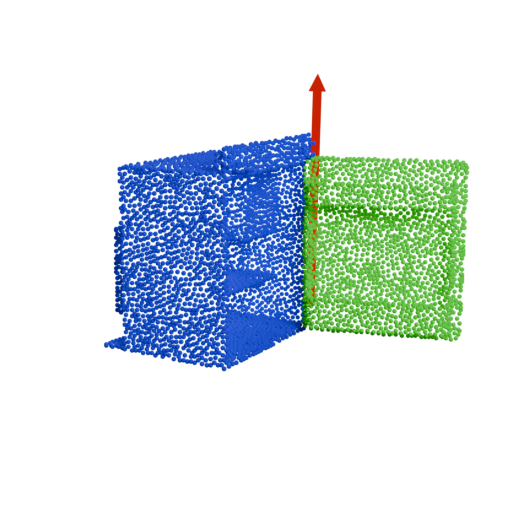}  &
\includegraphics[trim={3cm 2cm 3cm 2cm},clip,width=\qualitWidth]{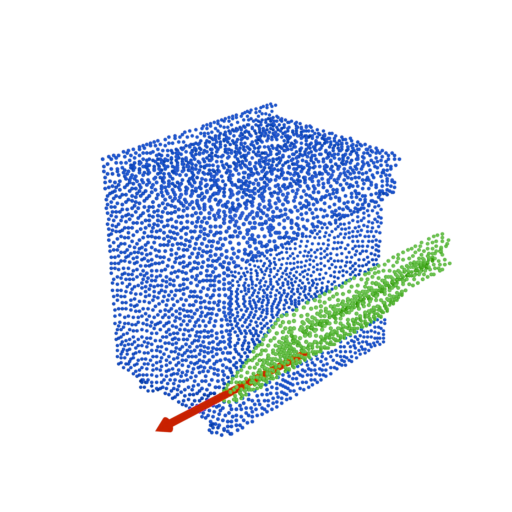}  &
\includegraphics[trim={3cm 3cm 3cm 2cm},clip,width=\qualitWidth]{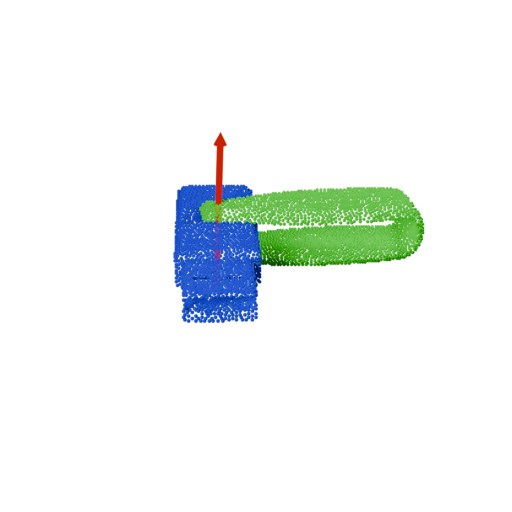} &
\includegraphics[trim={3cm 3cm 3cm 2cm},clip,width=\qualitWidth]{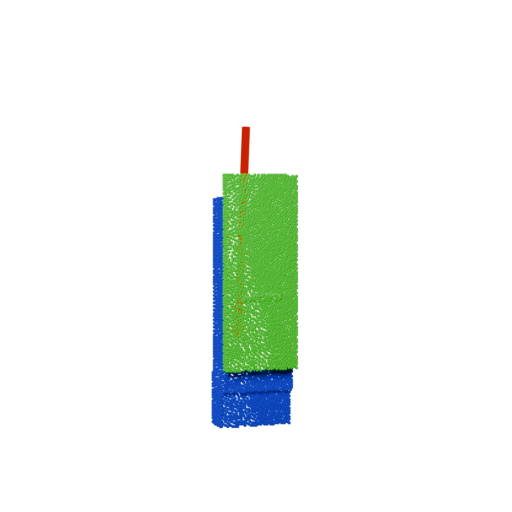}
\\[-0.0cm]

\rotatebox{90}{\hspace{0.7cm}} &
\rotatebox{90}{\hspace{0.8cm}Ours} &
\includegraphics[trim={3cm 1cm 3cm 0cm},clip,width=\qualitWidth]{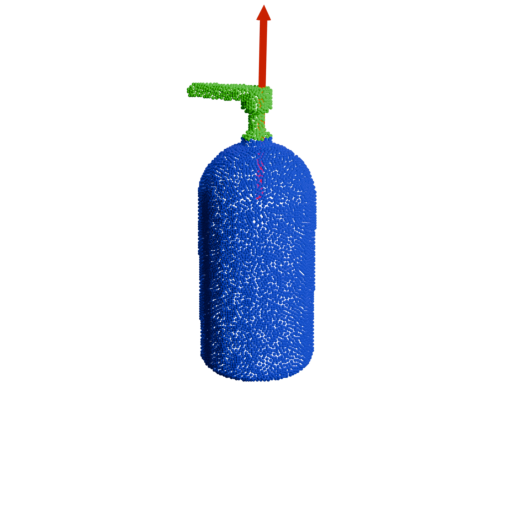}  &
\includegraphics[trim={3cm 3cm 3cm 3cm},clip,width=\qualitWidth]{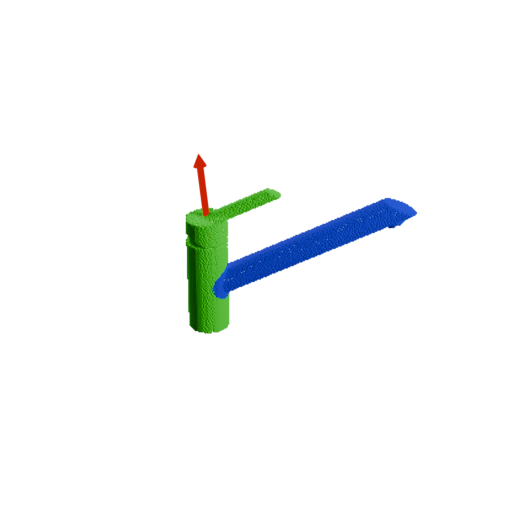} &
\includegraphics[trim={3cm 3cm 3cm 3cm},clip,width=\qualitWidth]{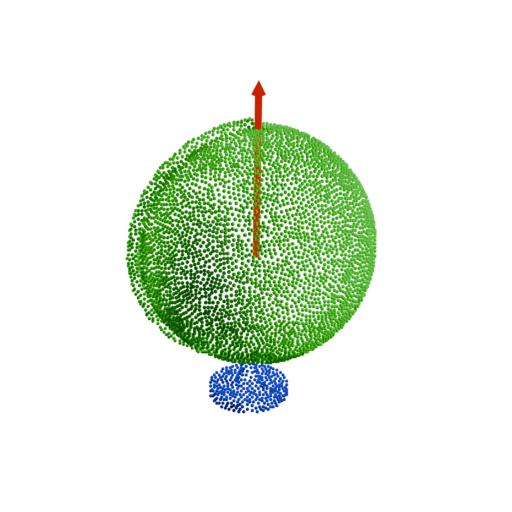} &
\includegraphics[trim={3cm 2cm 3cm 1cm},clip,width=\qualitWidth]{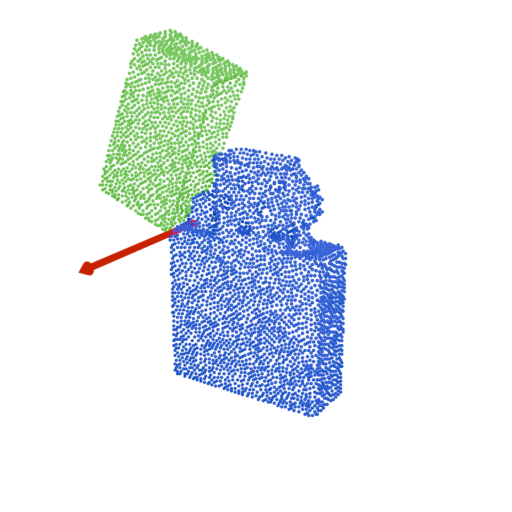} &
\includegraphics[trim={3cm 3cm 3cm 3cm},clip,width=\qualitWidth]{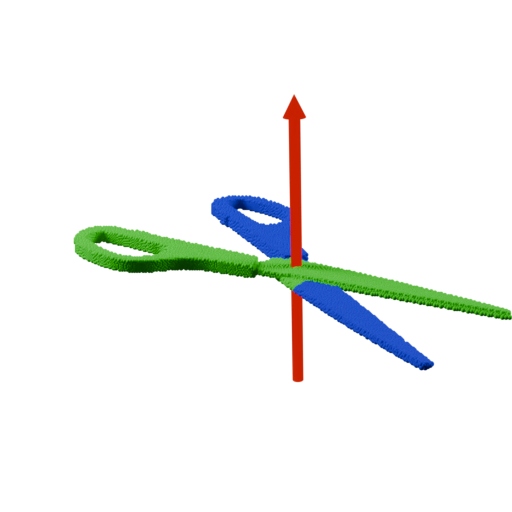} &
\includegraphics[trim={2cm 2cm 2cm 2cm},clip,width=\qualitWidth]{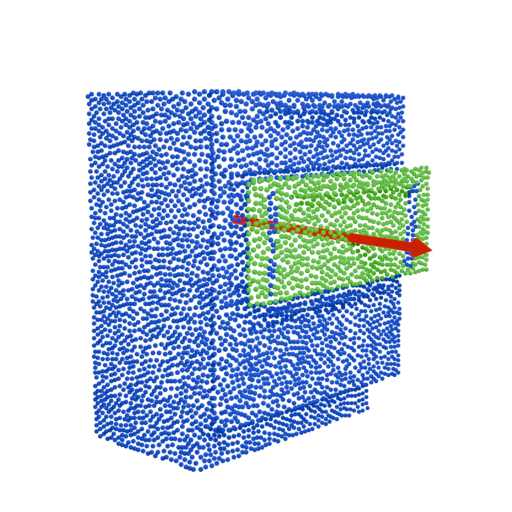} &
\includegraphics[trim={3cm 2cm 3cm 2cm},clip,width=\qualitWidth]{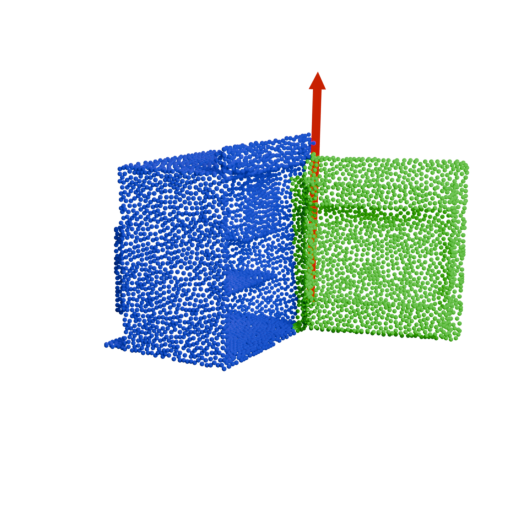} &
\includegraphics[trim={3cm 2cm 3cm 2cm},clip,width=\qualitWidth]{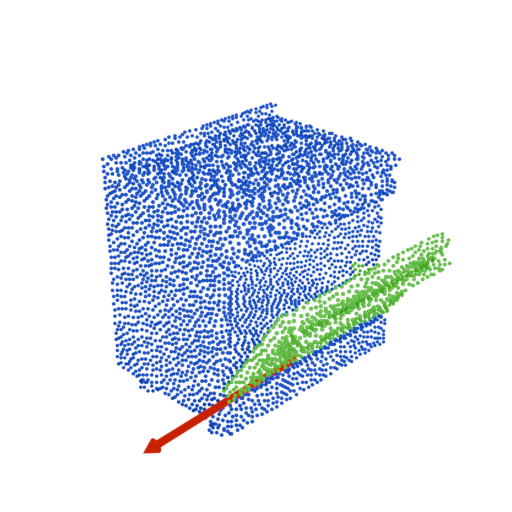} &
\includegraphics[trim={3cm 3cm 3cm 3cm},clip,width=\qualitWidth]{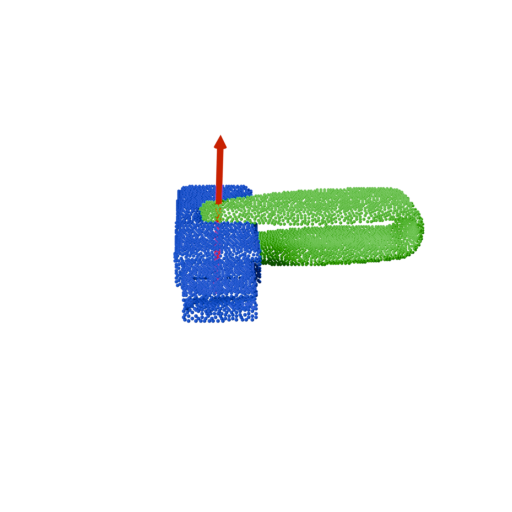} & 
\includegraphics[trim={2cm 2cm 2cm 2cm},clip,width=\qualitWidth]{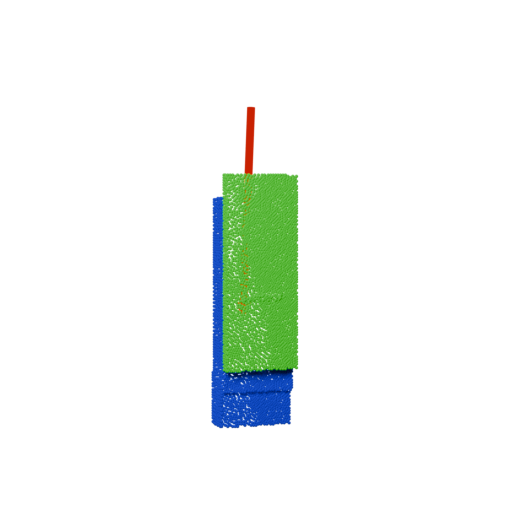}  
\\[-0.0cm]

\rotatebox{90}{\;\;\;\;\;\;Articulate-} &
\rotatebox{90}{\;\;\;\;\;\;Anything} &
\includegraphics[trim={3cm 3cm 3cm 2cm},clip,width=\qualitWidth]{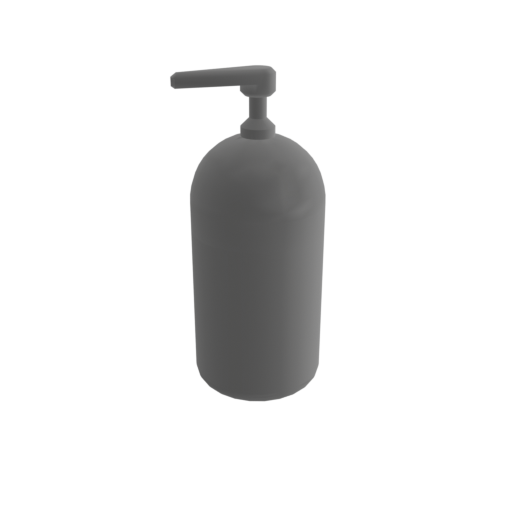}&
\epicfail&
\epicfail &
\includegraphics[trim={2cm 3cm 2cm 2cm},clip,width=\qualitWidth]{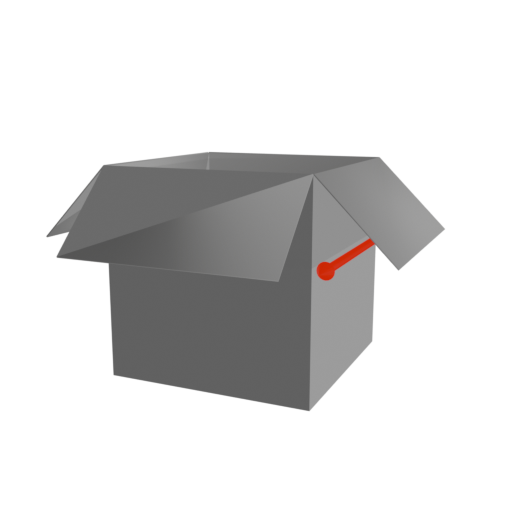}
&
\epicfail &
\includegraphics[trim={3cm 3cm 3cm 2cm},clip,width=\qualitWidth]{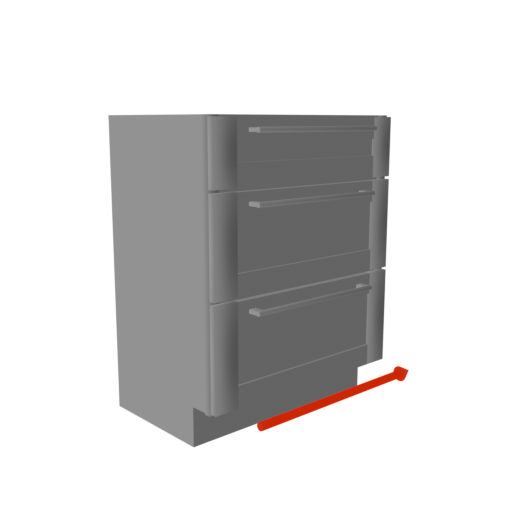} &
\epicfail &
\includegraphics[trim={3cm 2cm 3cm 2cm},clip,width=\qualitWidth]{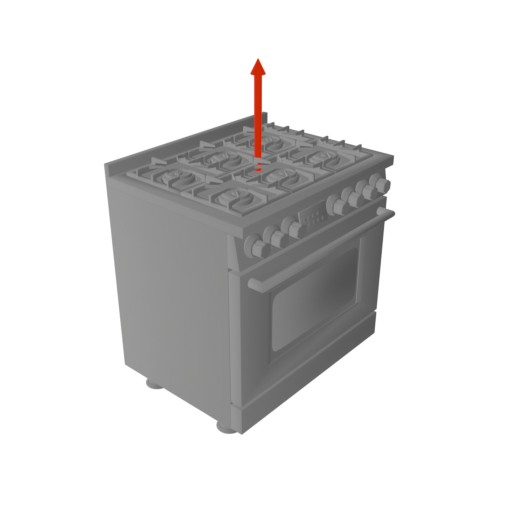}  
& \epicfail &
\includegraphics[trim={2cm 2cm 2cm 2cm},clip,width=\qualitWidth]{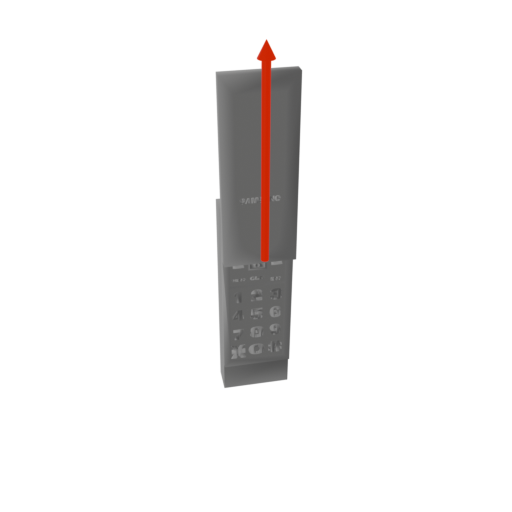}  
\\ [-0.5cm]

&
\rotatebox{90}{\hspace{1.0cm}\vphantom{A}ReArt} &
\epicfail &
\epicfail &
\epicfail &
\epicfail &
\includegraphics[trim={3cm 3cm 3cm 3cm},clip,width=\qualitWidth]{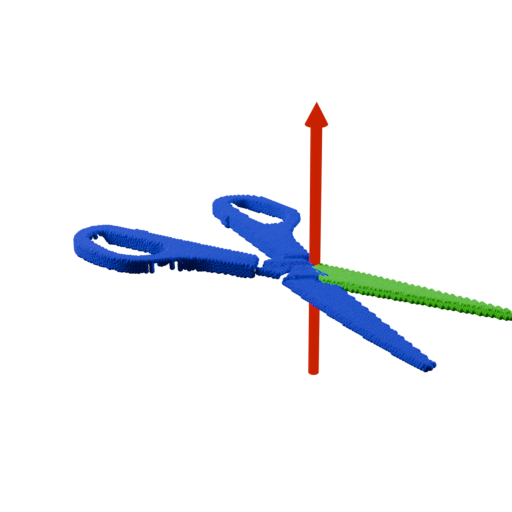} &
\epicfail &
\epicfail &
\includegraphics[trim={3cm 3cm 3cm 3cm},clip,width=\qualitWidth]{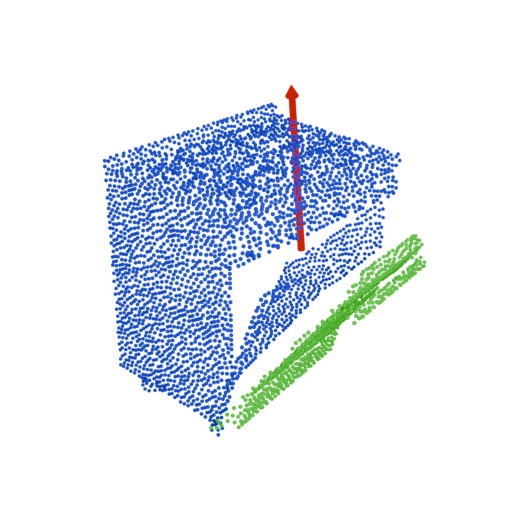} &
\epicfail &
\epicfail 
\\ [-0.5cm]

\rotatebox{90}{\hspace{0.6cm}Video2} &
\rotatebox{90}{\hspace{0.4cm}Articulation} &
\includegraphics[trim={3cm 3cm 3cm 2cm},clip,width=\qualitWidth]{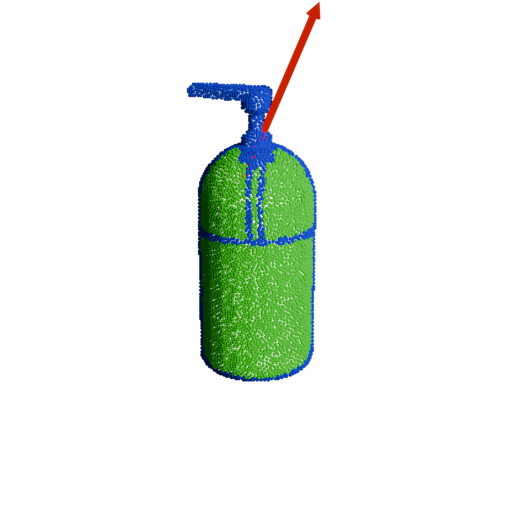}  &
\includegraphics[trim={2cm 2cm 2cm 2cm},clip,width=\qualitWidth]{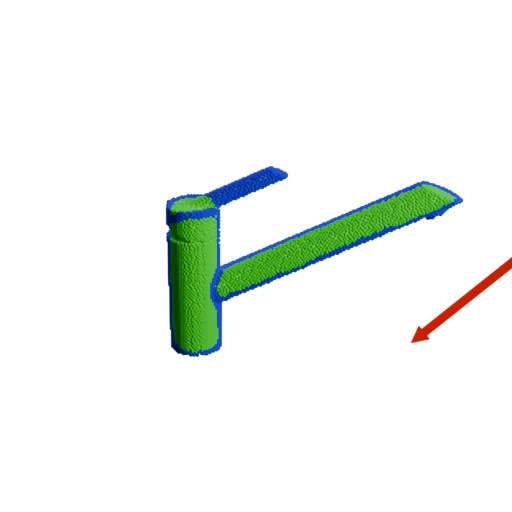}  &
\includegraphics[trim={2cm 2cm 2cm 2cm},clip,width=\qualitWidth]{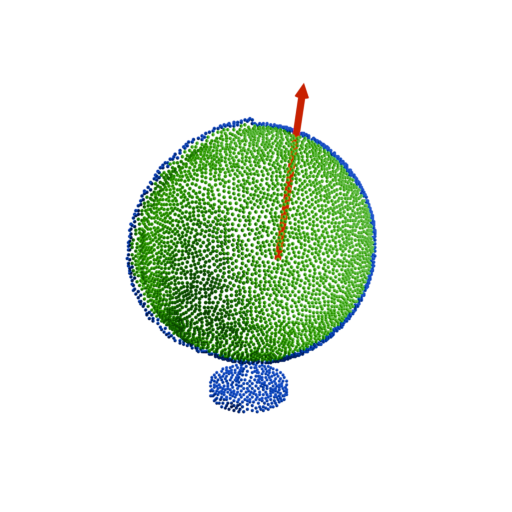}  &
\epicfail &
\includegraphics[trim={3cm 3cm 3cm 3cm},clip,width=\qualitWidth]{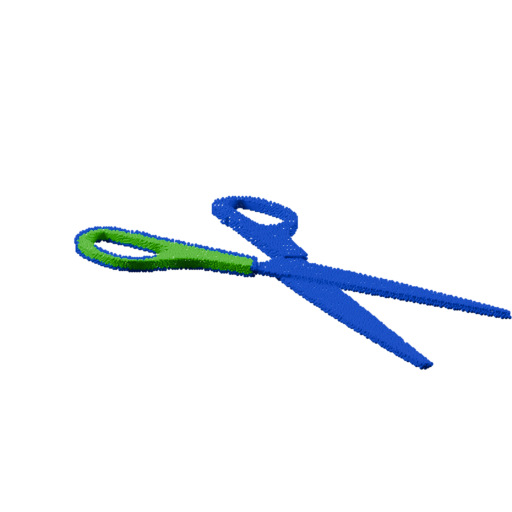}  &
\includegraphics[trim={2cm 2cm 2cm 2cm},clip,width=\qualitWidth]{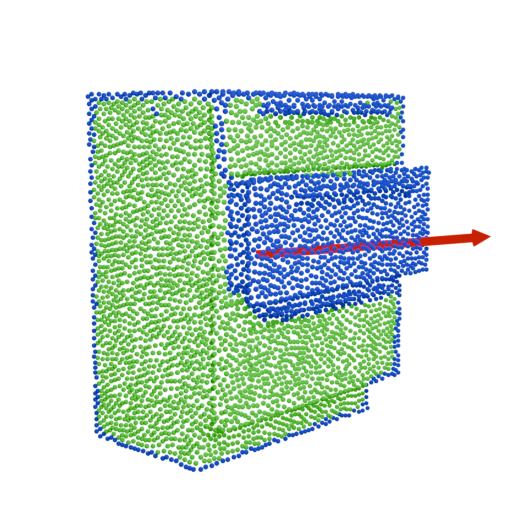}  &
\includegraphics[trim={3cm 3cm 3cm 3cm},clip,width=\qualitWidth]{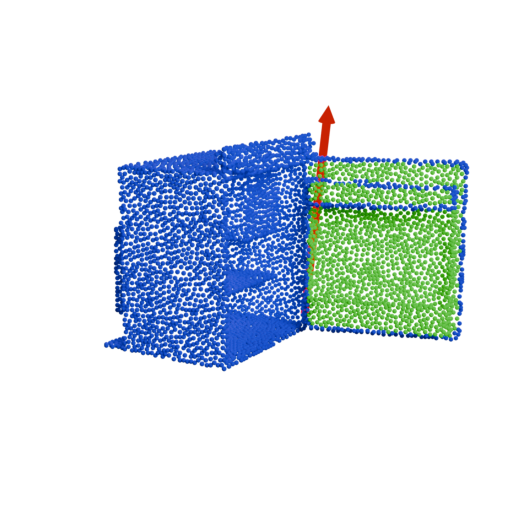}  &
\epicfail &
\includegraphics[trim={3cm 3cm 3cm 3cm},clip,width=\qualitWidth]{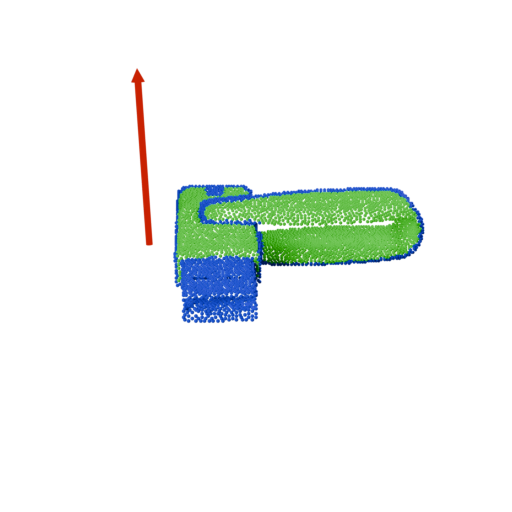} &
\includegraphics[trim={2cm 2cm 2cm 2cm},clip,width=\qualitWidth]{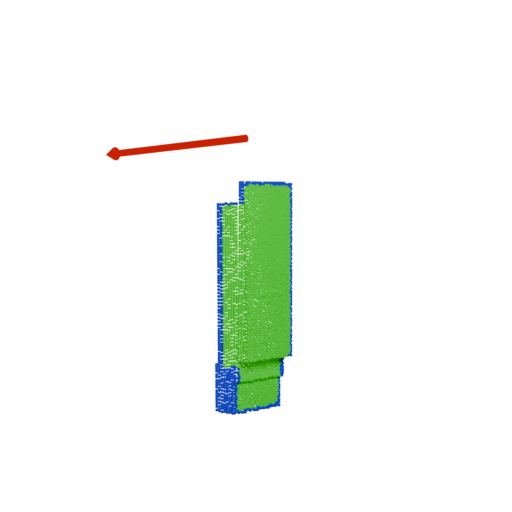}  
\\[-0.1cm]

&  & Bottle & Faucet & Globe & Lighter &Scissors & Drawer & Fridge & Oven & USB & Phone\\

\end{tabular}
} 
   \caption{\textbf{Qualitative results on the rest of the sequences from the AiP-synth dataset.} Red arrows denote predicted joint axes. Note how our method retrieves the part segmentations and the joint axes much more accurately and robustly than all the other methods. }
   \label{fig:qual_fig_merged}
\end{figure*}

\section{Visualization of Object Decompositions into Primitives and Parts}
\label{supp:viz_superquadric}

Beyond serving as proxies to group object points into geometrically and dynamically consistent regions, primitives also provide interpretability. We present primitive decompositions for AiP-synth and AiP-real objects in Figures~\ref{fig:qual_fig_sq} and~\ref{fig:qual_fig_sq_real}, respectively. We observe that certain parts automatically aggregate multiple primitives to reconstruct complex object structures beyond the modeling capacity of a single superquadric, such as the lamp base, the chair body or the excavator arm. Despite their limited expressivity, which prevents faithful reconstruction of real-world objects shapes, they prove sufficient for part segmentation and articulation discovery, supporting our design choice.


\def\qualitWidth{0.20\linewidth}

\setlength{\tabcolsep}{0pt}

\begin{figure*}[t]
  \centering

\resizebox{0.9\linewidth}{!}{
\begin{tabular}{c@{$\;$}c@{$\;\;$}cccccc}

\rotatebox{90}{\hspace{0.7cm}\vphantom{A}Representative} &
\rotatebox{90}{\hspace{1.2cm}\vphantom{A}Frame} &
\includegraphics[trim={3cm 2cm 3cm 1cm},clip,width=\qualitWidth]{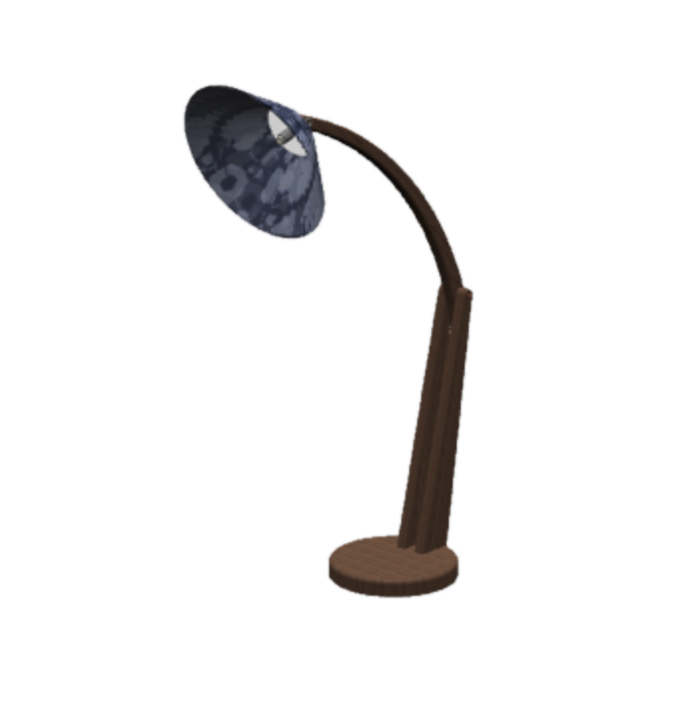} &
\includegraphics[trim={3cm 3cm 3cm 3cm},clip,width=\qualitWidth]{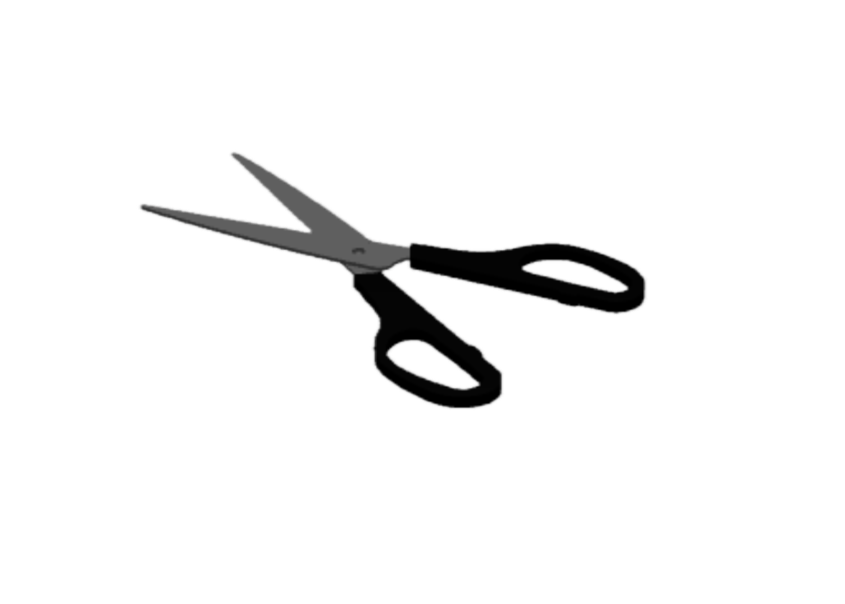} &
\includegraphics[trim={1cm 1cm 1cm 1cm},clip,width=\qualitWidth]{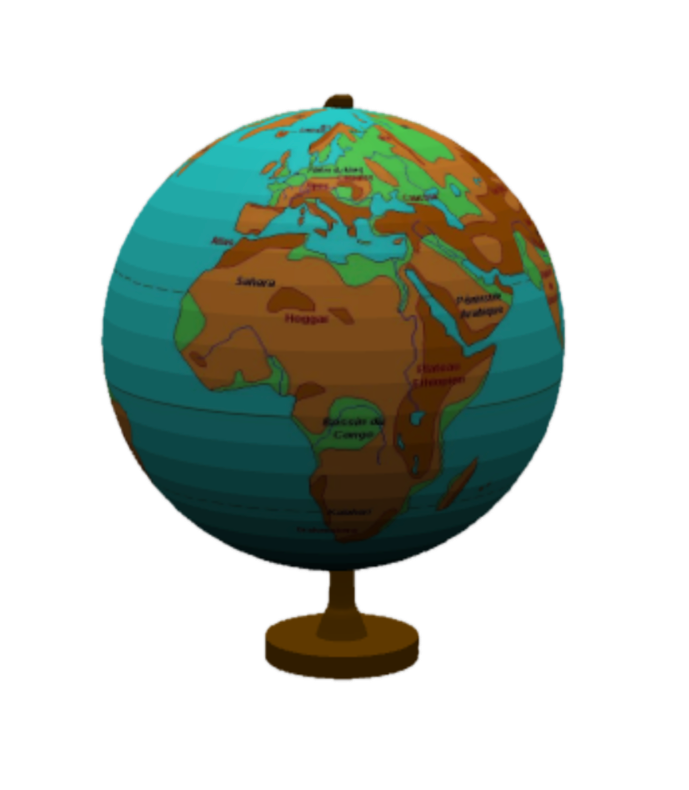} &
\includegraphics[trim={1cm 3cm 1cm 3cm},clip,width=\qualitWidth]{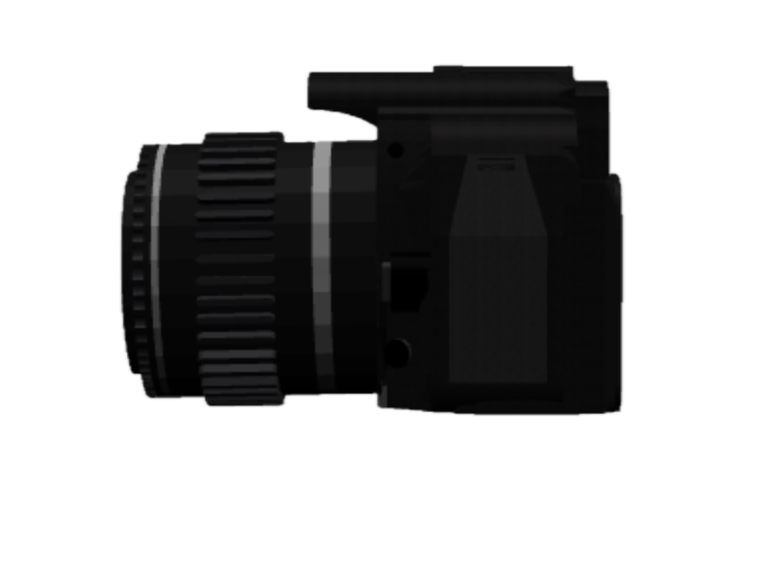} &
\includegraphics[trim={4cm 5cm 4cm 5cm},clip,width=\qualitWidth]{images/our_dataset_results/representative_figs/blade.png}  &
\includegraphics[trim={3cm 3cm 3cm 2cm},clip,width=\qualitWidth]{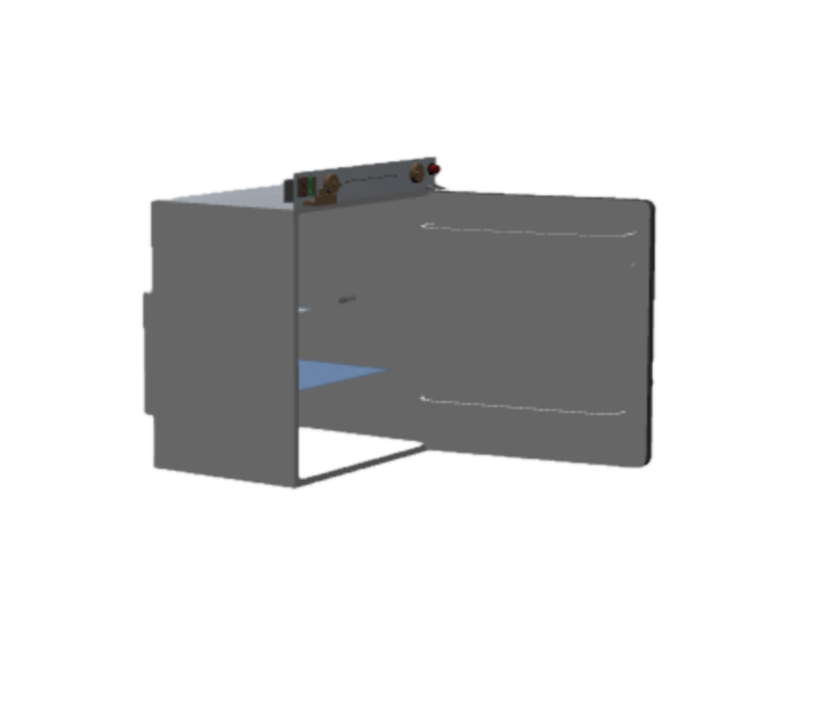} \\[-0.0cm]

\rotatebox{90}{\hspace{0.7cm} Superquadrics} &
\rotatebox{90}{\hspace{0.8cm}} &
\includegraphics[trim={3cm 1cm 3cm 1cm},clip,width=\qualitWidth]{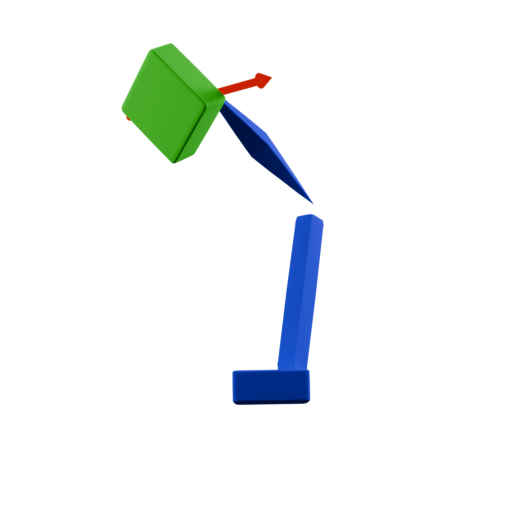}  &
\includegraphics[trim={2cm 1cm 2cm 1cm},clip,width=\qualitWidth]{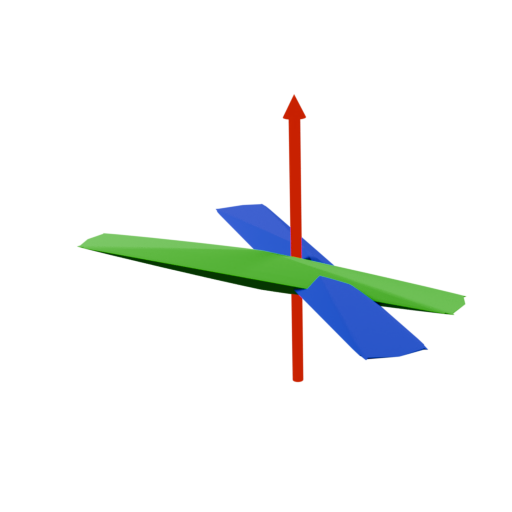}  &
\includegraphics[trim={2cm 1cm 2cm 1cm},clip,width=\qualitWidth]{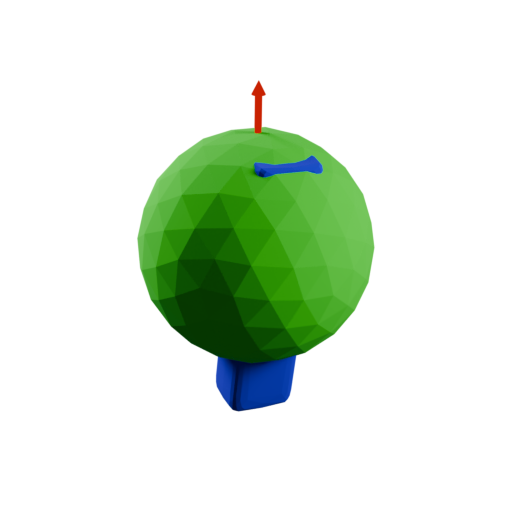}  &
\includegraphics[trim={1cm 3cm 1cm 3cm},clip,width=\qualitWidth]{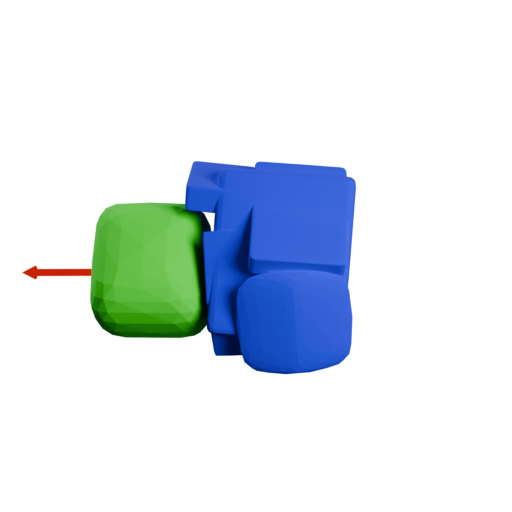}  &
\includegraphics[trim={2cm 3cm 2cm 3cm},clip,width=\qualitWidth]{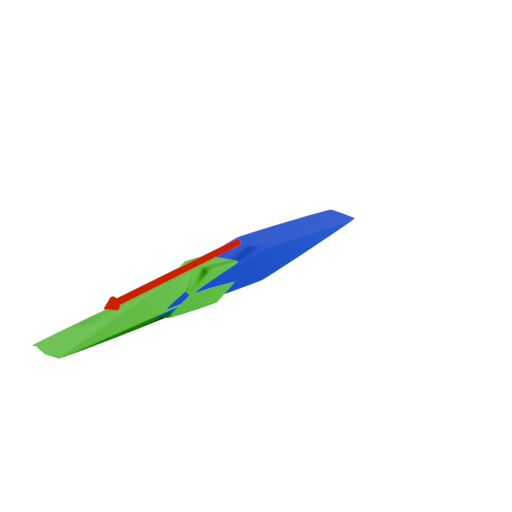}  &
\includegraphics[trim={2cm 5cm 3cm 1cm},clip,width=\qualitWidth]{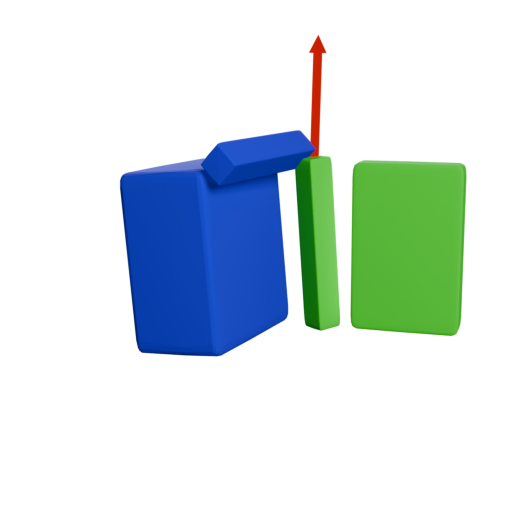}  
\\[-0.0cm]

& & Lamp & Scissors & Globe & Camera & Blade & Fridge \\

\end{tabular}
} 
   \caption{\textbf{Fitted superquadrics results on sequences from the AiP-synth dataset.} Red arrows denote predicted joint axes.   }
   \label{fig:qual_fig_sq}
\end{figure*}


\def\qualitWidth{0.20\linewidth}

\setlength{\tabcolsep}{0pt}

\begin{figure*}[t]
  \centering

\resizebox{0.9\linewidth}{!}{
\begin{tabular}{c@{$\;$}c@{$\;\;$}ccccc}

\rotatebox{90}{\hspace{0.7cm}\vphantom{A}Representative} &
\rotatebox{90}{\hspace{1.2cm}\vphantom{A}Frame} &
\includegraphics[trim={2cm 1.43cm 1cm 1.1cm},clip,width=\qualitWidth]{images/real_data_qual_res/rep_frame/box.png} &
\includegraphics[trim={2cm 1cm 1cm 1cm},clip,width=\qualitWidth]{images/real_data_qual_res/rep_frame/chair.png} &
\includegraphics[trim={2cm 1.4cm 1cm 1.1cm},clip,width=\qualitWidth]{images/real_data_qual_res/rep_frame/sliding_box.png} &
\includegraphics[trim={2cm 1.4cm 1cm 1.1cm},clip,width=\qualitWidth]{images/real_data_qual_res/rep_frame/globe.png} &
\includegraphics[trim={2cm 0.9cm 1cm 1.1cm},clip,width=\qualitWidth]{images/real_data_qual_res/rep_frame/excavator.png}  
\\[-0.0cm]

\rotatebox{90}{\hspace{0.7cm} Superquadrics} &
\rotatebox{90}{\hspace{0.8cm}} &
\includegraphics[trim={2cm 1cm 2cm 1cm},clip,width=\qualitWidth]{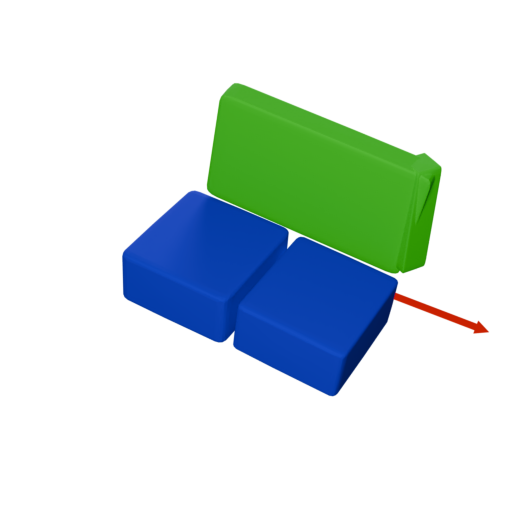}  &
\includegraphics[trim={2cm 1cm 2cm 1cm},clip,width=\qualitWidth]{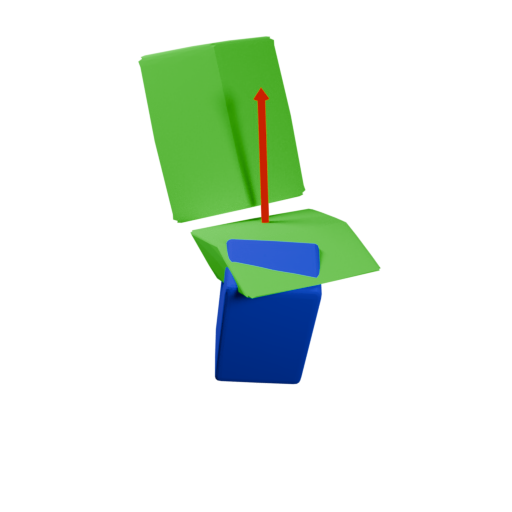}  &
\includegraphics[trim={2cm 1cm 2cm 1cm},clip,width=\qualitWidth]{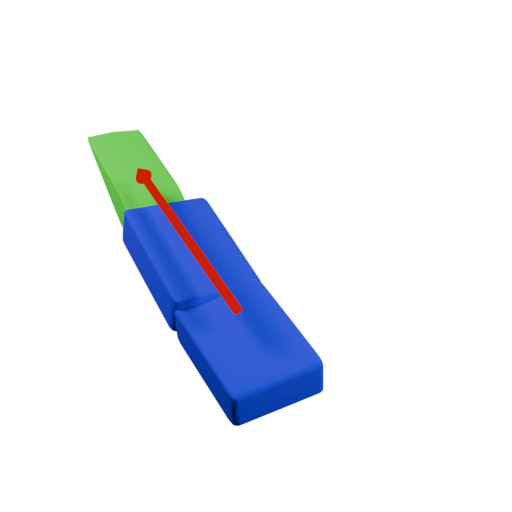}  &
\includegraphics[trim={2cm 1cm 2cm 1cm},clip,width=\qualitWidth]{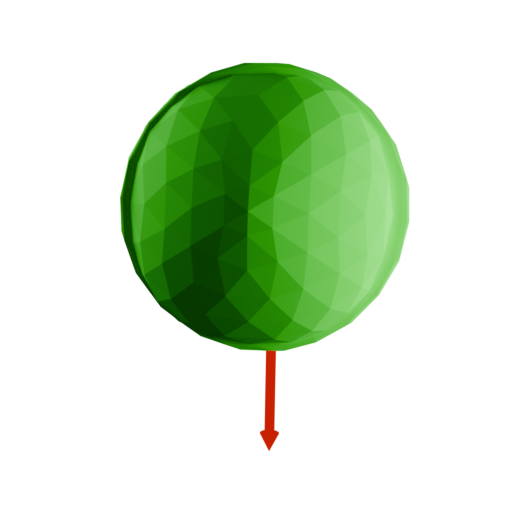}  &
\includegraphics[trim={2cm 1cm 2cm 1cm},clip,width=\qualitWidth]{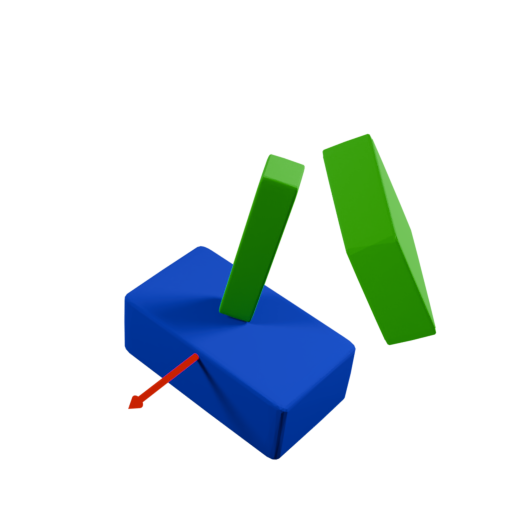}  

\\[-0.0cm]

& & Box & Chair & Sliding Box & Globe & Excavator  \\

\end{tabular}
} 
   \caption{\textbf{Fitted superquadrics results on the sequences from the AiP-real dataset.} Red arrows denote predicted joint axes.   }
   \label{fig:qual_fig_sq_real}
\end{figure*}

\clearpage
\section*{NeurIPS Paper Checklist}

\begin{enumerate}

\item {\bf Claims}
    \item[] Question: Do the main claims made in the abstract and introduction accurately reflect the paper's contributions and scope?
    \item[] Answer: \answerYes{}
    \item[] Justification: In the context of articulated object understanding, our abstract and introduction present our primitive-based optimization approach that uses superquadric primitives as a proxy representation to predict part segmentation into static and dynamic and articulation parameters. In both sections the main limitations of previous work and our contributions are presented.

\item {\bf Limitations}
    \item[] Question: Does the paper discuss the limitations of the work performed by the authors?
    \item[] Answer: \answerYes{}
    \item[] Justification: We have included a Limitations section discussing the main limitations of our approach.

\item {\bf Theory assumptions and proofs}
    \item[] Question: For each theoretical result, does the paper provide the full set of assumptions and a complete (and correct) proof?
    \item[] Answer: \answerNA{}
    \item[] Justification: Our approach does not involve theoretical mathematical proofs, thus Theorems and Lemmas are not included in the paper.

    \item {\bf Experimental result reproducibility}
    \item[] Question: Does the paper fully disclose all the information needed to reproduce the main experimental results of the paper to the extent that it affects the main claims and/or conclusions of the paper (regardless of whether the code and data are provided or not)?
    \item[] Answer: \answerYes{}
    \item[] Justification: We have included detailed assumptions made in order to ensure reproducibility of our results. A more detailed implementation description is included in the supplemental.

\item {\bf Open access to data and code}
    \item[] Question: Does the paper provide open access to the data and code, with sufficient instructions to faithfully reproduce the main experimental results, as described in supplemental material?
    \item[] Answer: \answerYes{}
    \item[] Justification: Our code and data will be made publicly available, and scripts to reproduce the results of the paper will be provided.

\item {\bf Experimental setting/details}
    \item[] Question: Does the paper specify all the training and test details (e.g., data splits, hyperparameters, how they were chosen, type of optimizer) necessary to understand the results?
    \item[] Answer: \answerYes{}
    \item[] Justification: All the parameters of our approach, experiments and ablations are included in the paper, and the supplemental material includes extra resutls and details.

\item {\bf Experiment statistical significance}
    \item[] Question: Does the paper report error bars suitably and correctly defined or other appropriate information about the statistical significance of the experiments?
    \item[] Answer: \answerYes{}
    \item[] Justification: Our experiments, in Table 1,3, and 4 report average results across the dataset with standard deviation. For individual object results (Table 2) we report best results based on lowest loss on a set of 5 independent runs, and for ablations (Table 5) we only report average results.

\item {\bf Experiments compute resources}
    \item[] Question: For each experiment, does the paper provide sufficient information on the computer resources (type of compute workers, memory, time of execution) needed to reproduce the experiments?
    \item[] Answer: \answerYes{}
    \item[] Justification: We have included details on the machine used to run our experiments in the the supplemental.
    
\item {\bf Code of ethics}
    \item[] Question: Does the research conducted in the paper conform, in every respect, with the NeurIPS Code of Ethics \url{https://neurips.cc/public/EthicsGuidelines}?
    \item[] Answer: \answerYes{}
    \item[] Justification: Our research conforms to NeurIPS guidelines.

\item {\bf Broader impacts}
    \item[] Question: Does the paper discuss both potential positive societal impacts and negative societal impacts of the work performed?
    \item[] Answer: \answerNA{}
    \item[] Justification: Our paper does not pose a risk of societal impacts.
    
\item {\bf Safeguards}
    \item[] Question: Does the paper describe safeguards that have been put in place for responsible release of data or models that have a high risk for misuse (e.g., pre-trained language models, image generators, or scraped datasets)?
    \item[] Answer: \answerNA{}
    \item[] Justification: Our paper does not pose such risks.

\item {\bf Licenses for existing assets}
    \item[] Question: Are the creators or original owners of assets (e.g., code, data, models), used in the paper, properly credited and are the license and terms of use explicitly mentioned and properly respected?
    \item[] Answer: \answerYes{}
    \item[] Justification: Data from previous work is properly referenced.

\item {\bf New assets}
    \item[] Question: Are new assets introduced in the paper well documented and is the documentation provided alongside the assets?
    \item[] Answer: \answerYes{}
    \item[] Justification: New data introduced in the paper is captured either by the authors or retrieved from publicly available sources.

\item {\bf Crowdsourcing and research with human subjects}
    \item[] Question: For crowdsourcing experiments and research with human subjects, does the paper include the full text of instructions given to participants and screenshots, if applicable, as well as details about compensation (if any)? 
    \item[] Answer: \answerNA{}
    \item[] Justification:  Our method did not involve a study with human subjects.

\item {\bf Institutional review board (IRB) approvals or equivalent for research with human subjects}
    \item[] Question: Does the paper describe potential risks incurred by study participants, whether such risks were disclosed to the subjects, and whether Institutional Review Board (IRB) approvals (or an equivalent approval/review based on the requirements of your country or institution) were obtained?
    \item[] Answer: \answerNA{}
    \item[] Justification: Our method did not involve a study with human subjects.

\item {\bf Declaration of LLM usage}
    \item[] Question: Does the paper describe the usage of LLMs if it is an important, original, or non-standard component of the core methods in this research? Note that if the LLM is used only for writing, editing, or formatting purposes and does \emph{not} impact the core methodology, scientific rigor, or originality of the research, declaration is not required.
    \item[] Answer: \answerNA{}
    \item[] Justification: Our method does not involve usage of LLMs.

\end{enumerate}

\end{document}